\documentclass[11pt]{article}

\usepackage[final]{acl}

\usepackage{times}
\usepackage{latexsym}

\usepackage[T1]{fontenc}
\usepackage[utf8]{inputenc}

\usepackage{microtype}

\usepackage{inconsolata}

\usepackage{graphicx}

\usepackage{paralist}
\usepackage{amsmath}
\usepackage{amssymb}
\usepackage{tcolorbox}
\tcbuselibrary{skins, listings, breakable}
\usepackage{xcolor}
\usepackage{algorithm}
\usepackage{algpseudocode}
\usepackage{adjustbox}
\usepackage{booktabs}
\usepackage{multirow}
\usepackage{xcolor,colortbl}
\newcommand{\bestres}[1]{\textbf{\textcolor{red}{#1}}}
\newcommand{\secondres}[1]{\textcolor{blue}{\underline{#1}}}
\usepackage{colortbl}
\usepackage{graphicx}
\usepackage{subcaption}
\usepackage{tabularx}
\usepackage{caption}
\usepackage{array}
\usepackage{hyperref}

\usepackage{listings}
\usepackage{xcolor}

\lstdefinelanguage{json}{
  basicstyle=\ttfamily\small,
  numbers=left,
  numberstyle=\tiny,
  stepnumber=1,
  numbersep=6pt,
  showstringspaces=false,
  breaklines=true,
  frame=single,
  backgroundcolor=\color{gray!5},
  tabsize=2
}

\title{STReasoner: Empowering LLMs for Spatio-Temporal Reasoning in Time Series via Spatial-Aware Reinforcement Learning}

\author{
Juntong Ni$^{1}$, Shiyu Wang\thanks{Corresponding Authors.},
Qi He$^{2}$, Ming Jin$^{3}$, Wei Jin$^{1*}$
\\
$^{1}$Emory University, $^{2}$Microsoft, $^{3}$Griffith University \\
\texttt{\{juntong.ni, wei.jin\}@emory.edu}
}

\begin{document}
\maketitle
\begin{abstract}
Spatio-temporal reasoning in time series involves the explicit synthesis of temporal dynamics, spatial dependencies, and textual context. This capability is vital for high-stakes decision-making in systems such as traffic networks, power grids, and disease propagation. However, the field remains underdeveloped because most existing works prioritize predictive accuracy over reasoning. To address the gap, we introduce \textbf{ST-Bench}, a benchmark consisting of four core tasks, including etiological reasoning, entity identification, correlation reasoning, and in-context forecasting, developed via a \emph{network SDE-based multi-agent data synthesis pipeline}. We then propose \textbf{STReasoner}, which empowers LLM to integrate time series, graph structure, and text for explicit reasoning. To promote spatially grounded logic, we introduce \emph{S-GRPO}, a reinforcement learning algorithm that rewards performance gains specifically attributable to spatial information. Experiments show that STReasoner achieves average accuracy gains between $17\%$ and $135\%$ at only $0.004\times$ the cost of proprietary models and generalizes robustly to real-world data. Our code is available at \href{https://github.com/LingFengGold/STReasoner}{https://github.com/LingFengGold/STReasoner}.
\end{abstract}

\section{Introduction}

Time series data are ubiquitous in real-world systems and often exhibit complex spatio-temporal structures~\cite{wang2020deep}, from traffic networks~\cite{liu2023largest} and climate systems~\cite{lin2018exploiting} to disease propagation across regions~\cite{liu2024review}.
As the community races to bring large language models (LLMs) to these domains~\cite{jin2023large}, it is critical to ensure that LMs can support decisions grounded in these data~\cite{merrill2024language}.

\begin{figure}[t!]
    \centering
    \includegraphics[width=\linewidth]{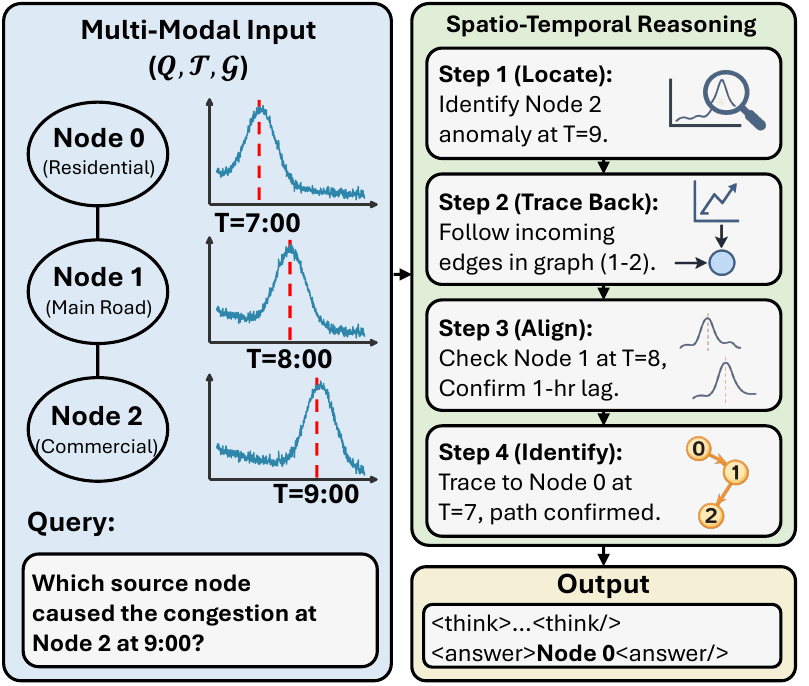}

    \caption{A traffic flow example of spatio-temporal reasoning. $Q$: query, $\mathcal{T}$: time series, $\mathcal{G}$: graph.}
    % \vskip -2em 
    \label{fig:introduction}
\end{figure}

However, most existing approaches primarily focus on predictive accuracy~\cite{liang2025foundation,liu2024spatial,liu2024can,li2024urbangpt}, offering limited support for decision-making that requires explicit reasoning rather than point forecasts.  
Thus, we focus on \textit{spatio-temporal reasoning}: the ability to answer natural-language queries about \emph{what happened, where, when, and why} in a system whose state evolves over time and is coupled across space (e.g., a graph).
Such queries require linking observations across \emph{multiple nodes} and \emph{multiple time steps} through the system's spatial dependencies and temporal dynamics, often involving propagation and time lag. As shown in Figure~\ref{fig:introduction}, when traffic congestion is observed, the query ``Which source node caused the congestion at Node~2 at 9:00?'' cannot be answered from a single-node prediction at 9:00; it requires tracing upstream dependencies in the traffic graph and connecting earlier upstream patterns to the downstream congestion under delay. Answering it thus integrates \emph{textual intent}, \emph{temporal evidence}, and \emph{spatial structure}. Spatio-temporal reasoning of this form is essential in many other critical systems,  such as tracking viral spread across urban populations~\cite{liu2024review}, assessing cascading impacts on power grids under extreme weather~\cite{panteli2015influence}, and tracing flood propagation along river networks~\cite{paola2006toward}.

Despite its importance, spatio-temporal reasoning in time series remains underexplored.
LLMs and vision language models (VLMs) demonstrate strong reasoning ability in textual and visual domains~\cite{guo2025deepseek,huang2025vision}, but they are not explicitly trained or evaluated for reasoning over \emph{spatio-temporal time series}. Existing spatio-temporal forecasting methods focus on predicting future values~\cite{chen2025learning,li2024urbangpt,yuan2024unist}, without explicitly reasoning, which limits their usefulness for decision-making. Meanwhile, time series language/reasoning models primarily consider univariate or multivariate series~\cite{guan2025timeomni,luo2025time,kong2025time,xie2024chatts,jin2023time,wang2023timemixer,wang2024timemixer++} and fail to consider \emph{spatial relationship}. 
As a result, spatio-temporal reasoning in time series is inherently distinct from these paradigms.

Spatio-temporal reasoning in time series faces three key fundamental challenges that hinder its development:
\textbf{(a) Data Challenges.} 
Existing spatio-temporal datasets rarely provide paired natural language descriptions to explain spatial entities, dependency, or temporal dynamics. Furthermore, current time series reasoning datasets often lack spatial interactions such as explicit graph structures.
\textbf{(b) Evaluation Gaps.}
There are no standardized, multi-dimensional benchmarks that decompose spatio-temporal reasoning into distinct reasoning tasks, making it difficult to systematically evaluate models’ capabilities across different spatio-temporal reasoning demands.
\textbf{(c) Modeling Limitations.} The effective architecture and optimization objective for spatio-temporal reasoning are largely underexplored.
\underline{\emph{First}}, it remains unclear how to effectively fuse time series, graph structure, and textual information for reasoning without sacrificing numerical precision or global context.
\underline{\emph{Second}}, it is unclear how to optimize models for spatio-temporal reasoning.
Existing RL methods rely on result-only rewards~\cite{guo2025deepseek} and fail to enforce spatial grounding, allowing models to exploit superficial temporal patterns rather than perform genuine spatial attribution.

To address these challenges, we propose three key solutions.
\textbf{First}, we introduce an \textit{SDE-based multi-agent data synthesis pipeline}~\cite{xu2025comprehensive}, which generates time series with controllable spatial dependencies and temporal evolution, together with aligned textual descriptions. 
\textbf{Second}, building on this pipeline, we build \emph{ST-Bench} and organize spatio-temporal reasoning into four tasks: \textit{etiological spatial reasoning}, \textit{spatial entity identification}, \textit{spatial correlation reasoning}, and \textit{in-context forecasting}, along with a real-world dataset for zero-shot evaluation.
\textbf{Third}, we propose \textit{STReasoner}, which integrates a dedicated time series encoder with LLM to process numerical time series and textual inputs jointly.
To explicitly incentivize spatial reasoning during training, we propose \textit{spatial-aware group relative policy optimization (S-GRPO)}, which assigns additional reward only when spatial information improves performance.

We conduct a comprehensive empirical study to evaluate spatio-temporal reasoning capabilities across existing language models and STReasoner.
Our contributions can be summarized as:

\begin{compactenum}[(a)] 

\item To the best of our knowledge, we take an early step toward formally studying the problem of spatio-temporal reasoning in time series.

\item We propose an SDE-based multi-agent pipeline to synthesize spatio-temporal data with aligned textual descriptions. Based on this pipeline, we construct ST-Bench to support evaluation.

\item We propose STReasoner, a unified spatio-temporal reasoning model, together with a spatial-aware RL objective.

\item Through systematic evaluation, we show that existing LMs struggle with spatio-temporal reasoning, while STReasoner achieves strong performance and robust zero-shot generalization at substantially lower cost.

\end{compactenum}

% \vspace{-0.15em}
\section{Related Work}
% \vspace{-0.2em}

Recent work has improved the reasoning ability of LLMs and extended them to non-text modalities~\cite{wei2022chain,guo2025deepseek,huang2025vision}. 
Meanwhile, time series language and reasoning models support QA and multi-step reasoning, but largely focus on univariate or multivariate series without spatial dependency~\cite{jin2023time,xie2024chatts,wang2025can}. 
Spatio-temporal reasoning has also been explored in videos and trajectories~\cite{feng2025video,li2025stbench}. 
In contrast, spatio-temporal reasoning in time series involves numerical signals on discrete entities, with graph-defined spatial dependencies and induced temporal dynamics.
A complete discussion of related work is provided in Appendix~\ref{app:related_work}.

\begin{figure*}[t!]
    \centering
    \includegraphics[width=0.98\textwidth]{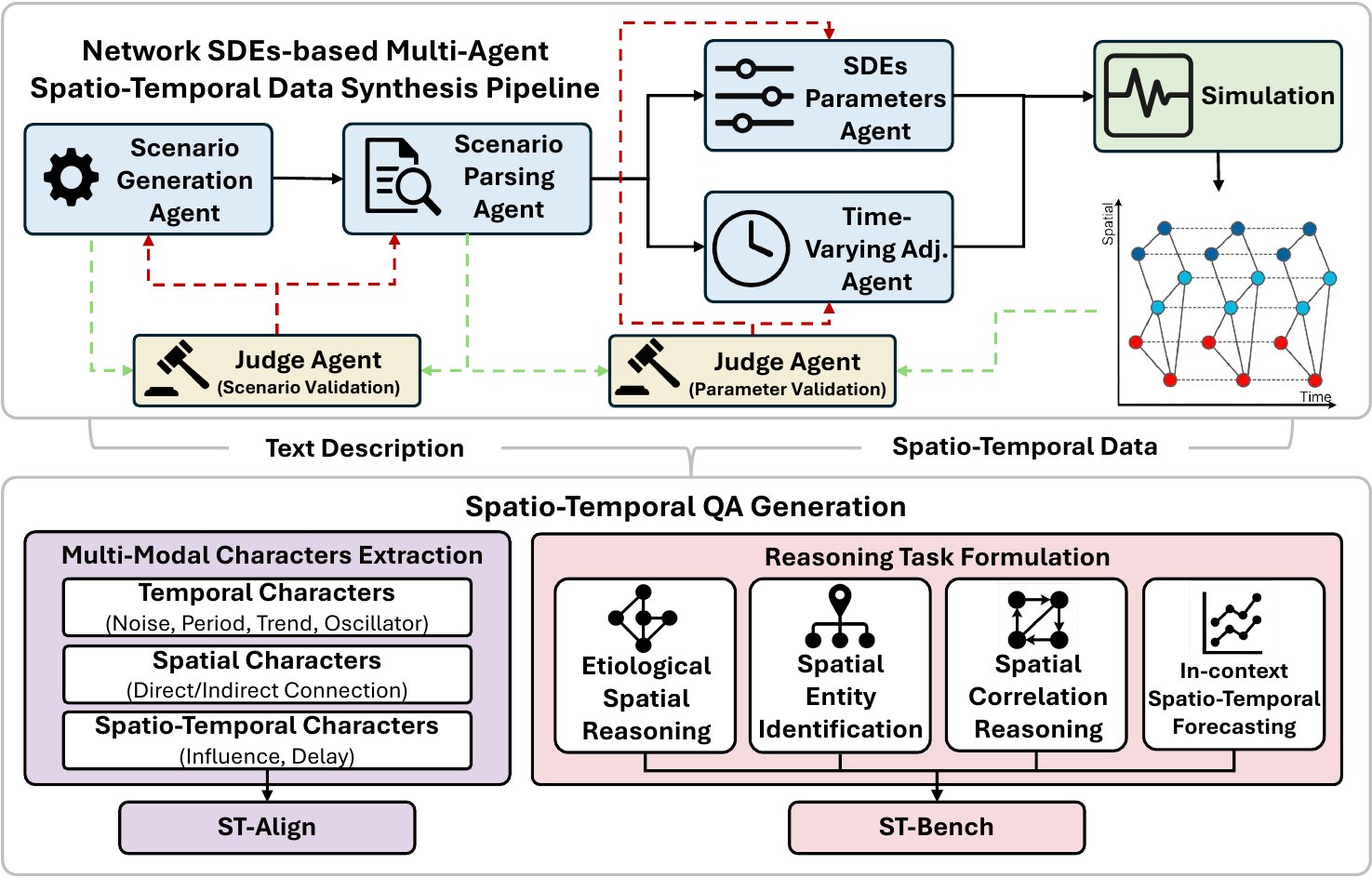}
    % \vskip -1em
    \caption{Overall framework of the Network SDEs-based multi-agent spatio-temporal data synthesis pipeline (upper) and spatio-temporal QA generation (lower). Detailed prompts for each agent are provided in Appendix~\ref{app:agent_prompt}, and the pseudo algorithm is given in Algorithm~\ref{alg:data_synthesis}.
}
    \label{fig:data_synthesis}
    % \vskip -1em
\end{figure*}

% \vspace{-0.2em}
\section{Problem and Dataset Construction}
% \vspace{-0.2em}

\subsection{Problem Definition}
We define spatio-temporal reasoning in time series as follows:
given a graph $\mathcal{G} = (\mathcal{V}, \mathcal{E})$ with $N$ nodes, 
a set of time series $\mathcal{T} = \{T_1, \dots, T_N\}$, where each $T_i = \{x_{i,1}, x_{i,2}, \dots, x_{i,T}\}, \quad x_{i,t} \in \mathbb{R}$ represents a sequence of $T$ observed values over time for the node $v_i \in \mathcal{V}$ and a natural language query $Q$, the goal is to generate a text-based response that requires explicit reasoning over temporal dynamics and spatial dependencies.
The model produces an intermediate reasoning sequence $R$ and a final answer $A$.
Formally, the task is defined as
$f : (Q, \mathcal{T}, \mathcal{G}) \rightarrow (R, A)$,
where $f$ denotes a model that interprets the query $Q$ and analyzes the time series $\mathcal{T}$ under the structural constraints imposed by the graph $\mathcal{G}$ to generate the reasoning process $R$ and the final answer $A$.

\subsection{Dataset Synthesis Pipeline}

\paragraph{Motivation.} High-quality datasets for spatio-temporal reasoning in time series are scarce, especially those that simultaneously provide controllable spatial dependencies, rich temporal dynamics, and aligned textual descriptions. Synthetic data, therefore, becomes a practical solution~\cite{xie2024chatts,fu2024synthetic,tremblay2018training}. However, naive generation methods often lack precise control over \emph{temporal dynamics} and \emph{spatial dependency}, leading to weak alignment between data and semantics~\cite{zhang2024survey}. To address this, we adopt \textit{network stochastic differential equations (Network SDEs)}~\cite{iafrate2024ergodic,liu2025graph} as the core generative mechanism. 

\paragraph{Network SDEs.}
Network SDEs enable fine-grained control over \emph{temporal dynamics} through node-wise drift and diffusion terms, and explicit modeling of \emph{spatial dependency} through structured coupling over a graph.
This makes them well-suited for simulating spatio-temporal processes with controllable dynamics and well-aligned semantics.
To model the spatio-temporal time series,
we assume that each discrete observation $x_{i,t}$ is sampled from an underlying
continuous-time latent process $X_i(t)$ associated with node $v_i \in \mathcal{V}$.
The latent dynamics over the graph $\mathcal{G} = (\mathcal{V}, \mathcal{E})$ are modeled with network SDEs.
For each node $i$, the state $X_i(t)$ evolves as
% \begin{equation}
% \label{eq:sde}
% \begin{aligned}[b]
% &dX_i(t)
% =
% f_i\!\left(X_i(t), t\right)\,dt
% +
% g_i\!\left(X_i(t), t\right)\,dW_i(t) \\
% & +
% \sum_{j \neq i}
% \lambda_i\, A_{ji}(t)
% \big(X_j(t-\tau_{ji}) - X_i(t)\big)\,dt  
% \end{aligned}
% \end{equation}
\begin{align}
\label{eq:sde}
&dX_i(t)
=
f_i(X_i(t),t)\,dt + g_i(X_i(t),t)\,dW_i(t) \notag \\
&+ \sum\nolimits_{j\ne i}\lambda_i A_{ji}(t)\bigl(X_j(t-\tau_{ji})-X_i(t)\bigr)\,dt
\end{align}
where $f_i$ denotes the node-specific drift function capturing intrinsic temporal dynamics,
$g_i$ controls stochastic diffusion, and $W_i(t)$ is a Wiener process.
The network coupling term models spatial dependencies induced by $\mathcal{G}$:
$A_{ji}(t)$ is a time-varying adjacency weight associated with the directed edge
$(v_j, v_i) \in \mathcal{E}$, $\lambda_i$ controls the coupling strength at node $i$,
and $\tau_{ji}$ represents the propagation delay from node $j$ to node $i$.
Discrete observations $x_{i,t}$ in the time series $T_i$ are obtained by sampling
$X_i(t)$ at the specified time points.

\paragraph{Network SDEs-based multi-agent spatio-temporal data synthesis pipeline.}
A key remaining challenge is how to align spatio-temporal dynamics with structured
textual descriptions. To address this issue, we propose a network SDEs-based multi-agent
data synthesis pipeline, illustrated in Figure~\ref{fig:data_synthesis}.
The framework consists of six agents.
The \textit{\textbf{Scenario Generation Agent}} produces high-level textual descriptions
of the spatio-temporal system.
The \textit{\textbf{Scenario Parsing Agent}} converts the text into structured,
machine-readable specifications, including nodes, edges, temporal patterns,
and spatial dependencies.
The \textit{\textbf{Judge Agent (Scenario Validation)}} evaluates scenario logic and
provides feedback to refine the outputs of the first two agents.
Based on the parsed specifications, the \textit{\textbf{SDEs Parameters Agent}}
instantiates the node-wise drift and diffusion functions $f_i(\cdot)$ and $g_i(\cdot)$,
together with coupling strengths $\lambda_i$, while the
\textit{\textbf{Time-Varying Adjacency Agent}} assigns dynamic adjacency weights
$A_{ji}(t)$ and propagation delays $\tau_{ji}$.
The \textit{\textbf{Judge Agent (Parameter Validation)}} assesses whether the simulated
data are consistent with the scenario description and provides feedback to refine
the SDE and adjacency parameters.
Finally, the \textit{\textbf{Simulation}} module generates spatio-temporal time series
by integrating Equation~\ref{eq:sde} with the instantiated parameters.

To ensure alignment between text and data, we design the network SDEs to be
expressive enough to faithfully realize the scenarios described by the first
two agents. Specifically, we distinguish demand source nodes from propagation
nodes to control where self-generated dynamics can occur, restrict exogenous
temporal patterns to demand sources, model spatial influence through
time-varying adjacency matrices, and introduce explicit time lags to capture
delayed propagation. These design choices enable the network SDEs formulation to
directly encode the semantic constraints specified in the textual scenario,
thereby enforcing consistency between textual descriptions, structured
specifications, and simulated spatio-temporal data.
As a result, the pipeline yields well-aligned textual descriptions and spatio-temporal data, which we use to generate spatio-temporal QA datasets. 
More details are provided in Appendix~\ref{app:Spatio-Temporal Data and Textual Description Pair Generation}.

\begin{figure*}[t!]
    \centering
    \includegraphics[width=0.98\textwidth]{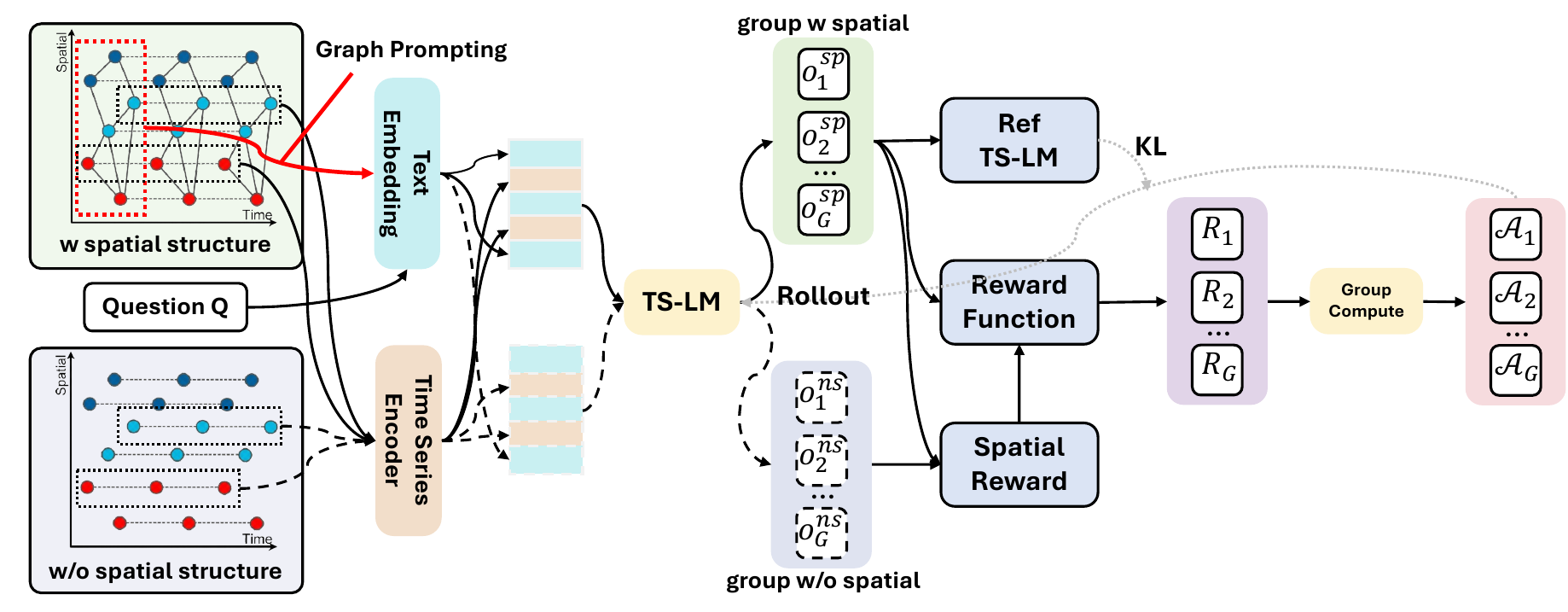}
    % \vskip -1em
    \caption{An illustration of our proposed STReasoner with the S-GRPO algorithm.
}
    \label{fig:streasoner}
    % \vskip -1em
\end{figure*}

\paragraph{Spatio-Temporal QA Generation.}
As shown in Figure~\ref{fig:data_synthesis}, we generate two spatio-temporal QA datasets: \emph{ST-Bench} and \emph{ST-Align}. ST-Bench focuses on training and evaluating complex spatio-temporal reasoning, while ST-Align supports alignment pretraining via basic understanding questions.

% \wei{please number these tasks} 
\underline{\emph{ST-Bench}} is a benchmark designed to assess spatio-temporal reasoning through four tasks:
\textit{\textbf{T1}:Etiological Spatial Reasoning} focuses on inferring global system-level
semantics from observed spatio-temporal dynamics.
\textit{\textbf{T2}:Spatial Entity Identification} examines whether a model can recognize
semantic roles of nodes from spatio-temporal observations.
\textit{\textbf{T3}:Spatial Correlation Reasoning} evaluates causal reasoning over spatial
structures, including direct causal influence and multi-hop propagation.
Finally, \textit{\textbf{T4}:In-context Spatio-Temporal Forecasting} assesses the ability to
predict future temporal evolution under spatial dependencies. Prompts for QA generation are provided in Appendix~\ref{app:dataset_prompt}.

\underline{\emph{ST-Align}} is designed to align textual representations with
spatio-temporal data during training through basic understanding questions:
\emph{temporal} questions focusing on node-level temporal dynamics,
\emph{spatial} questions probing graph connectivity and influence paths,
and \emph{spatio-temporal} questions that jointly reason over propagation
delays and time-varying spatial interactions.
Specifically, we generate QA
pairs using predefined templates (see Appendix~\ref{app:dataset_stalign_prompt}) based on the outputs of the \textit{SDEs Parameters Agent}, the \textit{Time-Varying Adjacency Agent}, and the simulated spatio-temporal data, 

\subsection{Dataset Splits and Usage}
We first synthesize 1,200 spatio-temporal samples using the \textit{Network SDE-based
Multi-Agent Spatio-Temporal Data Synthesis Pipeline} illustrated in Figure~\ref{fig:data_synthesis}. We provide a showcase of the synthesized data sample in Table~\ref{tab:showcase_data_sample}.
To ensure data reliability, all synthesized samples are manually inspected, as
described in Appendix~\ref{app:data_validation}, resulting in 1,064 retained samples.
We split the samples into 80\% for training and 20\% for testing.
We generate \textbf{ST-Align} using the training set, which contains 153,700 QA pairs. From the same training set, we further construct \textbf{ST-Bench}, and divide it in a 6:2 ratio to form \textbf{ST-SFT} and \textbf{ST-RL} for SFT and RL, respectively. The testing split is used exclusively to construct \textbf{ST-Test} for held-out evaluation on \textbf{ST-Bench}. More details can be found in Appendix~\ref{app:QA Dataset Generation}.

\section{Our Solution: STReasoner}
\subsection{Model Architecture}
As shown in Figure~\ref{fig:streasoner}, STReasoner processes multimodal inputs by
separately encoding time series data and textual inputs, including the natural
language query and the graph structure prompted in text.
Following prior work on time series encoding for LLMs~\cite{zhang2025time,chow2024towards,xie2024chatts,jin2023time},
we patchify the input time series~\cite{nie2022time} and encode each patch using a
lightweight five-layer MLP (Time Series Encoder).
The resulting time series embeddings are interleaved with text tokens according to node order to form a single input sequence for the LLM, thereby casting the model as a TS-LM.
Specifically, node-associated time series embeddings are inserted as special
placeholders in the textual stream, yielding an input of the form
$[\texttt{Node}_1: \langle \mathrm{TS}_1 \rangle,
 \texttt{Node}_2: \langle \mathrm{TS}_2 \rangle,
 \ldots,
 \texttt{Graph}, \texttt{Question}]$.
This representation preserves the contextual information of each time series.
We adopt value-preserving normalization following ChatTS~\cite{xie2024chatts} to
normalize time series features while preserving the original numerical information.

\subsection{Model Training}
\label{sec:model_training}
Training models for spatio-temporal reasoning in time series presents challenges
distinct from those in purely textual or visual domains.
First, the newly introduced time series encoder produces embeddings that are not
naturally aligned with the text embedding space of pretrained LLMs.
Second, pretrained LLMs lack inherent spatio-temporal priors, as they are only
minimally exposed to structured time series data during pretraining.
Third, models may rely on superficial temporal patterns and fail to consistently ground their reasoning in spatial structure, due to the absence of explicit training signals that encourage spatial reasoning.
To address these challenges, we adopt a three-stage training paradigm:
(1) large-scale alignment pretraining to bridge textual and time series modalities,
(2) injection of spatio-temporal reasoning priors to provide a reliable reasoning
cold start, and
(3) spatial-aware RL to refine reasoning behavior.

\paragraph{Stage 1: Large-Scale Alignment Pretraining.}
In the first stage, we perform large-scale alignment pretraining using \textbf{ST-Align}.
This dataset consists of QA pairs that probe fundamental
characteristics of spatio-temporal data, including temporal, spatial, and
spatio-temporal attributes.
The goal of this stage is to establish an initial alignment between textual
representations and time series embeddings within the STReasoner.
Leveraging large-scale synthetic data, the model learns to associate textual
descriptions with attributes encoded in spatio-temporal time series, providing
a foundation for subsequent reasoning-oriented training.

\paragraph{Stage 2: Spatio-Temporal Reasoning Priors Injection.}
To provide a reliable SFT cold start, we construct
chain-of-thought (CoT) annotations via rejection sampling~\cite{ahn2024large,jin2025search}.
Specifically, for each question in \textbf{ST-SFT}, we sample five candidate
responses using an instructed LLM (Claude-4.5-Sonnet) and retain only those that
lead to correct final answers.
The selected reasoning trajectories are aggregated to form the \textbf{ST-CoT}
dataset, which is then used to perform SFT.
This stage injects explicit spatio-temporal reasoning priors into the model,
establishing a strong initialization for subsequent RL.

\paragraph{Stage 3: Reinforcing Reasoning with Spatial-Aware GRPO.}
While GRPO~\cite{shao2024deepseekmath} has proven effective in text-based reasoning, it lacks explicit reward for spatial reasoning. To address this, we propose \textit{Spatial-Aware GRPO} (\textbf{S-GRPO}), which introduces a contrastive reward mechanism that explicitly encourages spatial reasoning, as illustrated in Figure~\ref{fig:streasoner}.
The core idea behind \textbf{S-GRPO} is to compare the model’s performance on the same question when spatial structure is provided or not. For each input question, we generate two groups of responses:
$\{o_i^{\text{sp}}\}_{i=1}^G$ with explicit spatial structure,
and $\{o_i^{\text{ns}}\}_{i=1}^G$ without spatial structure,
where $G$ denotes the number of responses per group, we then define the final reward $R_i$ used for calculating advantages as: 
\[
R_i =
\begin{cases}
r_i^{\text{sp}} + \alpha, & \text{if } r_i^{\text{sp}} > \beta\, r_i^{\text{ns}}, \\
r_i^{\text{sp}}, & \text{otherwise}.
\end{cases}
\]
where $r_i^{\text{sp}}$ and $r_i^{\text{ns}}$ denote the rewards of responses
$o_i^{\text{sp}}$ and $o_i^{\text{ns}}$, respectively, containing both the accuracy reward and the format reward (see Appendix~\ref{app:Task-Grounded Reward Design})~\cite{guo2025deepseek}, $\alpha$, $\beta$ are hyperparameters controlling the magnitude of the spatial reward and tolerance.
This contrastive design encourages the model to perform better when the spatial structure is presented than when it is removed. The model is only granted this positive reward if its current reasoning strategy for a given question demonstrates a reliance on spatial information.
Then, the advantage $\mathcal{A}_i$ is computed over the rewards within each group:
$
\mathcal{A}_i = \frac{R_i - \operatorname{mean}(\{R_j\})}{\operatorname{std}(\{R_j\})}.
$
Following DeepSeek R1~\cite{guo2025deepseek}, the final policy update is as follows:
% \begin{small}
% \begin{equation}
% \begin{aligned}
% &\mathcal{J}_{\text{S-GRPO}}(\theta)= 
% \mathbb{E}_{q,\{o_i\}} \Bigg[ \frac{1}{G}
% \sum_{i=1}^{G}
% \Bigg(
% \min \Bigg(
% \frac{\pi_\theta(o_i \mid q)}{\pi_{\theta_{\text{old}}}(o_i \mid q)} \mathcal{A}_i, \\
% &
% \operatorname{clip}\!\left(
% \frac{\pi_\theta(o_i \mid q)}{\pi_{\theta_{\text{old}}}(o_i \mid q)},
% 1 - \epsilon,
% 1 + \epsilon
% \right) \mathcal{A}_i
% \Bigg)  \;-\; \beta\, \mathbb{D}_{\mathrm{KL}}\!\left(\pi_\theta \,\|\, \pi_{\text{ref}}\right)\Bigg)
% \Bigg] \nonumber
% \end{aligned}
% \end{equation}
% \end{small}
\begin{equation}
\begin{aligned}
&\mathcal{J}_{\text{S-GRPO}}(\theta)= \\
&\mathbb{E}_{q,\{o_i\}} \Bigg[ \frac{1}{G}
\sum_{i=1}^{G}
\Bigg(
\min \Bigg(
\frac{\pi_\theta(o_i \mid q)}{\pi_{\theta_{\text{old}}}(o_i \mid q)} \mathcal{A}_i, \\
&
\operatorname{clip}\!\left(
\frac{\pi_\theta(o_i \mid q)}{\pi_{\theta_{\text{old}}}(o_i \mid q)},
1 - \epsilon,
1 + \epsilon
\right) \mathcal{A}_i
\Bigg) \\
& \;-\; \beta\, \mathbb{D}_{\mathrm{KL}}\!\left(\pi_\theta \,\|\, \pi_{\text{ref}}\right)\Bigg)
\Bigg]
\end{aligned}
\end{equation}

By explicitly comparing the model’s performance under inputs with and without spatial structure, \textbf{S-GRPO} introduces a contrastive training signal that drives the model to prefer reasoning strategies that leverage spatial patterns.

\section{Experiment}

\begin{table*}[!t]
% \vspace{-2mm}
\centering
\caption{Model comparison on four tasks, reporting \textbf{ACC} (\%) for \textbf{T1: Etiological}, \textbf{T2: Entity}, and \textbf{T3: Correlation} tasks, 
and \textbf{MAE} for \textbf{T4: Forecasting}. 
We also report \textbf{Tokens} (M) and \textbf{Est. Cost} (\$), where token usage refers to input token only, and cost is estimated based on publicly available API pricing.
\textbf{Prompting} indicates the input modality used by each model.
$^{\dagger}$ indicates that STReasoner outperforms the best baseline with a p-value less than 0.05.
The best performance is highlighted in \bestres{red}, and the second-best is \secondres{underlined}.}
\label{tab:main}
% \vskip -0.5em
\resizebox{\linewidth}{!}{
\renewcommand{\arraystretch}{1.25}

\setlength{\tabcolsep}{2mm}
\begin{tabular}{l l l cc cc cc cc c}
\toprule
\multirow{2}{*}{\textbf{Category}} &
\multirow{2}{*}{\textbf{Models}} &
\multirow{2}{*}{\textbf{Prompting}} &
\multicolumn{2}{c}{\textbf{T1: Etiological}} &
\multicolumn{2}{c}{\textbf{T2: Entity}} &
\multicolumn{2}{c}{\textbf{T3: Correlation}} &
\multicolumn{2}{c}{\textbf{T4: Forecasting}} &
\multirow{2}{*}{\textbf{Est. Cost}} \\

\cmidrule(lr){4-5}\cmidrule(lr){6-7}\cmidrule(lr){8-9}\cmidrule(lr){10-11}
& & &
\textbf{ACC} & \textbf{Tokens} &
\textbf{ACC} & \textbf{Tokens} &
\textbf{ACC} & \textbf{Tokens} &
\textbf{MAE} & \textbf{Tokens} \\
\midrule

\multirow{4}{*}{\textbf{Proprietary Models}}
& GPT-5.2 & Text
& 83.09 & 0.66 & 38.78 & 4.86 & 58.79 & 6.84 & 63.99 & 0.49 & 22.48 \\
& GPT-5.2 & Image
& 86.47 & 0.24 & 40.54 & 1.38 & 65.08 & 1.89 & 64.70 & 0.31 & 6.69 \\
& Claude-4.5-Sonnet & Text
& 78.64 & 0.81 & 41.93 & 5.67 & 77.87 & 8.05 & \secondres{63.74} & 0.74 & 45.80 \\
& Claude-4.5-Sonnet & Image
& 79.61 & 0.39 & 41.85 & 2.12 & 76.04 & 2.84 & \bestres{63.36} & 0.48 & 17.48 \\

\midrule
\multirow{6}{*}{\textbf{Open-Source Models}}
& Qwen3-8B & Text
& 21.26 & 0.98
& 5.28  & 7.31
& 5.53  & 10.26
& 94.03 & 0.72
& 3.85 \\
& \quad - SFT & Text
& 82.13 & 0.98 & 62.65 & 7.31 & 79.84 & 10.26 & 66.49 & 0.72 & 3.85\\
& \quad - SFT+S-GRPO & Text
& 89.37 & 0.98 & 65.41 & 7.31 & 81.34 & 10.26 & 66.35 & 0.72 & 3.85\\

& Qwen3-VL-8B-Instruct & Image
& 68.60 & 0.21
& 31.16 & 1.20
& 53.52 & 1.65
& 74.21 & 0.26
& 0.66 \\
& \quad - SFT & Image
& 90.82 & 0.21 & 61.31 & 1.20 & 80.78 & 1.65 & 68.53 & 0.26 & 0.66\\
& \quad - SFT+S-GRPO & Image
& \secondres{91.79} & 0.21 & \secondres{69.43} & 1.20 & \secondres{83.92} & 1.65 & 67.29 & 0.26 & 0.66\\

\midrule
\multirow{2}{*}{\textbf{Time Series LMs}}
& Time-MQA-7B & Text
& 19.32 & 0.98 & 14.24 & 7.31 & 17.90 & 10.26 & 84.70 & 0.72 & 3.85 \\
& ChatTS-8B & TS Encoder
& 56.52 & 0.08 & 19.51 & 0.50 & 41.08 & 0.72 & 85.14 & 0.08 & \secondres{0.27} \\

\midrule
\textbf{Time Series RMs}
& Time-R1-7B & Text
& 60.39 & 0.98 & 29.65 & 7.31 & 48.62 & 10.26 & 68.15 & 0.72 & 3.85 \\

\midrule
\rowcolor{gray!15}
\textbf{Spatio-Temporal RMs}
& STReasoner-8B (Ours) & TS Encoder
& \bestres{95.65}$^{\dagger}$ & 0.08
& \bestres{75.71}$^{\dagger}$ & 0.50
& \bestres{87.12}$^{\dagger}$ & 0.72
& 65.59 & 0.08
& \bestres{0.27} \\

\bottomrule
\end{tabular}
}
% \vskip -1em
\end{table*}

\paragraph{Baselines.}
We compare \textbf{STReasoner} with models from four categories.
(1) \textbf{Proprietary Models}: GPT-5.2~\cite{openai2025gpt5} and Claude-4.5-Sonnet~\cite{anthropic2025claude_sonnet}.
(2) \textbf{Open-Source Models}: Qwen3-8B and Qwen3-VL-8B-Instruct~\cite{yang2025qwen3}.
(3) \textbf{Time Series Language / Reasoning Models (LMs/RMs)}: Time-MQA-7B~\cite{kong2025time}, ChatTS-8B~\cite{xie2024chatts}, and Time-R1-7B~\cite{luo2025time}. We provide implementation details of STReasoner in Appendix~\ref{app:implementation}.

\begin{figure*}[t!]
    \centering
    \begin{subfigure}[b]{0.27\textwidth}
        \centering
        \includegraphics[width=\textwidth]{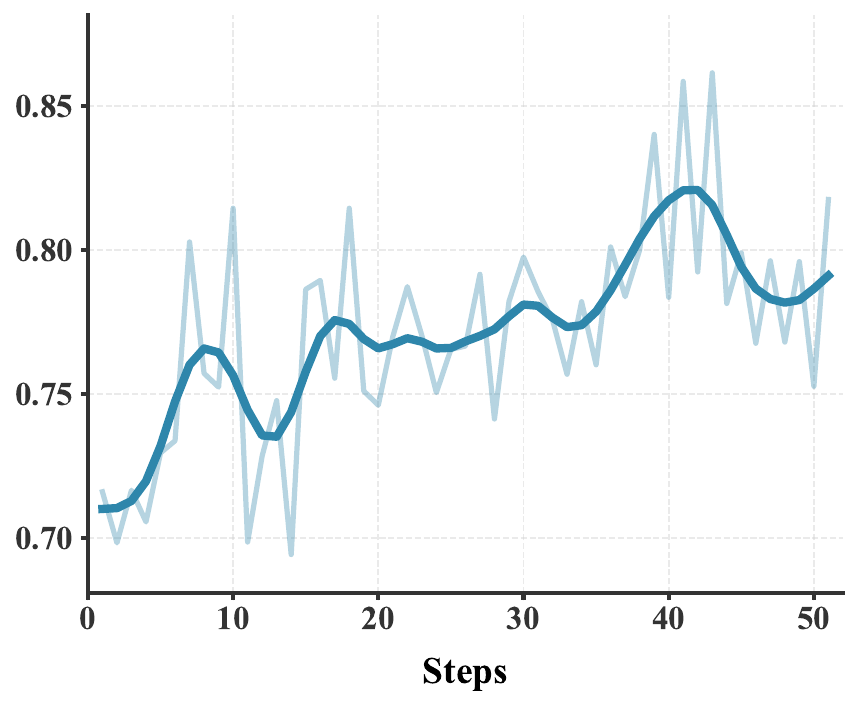}
        % \vskip -0.5em
        \caption{Accuracy Reward}
        \label{fig:accuracy}
    \end{subfigure}
    \hfill
    \begin{subfigure}[b]{0.27\textwidth}
        \centering
        \includegraphics[width=\textwidth]{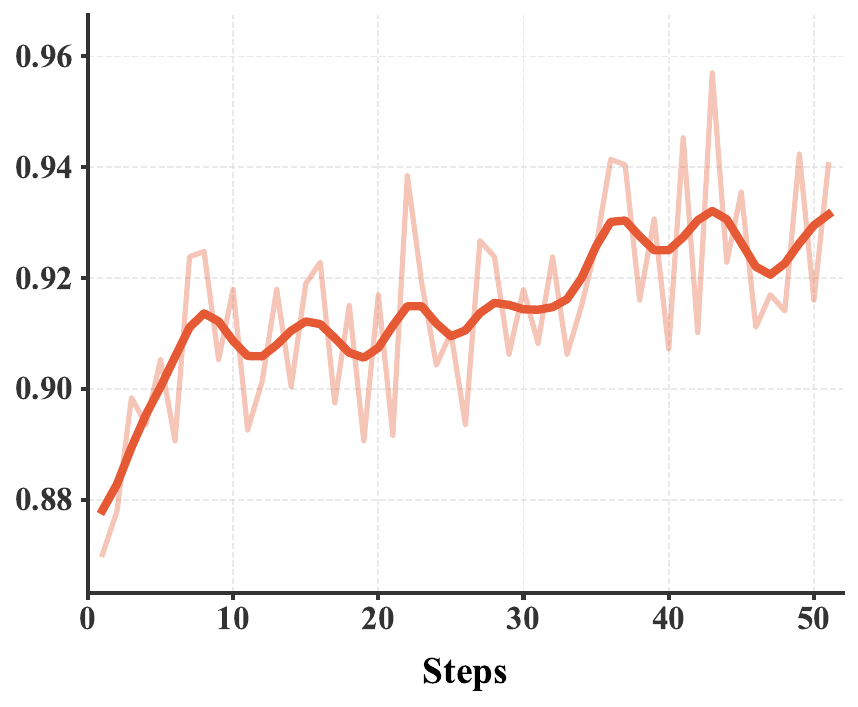}
        % \vskip -0.5em
        \caption{Spatial Reward}
        \label{fig:spatial}
    \end{subfigure}
    \hfill
    \begin{subfigure}[b]{0.27\textwidth}
        \centering
        \includegraphics[width=\textwidth]{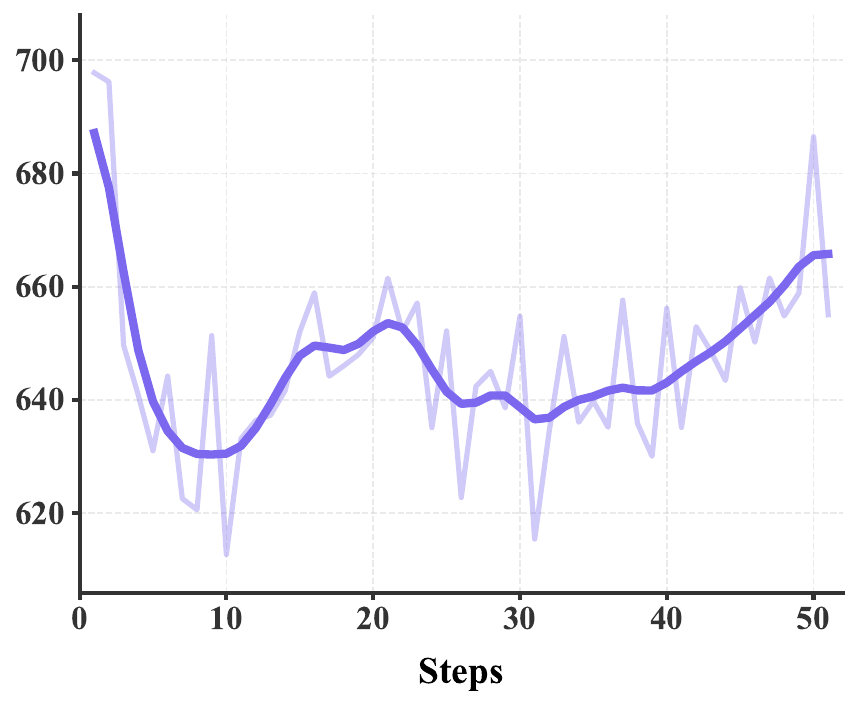}
        % \vskip -0.5em
        \caption{Response Length}
        \label{fig:response}
    \end{subfigure}
    % \vskip -0.5em
    \caption{RL training curves over steps.}
    \label{fig:training_curves}
    % \vskip -1em
\end{figure*}

\subsection{Main Results}
\paragraph{Performance of Existing LLMs.}
As shown in Table~\ref{tab:main}, proprietary models demonstrate strong zero-shot performance on all tasks, especially forecasting tasks. However, their large model sizes lead to substantially higher computational costs.
In contrast, most lightweight open-source models and time-series LMs/RMs perform significantly worse than proprietary models. Among them, time series RMs consistently outperform time series LMs, indicating that these tasks require explicit reasoning capability.
From a cost–performance perspective, prompting time series as images is generally more effective than text, as it achieves better performance while incurring lower token cost.

 % \vskip 0.15em 
\paragraph{Superior Performance.}
\textbf{STReasoner} achieves the best overall performance among open-source and time series models across all four tasks. 
Compared with proprietary models, \textbf{STReasoner} achieves these gains at an average cost of $\textbf{0.004}\times$ that of proprietary models, delivering average improvements of \textbf{17\%}, \textbf{135\%}, and \textbf{40\%} on the T1: Etiological, T2: Entity, and T3: Correlation tasks, respectively, while achieving comparable performance on T4: Forecasting with only \textbf{2.6\%} difference.
These results validate the effectiveness of the time series encoder and the proposed three-stage training strategy for spatio-temporal reasoning.

 % \vskip 0.15em 
\paragraph{Study of Time Series Modality.}
To study the effect of time series modality design and the use of a separate time series encoder, we apply the same training strategy, including SFT and S-GRPO, to models that prompt time series as text or images.
Image-based prompting performs well on tasks that rely on global shapes, such as Etiological, Entity, and Correlation reasoning, but underperforms text-based prompting on tasks that require precise numerical information, namely forecasting.
Overall, models using only text modality or image-based multimodality perform significantly worse than STReasoner.
This result shows that explicitly encoding time series through a dedicated encoder is necessary to preserve both global shape and numerical information for spatio-temporal reasoning.

\subsection{Zero-Shot Results on Real-World Data}
To evaluate the practical zero-shot performance of \textbf{STReasoner} on real-world spatio-temporal data, we construct a causal QA dataset based on CausalRivers~\cite{stein2025causalrivers}. 
Specifically, given the underlying causal graph, we define a causal effect from node $A$ to node $B$ if there exists a directed edge from $A$ to $B$, and generate one QA instance for each such edge by asking whether node $A$ has a causal effect on node $B$.
We evaluate \textbf{STReasoner} in a strict zero-shot setting, where no finetuning is performed on this dataset.
As shown in Table~\ref{tab:causal_results_single}, despite being trained only on synthetic data, \textbf{STReasoner} significantly outperforms two large proprietary models on this real-world benchmark.
These results indicate strong generalization ability and validate the effectiveness of our synthetic data construction and training strategy.

\begin{table}[tbp!]
\centering
\caption{Model comparison on the real-world dataset.}
% \vskip -0.5em
\label{tab:causal_results_single}
\resizebox{\columnwidth}{!}{
\begin{tabular}{l l c c c}
\toprule
\textbf{Model} & \textbf{Prompting} & \textbf{ACC} & \textbf{Tokens} & \textbf{Cost} \\
\midrule
GPT-5.2 & Text & 22.32 & 3.49M & \$6.15 \\
Claude-4.5-Sonnet & Text & 83.18 & 0.78M & \$2.35 \\
\midrule
STReasoner & TS Encoder & \textbf{98.82} & \textbf{0.26M} & \textbf{\$0.05} \\
\bottomrule
\end{tabular}
}
% \vskip -0.5em
\end{table}

\begin{table}[t!]
\centering
\caption{Ablation results of different training stages.}
% \vskip -0.5em
\label{tab:ablation}
% \renewcommand{\arraystretch}{1}
% \scalebox{0.9}{\setlength{\tabcolsep}{4pt}
\resizebox{\columnwidth}{!}{
\begin{tabular}{lcccc}
\toprule
\multirow{2}{*}{\textbf{Models}} & \textbf{Etiol.} & \textbf{Entity} & \textbf{Corr.} & \textbf{Fore.} \\
 & ACC & ACC & ACC & MAE \\
\midrule
% \rowcolor{gray!15}
\textbf{STReasoner} & \multirow{2}{*}{\textbf{95.65}} & \multirow{2}{*}{\textbf{75.71}}  & \multirow{2}{*}{\textbf{87.12}} & \multirow{2}{*}{\textbf{65.593}} \\
\footnotesize \texttt{Align+SFT+S-GRPO} &&&& \\
\midrule
\texttt{Align+SFT+GRPO} & 91.79 & 69.60  & 86.12 & 69.961 \\
\texttt{SFT+S-GRPO} & 91.30 & 67.76  & 83.98 & 69.014 \\
\texttt{Align+SFT} & 88.41 & 63.32  & 80.97 & 66.653 \\
\texttt{SFT} & 90.34 & 61.47  & 81.47 & 71.096 \\
% \texttt{Align+S-GRPO} & \\
\texttt{S-GRPO} & 47.34 & 23.20  & 39.20 & 91.921 \\
\texttt{Align} & 3.38 & 8.79  & 3.77 & 75.360 \\
\bottomrule
\end{tabular}
}
% \vskip -1em
\end{table}

% \vspace{-0.1em}
\subsection{Ablation Study}
% \vspace{-0.1em}
\paragraph{Does \texttt{Align} help?}
\texttt{Align} aims to align textual instructions with time series representations, providing a foundation for subsequent reasoning training.
As shown in Table~\ref{tab:ablation}, the model trained with \texttt{Align} alone performs poorly across all tasks.
This is because the \texttt{Align} objectives are not directly optimized for downstream reasoning tasks and therefore do not induce reasoning capability on their own.
Comparing \texttt{SFT+S-GRPO} with \textbf{STReasoner}, the addition of \texttt{Align} improves the final performance by an average of \textbf{6.3\%}.
This demonstrates that alignment provides complementary benefits when combined with \texttt{SFT} and \texttt{S-GRPO}.

\paragraph{Does \texttt{SFT} help?}
\texttt{SFT} plays a critical role in establishing instruction-following and basic reasoning ability.
Models that skip \texttt{SFT} and directly apply \texttt{S-GRPO} exhibit significantly worse performance, which is due to sparse and unstable reward signals during rollout under weak reasoning initialization.

\paragraph{Does \texttt{S-GRPO} help?}
After \texttt{SFT}, RL becomes effective.
Comparing \texttt{Align+SFT+GRPO} with \texttt{Align+SFT} shows that RL further improves accuracy.
Moreover, comparing \textbf{STReasoner} with \texttt{Align+SFT+GRPO} shows that replacing \texttt{GRPO} with the proposed \texttt{S-GRPO} yields an average improvement of \textbf{5.10\%}, demonstrating the effectiveness of spatial-aware RL for spatio-temporal reasoning.

\subsection{Effect of Model Scale}

Model scale is a key factor that can influence performance, making it important to distinguish improvements brought by the proposed method from those due to increased model capacity.
To ensure a fair comparison, all results in the main experiments are obtained using an 8B backbone.
We further conduct scaling experiments using Qwen3 as the backbone at 4B, 8B, and 14B, where all models are trained with \texttt{Align+SFT} under the same setting.

\begin{table}[t]
\centering
\caption{Scaling analysis across model sizes.}
\label{tab:scaling}
\resizebox{\columnwidth}{!}{
\begin{tabular}{lcccc}
\toprule
\textbf{Model} & \textbf{Etiol.} & \textbf{Entity} & \textbf{Corr.} & \textbf{Fore.} \\
 & ACC & ACC & ACC & MAE \\
\midrule
Qwen3-4B & 82.53 & 52.85 & 76.07 & 72.323 \\
STReasoner-4B & 90.34 & 64.66 & 83.29 & 70.321 \\
\midrule
Qwen3-8B & 88.41 & 63.32 & 80.97 & 66.653 \\
STReasoner-8B & 95.65 & 75.71 & 87.12 & 65.593 \\
\midrule
Qwen3-14B & 93.24 & 62.90 & 81.16 & 66.041 \\
STReasoner-14B & 96.69 & 76.60 & 87.61 & 65.478 \\
\bottomrule
\end{tabular}
}
\end{table}

As shown in Table~\ref{tab:scaling}, STReasoner consistently outperforms the corresponding Qwen3 baseline at all model scales.
While increasing the baseline size from 4B to 14B leads to moderate improvements, the gains are limited compared with those achieved by applying the proposed method.
Notably, STReasoner-8B already surpasses Qwen3-14B on multiple tasks, suggesting that the proposed method provides stronger benefits than simply scaling up the model.

\begin{figure}[t!]
    \centering
    \includegraphics[width=\linewidth]{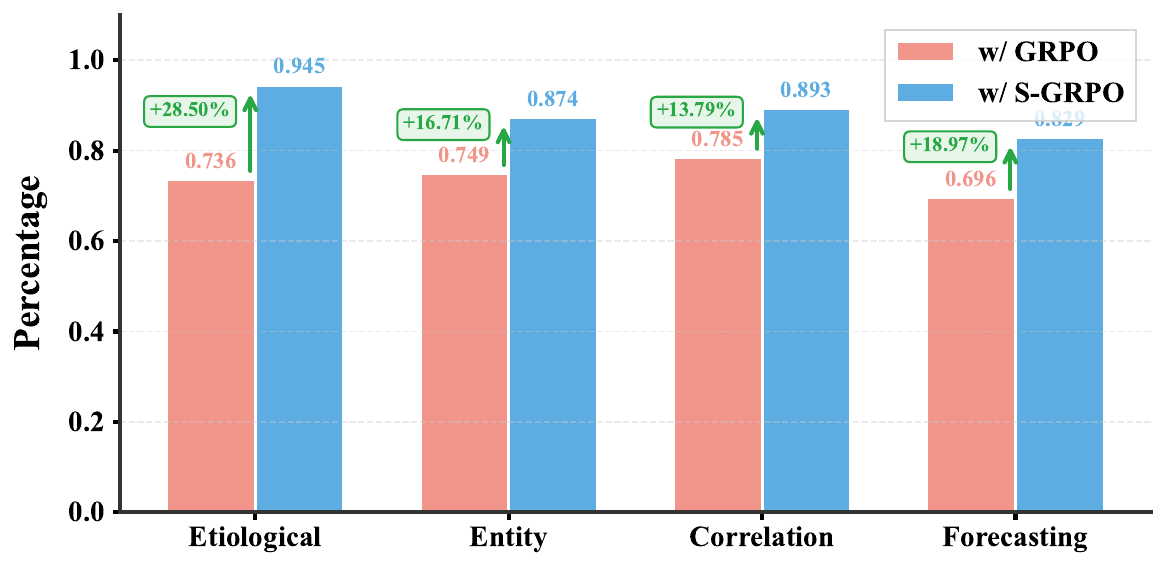}
    % \vskip -0.5em
    \caption{Percentage of spatial reasoning responses.}
    \label{fig:spatial_effect}
    % \vskip -1em
\end{figure}

% \vspace{-0.1em}
\subsection{Effect of Spatial Reward Analysis}
\paragraph{Training Curves of RL.}
Figure~\ref{fig:training_curves} illustrates the RL training dynamics of \textbf{STReasoner}. 
As shown in Figure~\ref{fig:accuracy}, the accuracy reward exhibits an overall upward trend, indicating that the model gradually improves its ability to generate correct answers during RL training. 
Similarly, Figure~\ref{fig:spatial} shows a steady increase in the spatial reward, suggesting that the model increasingly relies on spatially grounded reasoning behaviors under \textbf{S-GRPO}. 
Notably, the response length in Figure~\ref{fig:response} decreases in the early stage of RL training, followed by a gradual increase and stabilization. 
This pattern likely reflects a transition in the learned reasoning policy: the model initially departs from sub-optimal SFT reasoning patterns and eventually converges to a more stable and effective reasoning strategy.

\paragraph{Effect of S-GRPO on Spatial Reasoning Usage.}
To further assess whether \textbf{S-GRPO} encourages models to actively rely on spatial structure during reasoning, we measure the percentage of responses that explicitly use spatial information on questions requiring spatial reasoning by GPT-5.2. The evaluation prompt can be found in Appendix~\ref{app:spatial_effect_prompt}.
As shown in Figure~\ref{fig:spatial_effect}, \textbf{STReasoner} trained with \textbf{S-GRPO} consistently exhibits a substantially higher spatial reasoning usage ratio across all tasks compared with vanilla GRPO. 
These results indicate that \textbf{S-GRPO} does not merely improve final accuracy, but more importantly, shifts the model’s reasoning behavior toward spatially grounded strategies.

\paragraph{Effect of Spatial Information Usage During RL.}
To further examine whether \textbf{S-GRPO} improves the model’s ability to leverage spatial knowledge, we evaluate model checkpoints at different RL training steps under two input settings: w/ and w/o graph structure.
Specifically, we report accuracy on Task 1 at steps 1--51 and measure the performance gap $\Delta$ between the two settings.
As shown in Table~\ref{tab:spatial_usage_rl}, overall accuracy improves steadily during RL training.
More importantly, the performance gain from spatial input becomes increasingly significant.
The gap $\Delta$ grows from 0.5 at step 1 to 6.2 at step 51, indicating that the model progressively relies more on spatial structure.
This result provides direct evidence that \textbf{S-GRPO} enhances the model’s ability to utilize spatial information, rather than merely improving overall accuracy.

\begin{table}[t]
\centering
\caption{Effect of spatial input (w/ vs. w/o graph) during RL training.}
\label{tab:spatial_usage_rl}
\resizebox{\columnwidth}{!}{
\begin{tabular}{lcccccc}
\toprule
\textbf{Step} & 1 & 11 & 21 & 31 & 41 & 51 \\
\midrule
ACC (w/ graph) & 82.4 & 86.9 & 90.9 & 93.0 & 94.2 & 95.65 \\
ACC (w/o graph) & 81.9 & 83.9 & 85.7 & 87.1 & 88.6 & 89.4 \\
\midrule
$\Delta$ & 0.5 & 3.0 & 5.2 & 5.9 & 5.6 & 6.2 \\
\bottomrule
\end{tabular}
}
\end{table}

\begin{figure}[t!]
    \centering
    \includegraphics[width=\linewidth]{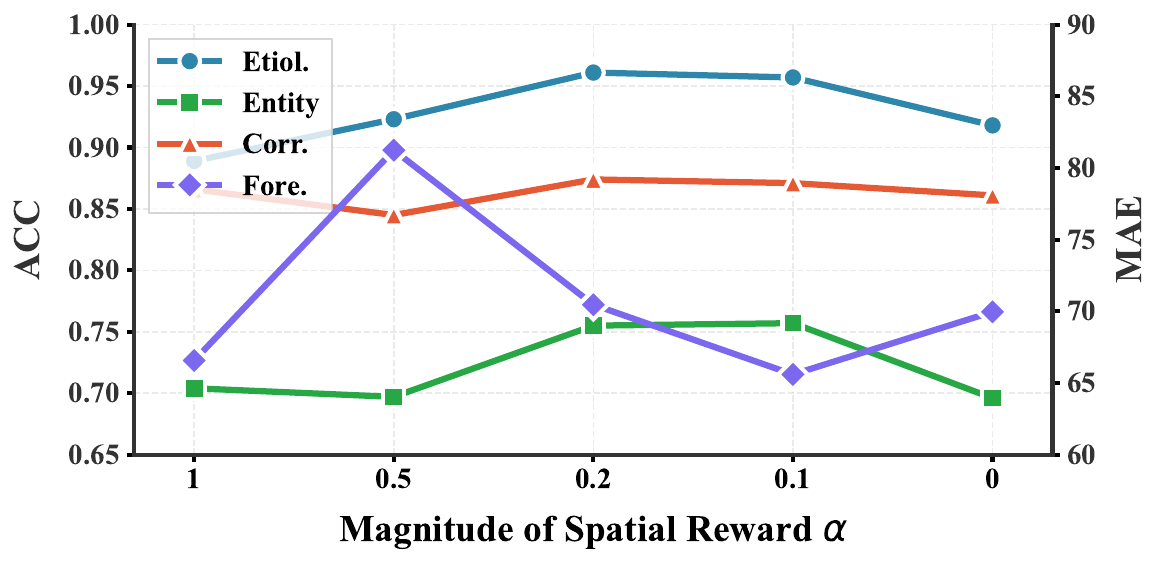}
    % \vskip -0.5em
    \caption{S-GRPO Sensitivity Analysis.}
    \label{fig:hyperparameters}
    % \vskip -1em
\end{figure}

\paragraph{Sensitivity Analysis.}
Figure~\ref{fig:hyperparameters} presents a sensitivity analysis on the spatial reward magnitude $\alpha$. 
Intermediate values ($\alpha=0.2$ and $0.1$) yield consistently strong performance, while extreme settings ($\alpha=1$, $0.5$, and $0$) degrade one or more metrics. 
A large $\alpha$ causes the spatial reward to dominate optimization and bias the model away from answer correctness, whereas removing $\alpha$ fails to encourage spatial-aware reasoning.

% \vspace{-0.1em}
\subsection{Case Study}
% \vspace{-0.1em}
\label{sec:case_study}
We present several case studies for each task in Appendix~\ref{app:case_study}. 
As shown in Table~\ref{tab:etiological_example}, STReasoner first parses the graph structure and, with the help of the time series encoder, accurately identifies key values and their corresponding time steps at each node. 
By aligning these temporal patterns with the directed topology, the model detects transport lags and attenuation along the propagation path, and finally infers the etiological scenario that best matches the observed spatio-temporal dynamics.

% \vspace{-0.2em}
\section{Conclusion}
% \vspace{-0.2em}
Spatio-temporal reasoning in time series is essential for real-world decision-making, but progress is limited by the lack of paired spatio-temporal data and text. We formally define spatio-temporal reasoning and propose an SDE-based data synthesis method that produces diverse spatio-temporal data with precise descriptions. We also build ST-Bench to evaluate multiple tasks. We propose STReasoner, the first TS-LM for complex spatio-temporal reasoning, trained in three stages on synthetic data, and introduce S-GRPO to improve spatial reasoning. Experiments show clear gains, indicating that our method effectively links spatio-temporal data with natural language understanding.

\clearpage
\section{Limitations}
Due to limited prior work on spatio-temporal reasoning in time series, STReasoner, while effective, still has several limitations.
\textbf{First}, although our experiments show that carefully designed synthetic data can support effective alignment and spatio-temporal reasoning, broader and more diverse real-world datasets are still essential for fully assessing and improving STReasoner.
\textbf{Second}, while a simple MLP-based time series encoder is sufficient for the structured signals considered in this work, more expressive architectures and multimodal integration strategies may be required for complex real-world scenarios.

\section{Potential Risks}
\begin{table}[h]
\centering
\caption{Energy consumption and CO$_2$ emissions by training stages.}
\label{tab:energy}
\resizebox{\columnwidth}{!}{
\begin{tabular}{lccccc}
\toprule
Stage & Run Time (h) & GPU-hours & GPU Energy (kWh) & Total Energy (kWh, incl. PUE) & CO$_2$e (kg) \\
\midrule
Align & 13.41 & 107.27 & 37.54 & 45.05 & 15.14 \\
SFT   & 8.18  & 65.44  & 22.90 & 27.48 & 9.23  \\
RL    & 5.38  & 43.07  & 15.07 & 18.09 & 6.08  \\
\midrule
Total & 26.97 & 215.77 & 75.52 & 90.62 & 30.45 \\
\bottomrule
\end{tabular}
}
\end{table}
A potential risk of this work is its environmental impact. Training and deploying large-scale models can incur substantial computational cost and energy consumption, which may contribute to increased carbon emissions. While our approach relies on synthetic data and relatively lightweight components, future extensions to larger models or datasets should carefully consider efficiency and sustainability.

\bibliography{acl}

\clearpage
\appendix

\section{Related Work}
\label{app:related_work}
\subsection{LMs Reasoning}
Improving LMs’ ability to perform complex and multi-step reasoning has become a major focus of recent research~\cite{wei2022chain, kojima2022large, yao2023tree, lightman2023let, besta2024graph,gong2025beyond}. 
DeepSeek-R1~\cite{guo2025deepseek} shows that large-scale RL with result-only rewards can induce self-emergent reasoning behaviors in LMs, and several subsequent studies have attempted to extend this paradigm to other modalities and tasks~\cite{jin2025search, huang2025vision, xie2025logic, wang2025ragen, feng2025video,liu2025time,wang2026position,lei2025mixture,liu2025hypervectors,glenn2024blendsql,wang2023accurate,guan2024intelligent}. 
Despite these advances, LM-based reasoning for spatio-temporal time series remains underexplored.

\subsection{Time Series Reasoning}

Time series language models leverage the implicit knowledge within LMs and contextual textual information to support downstream tasks and simple time series QA, typically without explicit reasoning~\cite{jin2023time,liu2024time,xie2024chatts,chow2024towards,cai2024timeseriesexam,merrill2024language,liang2024foundation,kong2025time,yang2025time,zhang2025tempogpt,wang2025chattime,chen2025mtbench,wang2025itformer,langer2025opentslm,zhou2025cats,zhang2025time,niu2025promedts,liu2025picture,liu2025can}. 
In contrast, time series reasoning models aim to develop genuine reasoning capabilities rather than pattern matching, often employing RL to encourage explicit multi-step reasoning and to produce interpretable, step-by-step explanations~\cite{wang2025can,tan2025inferring,luo2025time,guan2025timeomni,zhang2025timemaster,ning2025towards}. 
However, existing studies in this line of research mainly focus on univariate or multivariate time series and largely overlook spatio-temporal reasoning over spatially structured time series data.

\subsection{Spatio-Temporal Reasoning}
Spatio-temporal reasoning has been studied in data modalities such as videos~\cite{li2024llava,zhang2025open3dvqa,feng2025video,cheng2025v,wang2025time}, sensor-driven or event-based settings~\cite{quan2025benchmarking}, and trajectories~\cite{li2025stbench,feng2025citybench,feng2025urbanllava}. 
In contrast, spatio-temporal reasoning in time series involves reasoning over \emph{numerical signals associated with discrete entities}, where spatial relations are explicitly defined by a graph and temporal dependencies arise through time-lagged interactions, rather than spatial relationships being implicitly encoded in visual layouts or physical coordinates.

The closest prior work, STReason~\cite{hettige2025modular}, introduces a modular, training-free framework that decomposes spatio-temporal queries into programs executed by external tools, but it relies on textual encoding of time series, does not explicitly model or optimize graph structure, and mainly targets tasks such as anomaly detection or prediction that require limited reasoning. In contrast, our work trains a unified model that directly integrates time series, graph structure, and text, introduces spatial-aware reinforcement learning to enforce genuine spatio-temporal reasoning, and proposes ST-Bench to systematically evaluate spatio-temporal reasoning across multiple complementary tasks.

\subsection{Spatio-Temporal Forecasting.}
A large body of work studies spatio-temporal modeling for numeric prediction in general domains~\cite{chen2025learning,gao2025different,liu2024time,liu2026timer,wang2025accuracy} or specific application domains, such as traffic~\cite{yu2017spatio,wu2019graph}, air quality~\cite{liang2023airformer}, and epidemiology~\cite{liu2025cape,lau2026toward}.
The key difference between spatio-temporal forecasting and multivariate time series forecasting~\cite{ni2025u,ni2025timedistill,Fang2026,shi2024time} is the explicitly usage of the known graph structure.
These spatio-temporal forecasting models are carefully designed to capture spatial dependencies and temporal dynamics for forecasting accuracy, but they operate purely on numerical inputs and outputs.
As a result, they do not produce interpretable, long-form answers or support multi-step reasoning expressed in natural language.
They cannot function as meaningful baselines for evaluating language-based, multi-task spatio-temporal reasoning.

\subsection{Spatio-Temporal Language Models.}
Several recent works integrate LMs with spatio-temporal data~\cite{liang2025foundation}, including GPT-ST~\cite{li2023gpt}, UrbanGPT~\cite{li2024urbangpt}, UrbanLLM~\cite{jiang2024urbanllm},  UniST~\cite{yuan2024unist}, STG-LLM~\cite{liu2024can}, ST-LLM~\cite{liu2024spatial}, ST-LLM+~\cite{liu2025st} and ST-Vision-LLM~\cite{yang2025vision}.
Although these methods incorporate language models, their primary objective remains numeric forecasting or short-form question answering.
They generally lack explicit mechanisms for structured, multi-step reasoning and do not generate coherent, long-form explanations grounded in graph structure and temporal dynamics.
The closest prior work, UrbanLLM~\cite{jiang2024urbanllm}, targets urban question answering, but its responses are typically brief and surface-level, without sustained reasoning chains.
Therefore, these models are not directly comparable to our setting, which focuses on unified, language-based spatio-temporal reasoning across diverse tasks.

% \wei{this is a bit vague -- you should explain how their "spatialtemporal" is specifically different from our case instead of one simple comment "we are studying a different data modality"}

\section{Dataset Details}

\subsection{Spatio-Temporal Data and Textual Description Pair Generation}
\label{app:Spatio-Temporal Data and Textual Description Pair Generation}

As shown in Figure~\ref{fig:data_synthesis}, we first use the proposed network SDE--based multi-agent spatio-temporal data synthesis pipeline (detailed in Algorithm~\ref{alg:data_synthesis}) to generate aligned spatio-temporal data (that is, time series and graphs) together with aligned textual descriptions. The prompt for each agent is provided in Appendix~\ref{app:agent_prompt}. We use Claude-4.5-Sonnet~\cite{anthropic2025claude_sonnet} as the default LLM agent. In total, we generate \textbf{1200 pairs} of spatio-temporal data and textual descriptions.
Specifically, we generate spatio-temporal data with \textbf{3, 5, and 10 nodes}, covering \textbf{10 key domains}: ``Transportation,'' ``Energy,'' ``Environment\&Pollution,'' ``Ecology,'' ``Public Health,'' ``Hydrology,'' ``Oceanography,'' ``Agriculture,'' ``Mobility,'' and ``Climate.''

To better capture realistic spatio-temporal dynamics, our data synthesis pipeline incorporates three key design choices:

\paragraph{1. Demand Source and Propagation Nodes.}
We divide nodes into demand source nodes and propagation nodes. Concretely, we force the drift function $f_i$ of demand source nodes to be sinusoidal or mean-reverting, and assign only the mean-reverting drift function $f_i$ to propagation nodes. The drift function of demand source nodes could be time-varying. We also tried more diverse assignments of drift functions, but the results were not good. With this design, trends from demand source nodes propagate through the graph to other propagation nodes, while propagation nodes themselves only have mean-reverting trends and therefore are mainly influenced by their neighbors. This setup simulates real-world scenarios. For example, in a traffic setting with three nodes representing a residential area, a highway, and a commercial area, the residential and commercial areas are demand source nodes because traffic naturally originates there (for instance, people driving out of garages), while the highway is a propagation node whose traffic mainly comes from other regions and does not originate locally.

\paragraph{2. Time-Varying Adjacency Matrix.}
To simulate different spatial dependencies at different times, we set the graph adjacency matrix to be time-varying. Using the same example, in the morning traffic should mainly flow from residential areas to roads and then to commercial areas, while in the evening the direction is reversed. Therefore, the edge weights in the adjacency matrix should change over time, such as increasing the weight of residential-to-road edges in the morning and decreasing them at night, and doing the opposite for road-to-residential edges.

\paragraph{3. Propagation Time Lags.}
In addition, we introduce time lags to simulate propagation delays, since interactions between nodes are not instantaneous. Each edge is assigned a time lag to reflect the speed of propagation.

Through these three components, we expect our pipeline to simulate spatio-temporal data with dynamics that are close to those observed in real-world systems.

\subsection{Spatio-Temporal Data and Textual Description Pair Validation}
\label{app:data_validation}

\begin{figure}[htbp!]
    \centering
    \includegraphics[width=\linewidth]{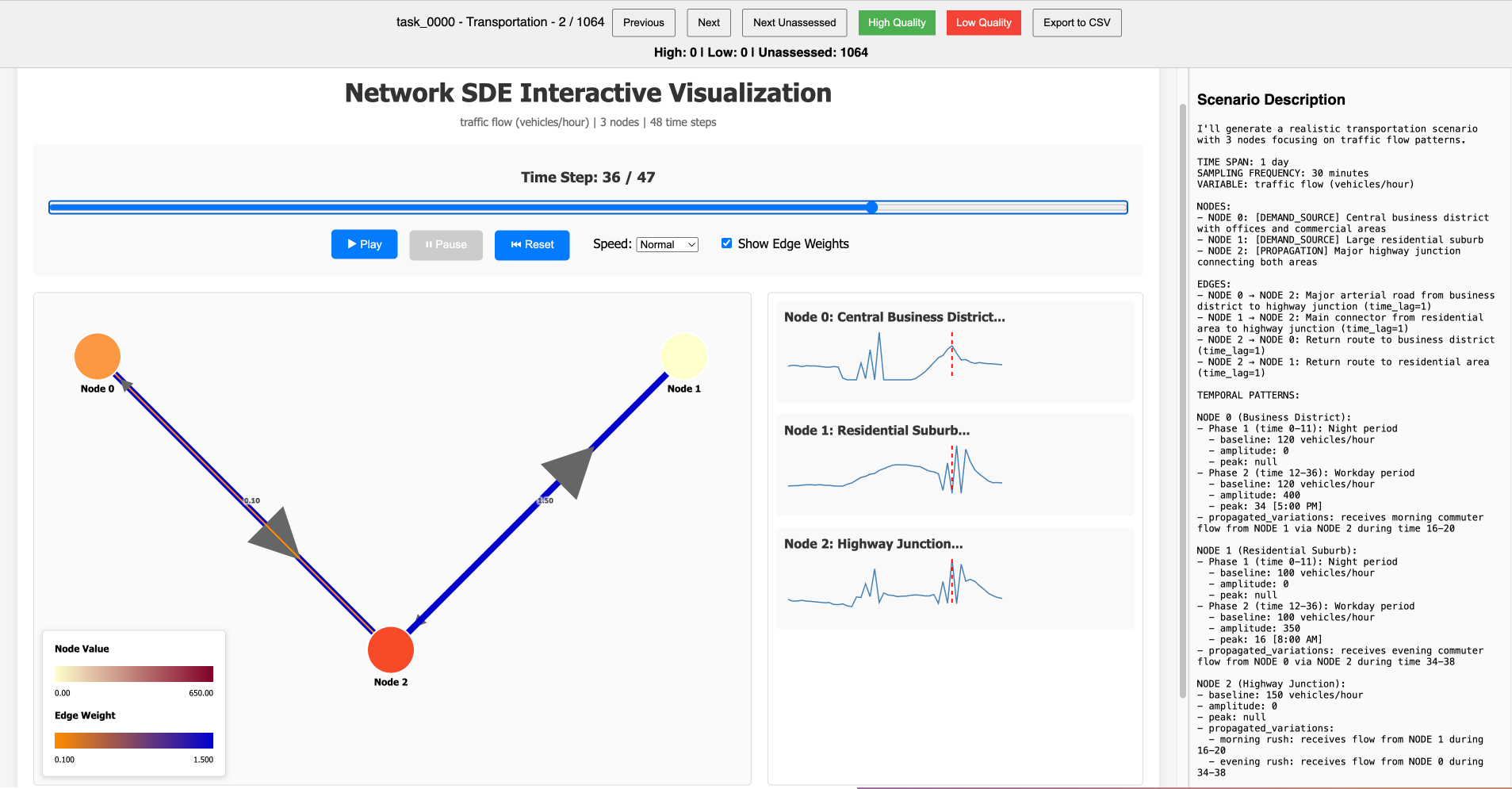}
    \caption{Screenshot of the human evaluation interface.}
    \label{fig:interface}
\end{figure}

To validate the usefulness of the generated data pairs and assess whether the textual descriptions are accurately aligned with the corresponding spatio-temporal data, we build a human evaluation interface, shown in Figure~\ref{fig:interface}. We manually inspect each pair by jointly examining the spatio-temporal data and its textual description, and then assign a quality label based on alignment correctness and overall coherence.

In total, 1064 spatio-temporal data and textual description pairs are labeled as ``High Quality,'' while 136 pairs are labeled as ``Low Quality,'' indicating that the majority of generated samples meet the desired quality criteria. We retain only the 1064 samples labeled as ``High Quality.''

\begin{figure}[htbp!]
    \centering
    \includegraphics[width=\linewidth]{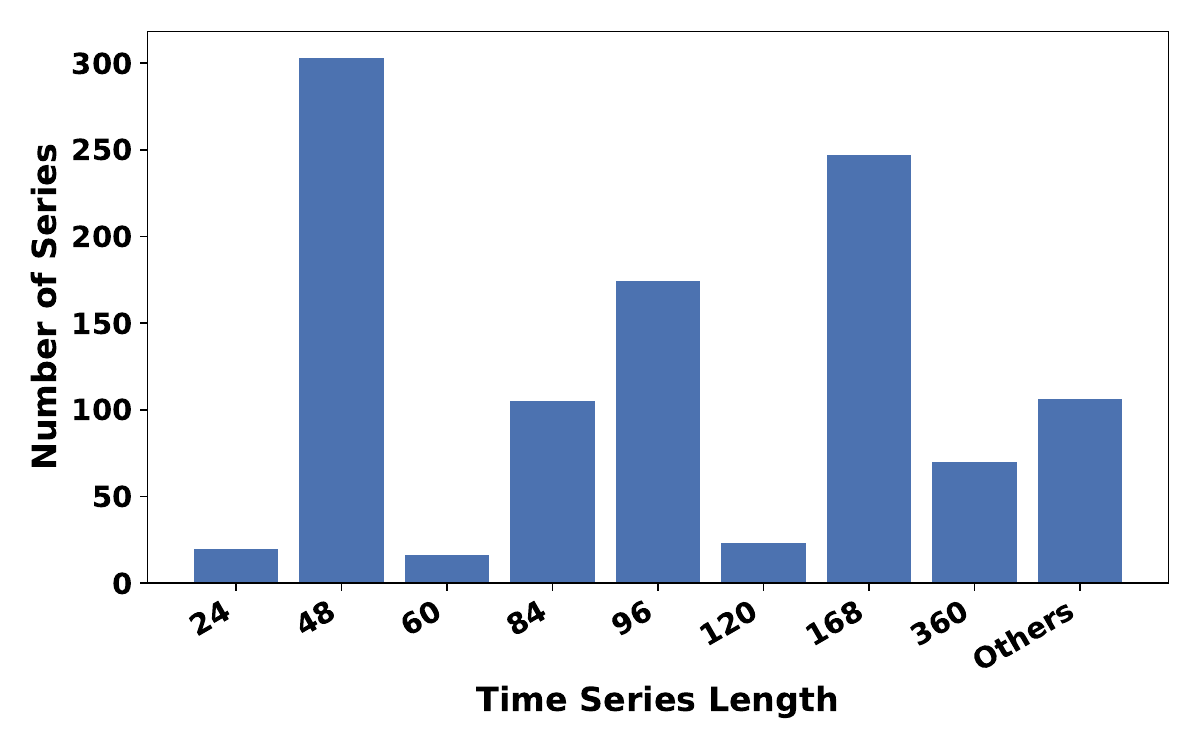}
    \caption{Distribution of Time Series Lengths.}
    \label{fig:time_series_length_distribution}
\end{figure}
\begin{figure}[htbp!]
    \centering
    \includegraphics[width=\linewidth]{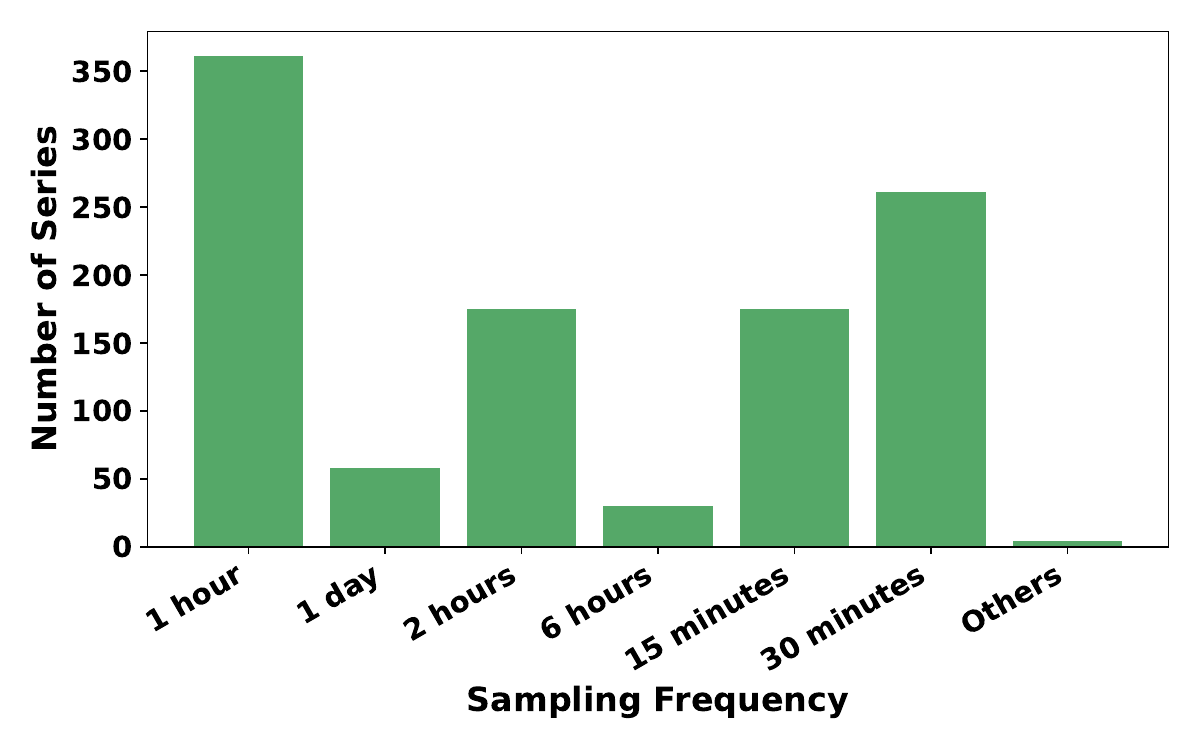}
    \caption{Distribution of Sampling Frequencies.}
    \label{fig:sampling_frequency_distribution}
\end{figure}
\begin{figure}[htbp!]
    \centering
    \includegraphics[width=\linewidth]{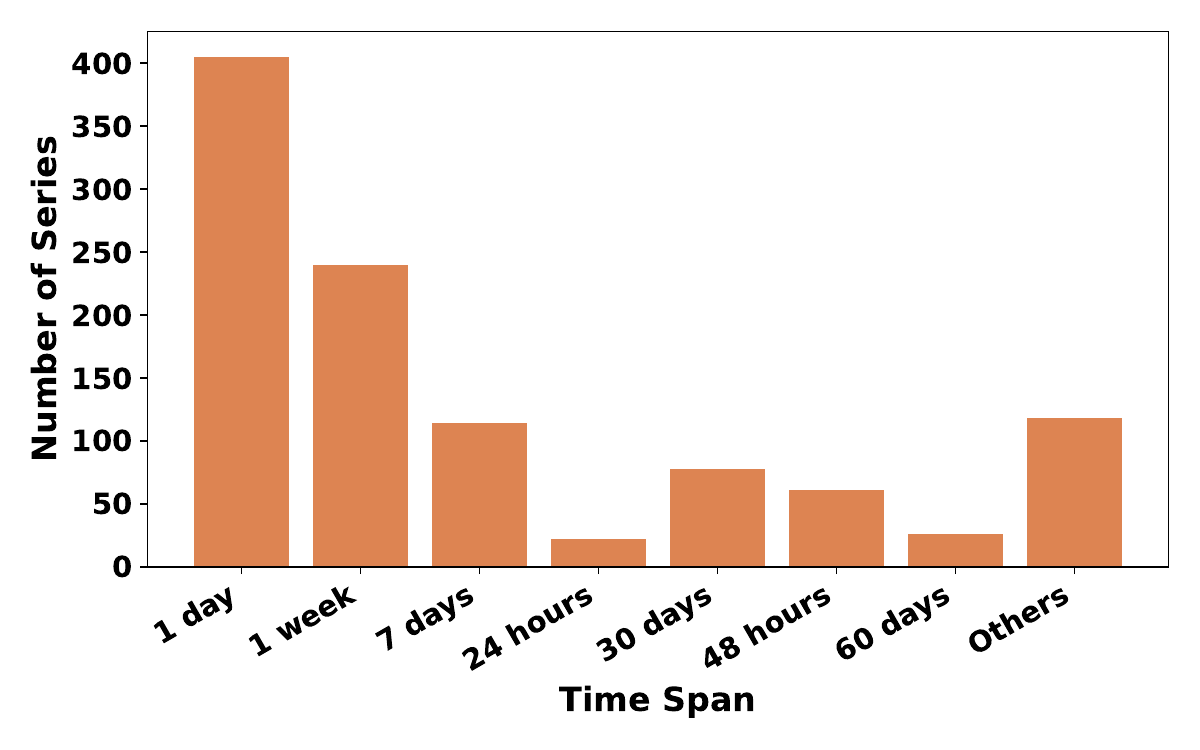}
    \caption{Distribution of Time Spans.}
    \label{fig:time_span_distribution}
\end{figure}

We further summarize the statistical characteristics of the selected 1064 samples in Figures~\ref{fig:time_series_length_distribution}–\ref{fig:time_span_distribution}.
Figure~\ref{fig:time_series_length_distribution} shows the distribution of time series lengths, covering both short sequences (tens of steps) and long sequences spanning thousands of steps. Figure~\ref{fig:sampling_frequency_distribution} illustrates the sampling frequency distribution, which ranges from high-frequency measurements (15–30 minutes) to low-frequency observations (daily). Figure~\ref{fig:time_span_distribution} reports the temporal coverage of the data, including samples that span from hours and days to multiple months.
In addition, the distributions of 10 domains and the number of nodes per graph (3,5,10 nodes) are approximately uniform, avoiding over-concentration on specific scenarios or graph sizes. Overall, these statistics indicate that the dataset exhibits broad coverage and high diversity along multiple dimensions, providing a challenging and representative testbed for spatio-temporal reasoning.

\subsection{QA Dataset Generation}
\label{app:QA Dataset Generation}
Based on the selected high-quality 1064 spatio-temporal data and aligned textual description pairs, we use Claude-4.5-Sonnet to generate two types of QA datasets: reasoning QA (ST-Bench) and alignment QA (ST-Align).

\begin{table}[t]
\centering
\caption{Data statistics of ST-Bench for each task.}
\label{tab:stbench_statistics}
\renewcommand{\arraystretch}{1.2}
\setlength{\tabcolsep}{5pt}
\resizebox{\columnwidth}{!}{
\begin{tabular}{l c c c c}
\toprule
\textbf{Task} & \textbf{ST-CoT} & \textbf{ST-RL} & \textbf{ST-Test} & \textbf{Total} \\
\midrule
T1:Etiological   & 1,576  & 405  & 207  & 2,188  \\
T2:Entity        & 4,974  & 2,451 & 1,194 & 8,619  \\
T3:Correlation   & 11,124 & 3,231 & 1,592 & 15,947 \\
T4:Forecasting   & 650   & 506  & 280  & 1,436  \\
\midrule
Total            & 18,324 & 6,593 & 3,273 & 28,190 \\
\bottomrule
\end{tabular}
}
\end{table}

ST-Bench~\footnote{CC-BY 4.0} is designed for Stage~2 and Stage~3 training and for evaluating the model’s spatio-temporal reasoning ability. We split ST-Bench into three subsets with a 6:2:2 ratio: ST-CoT for supervised fine-tuning (SFT), ST-RL for reinforcement learning (RL), and ST-Test for evaluation. The prompt templates used to generate these QA pairs are provided in Appendix~\ref{app:dataset_prompt}. Dataset statistics for each task in ST-Bench are reported in Table~\ref{tab:stbench_statistics}.

ST-Align is used in Stage~1 alignment training to align time series embeddings with graph structure and textual representations. We first split all spatio-temporal samples with an 8:2 ratio and use only the training split to generate alignment QA. The corresponding prompts are provided in Appendix~\ref{app:dataset_stalign_prompt}. ST-Align consists of three categories of questions: Temporal Characters, Spatial Characters, and Spatio-Temporal Characters.
For \textbf{temporal characters}, each node may contain multiple drift patterns. For each pattern, we generate one QA instance for every matched template key. If the drift type is sinusoidal, we additionally generate three QA instances that query the amplitude $A$, frequency $\omega$, and phase $\phi$, respectively.
For \textbf{spatial characters}, we generate two questions for each ordered node pair $(\text{src}, \text{tgt})$: (1) \texttt{edge\_relationship}, which asks whether a direct edge exists from \text{src} to \text{tgt}, and (2) \texttt{indirect\_connection}, which asks whether an indirect path exists between them. This results in a total of $2 \times N^2$ QA instances, where $N$ is the number of nodes.
For \textbf{spatio-temporal characters}, the total number of questions is $N + E + 2M$, where $N$ is the number of nodes (one question per node querying its type), $E$ is the number of edges (one question per edge querying its propagation delay), and $M$ is the number of time-varying edge modulations. For each modulation, we generate two questions that query the modulation multiplier and the effective coupling strength.

\section{Task-Grounded Reward Design}
\label{app:Task-Grounded Reward Design}
For each response, we compute a format reward $r_{\text{format}}$ to enforce a strict output schema:
\[
r_{\text{format}}
= \mathbb{I}\!\left(\text{valid output format}\right).
\]
where a valid output format requires the response to follow the schema
\texttt{<think>...</think><answer>...</answer>}.
The task reward $r_{\text{task}}$ is defined according to the type of ground-truth answer.
If the ground truth $y$ is a single discrete label (e.g., multiple-choice), we assign
\[
r_{\text{task}} =
\begin{cases}
1, & \hat{y} = y,\\
0, & \text{otherwise}.
\end{cases}
\]
$\hat{y}$ is the choice from the answer field. If the ground truth is a numeric sequence $y = \{y_t\}_{t=1}^{T}$ (e.g. T4:Forecasting), we first extract all predicted numbers
$\hat{y} = \{\hat{y}_t\}$ from the \texttt{<answer>} field. If no valid numbers are produced,
$r_{\text{task}} = 0$.
Otherwise, predictions are padded or truncated to length $T$, and we compute a relative-error–based score:
\[
r_{\text{task}} = \frac{1}{T} \sum_{t=1}^{T} 1 - \min\!\left(1,\;
\frac{|\hat{y}_t - y_t| + \epsilon}{|y_t| + \epsilon}
\right),
\]
where $\epsilon = 10^{-9}$.
If the predicted sequence length exactly matches the ground truth, we add a small bonus of $0.1$ and clip
the final score to $[0,1]$.

Finally, the per-sample reward is a weighted combination of format and task rewards:
\[
r
= (1 - \lambda)\,r_{\text{task}}
+ \lambda\,r_{\text{format}},
\]
where $\lambda = 0.5$ in all experiments.

\section{Implementation Details}
\label{app:implementation}
We use Qwen3-8B~\cite{yang2025qwen3} as the base model of \textbf{STReasoner}. All experiments are conducted using 8 NVIDIA A100 (80GB) GPUs.
Stage~1 (Alignment) is trained on ST-Align for 1{,}000 steps, and Stage~2 (SFT) is trained on ST-CoT for 400 steps.
Stages~1 and~2 are implemented using LlamaFactory~\cite{zheng2024llamafactory}.
We use a cosine learning rate scheduler with a warm-up ratio of 0.2.
The \texttt{per\_device\_train\_batch\_size} is set to 2, and \texttt{gradient\_accumulation\_steps} is set to 32.
The learning rate for both the Qwen3-8B and the separate time series encoder is set to $1\times10^{-5}$.
Stage~3 (RL) is trained on ST-RL for one epoch and is implemented using EasyR1~\cite{zheng2025easyr1} and \texttt{verl}~\cite{sheng2025hybridflow}.
The rollout group size $G$ is set to 8.
The spatial reward hyperparameters are set to $\alpha=0.1$ and $\beta=0.8$.
We use a linear learning rate scheduler with a warm-up ratio of 0.2.
The \texttt{rollout\_batch\_size} is set to 128, and the learning rate is set to $1\times10^{-7}$.

\begin{figure}[tbp!]
    \centering
    \includegraphics[width=\linewidth]{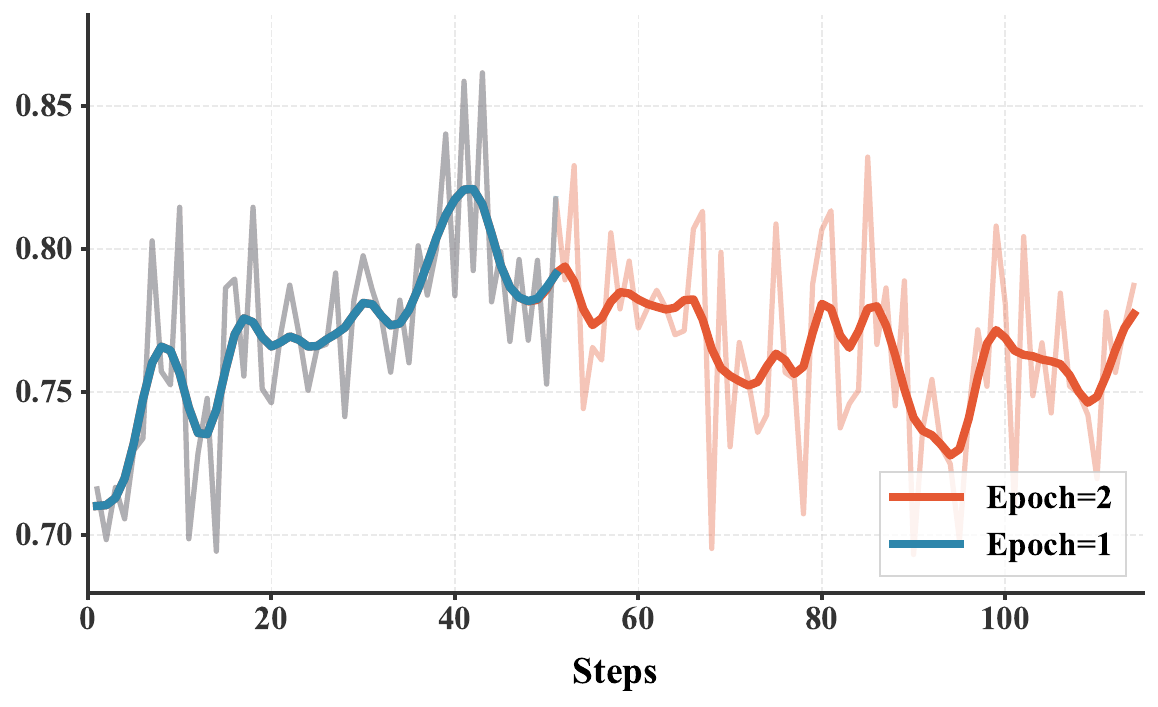}
    \caption{Scaling Up RL Training from 1 epoch to 2 epochs.}
    \label{fig:scaling_up_rl}
\end{figure}

\section{Performance with Confidence Intervals}

We report the results as mean $\pm$ 95\% confidence intervals over five runs.
The T1:Etiological achieves an accuracy of $95.39 \pm 0.74$, the T2:Entity reaches $75.78 \pm 0.49$, and the T3:Correlation attains $87.20 \pm 0.47$ accuracy.
For the T4:Forecasting, the model obtains a mean absolute error of $65.61 \pm 0.06$.

\section{Scaling Up RL Training}

Figure~\ref{fig:scaling_up_rl} shows that extending RL training from 1 to 2 epochs yields further improvements in the accuracy reward at early stages, but the gains gradually saturate and become less stable.
This suggests diminishing returns from prolonged RL training under the same reward setting.

\onecolumn

\section{Synthesized Data Pair Showcase}

\begin{table*}[ht!]
  \captionsetup{font=small,aboveskip=3pt,belowskip=4pt}
  \centering
  \newcolumntype{L}[1]{>{\raggedright\arraybackslash}p{#1}}
  \newcolumntype{Y}{>{\raggedright\arraybackslash}X}

  \caption{Showcase of the synthesized spatio-temporal data and text description pair.}
  \label{tab:showcase_data_sample}
  \begingroup
  \small
  \setlength{\tabcolsep}{4pt}
  \renewcommand{\arraystretch}{0.92}
  \begin{tabularx}{\linewidth}{@{} L{2.4cm} Y @{}}
    \toprule
    % \multicolumn{2}{@{}l@{}}{\textbf{Showcase of the synthesized spatio-temporal data and text description pair}}\\
    % \midrule

    \textbf{Spatio-Temporal Data} &
    \\
    \addlinespace[0.5em]
    & \begin{minipage}[t]{\linewidth}
        \vspace{0pt}
        \begin{center}
          \begin{minipage}[t]{0.4\linewidth}
            \centering
            \includegraphics[width=\linewidth]{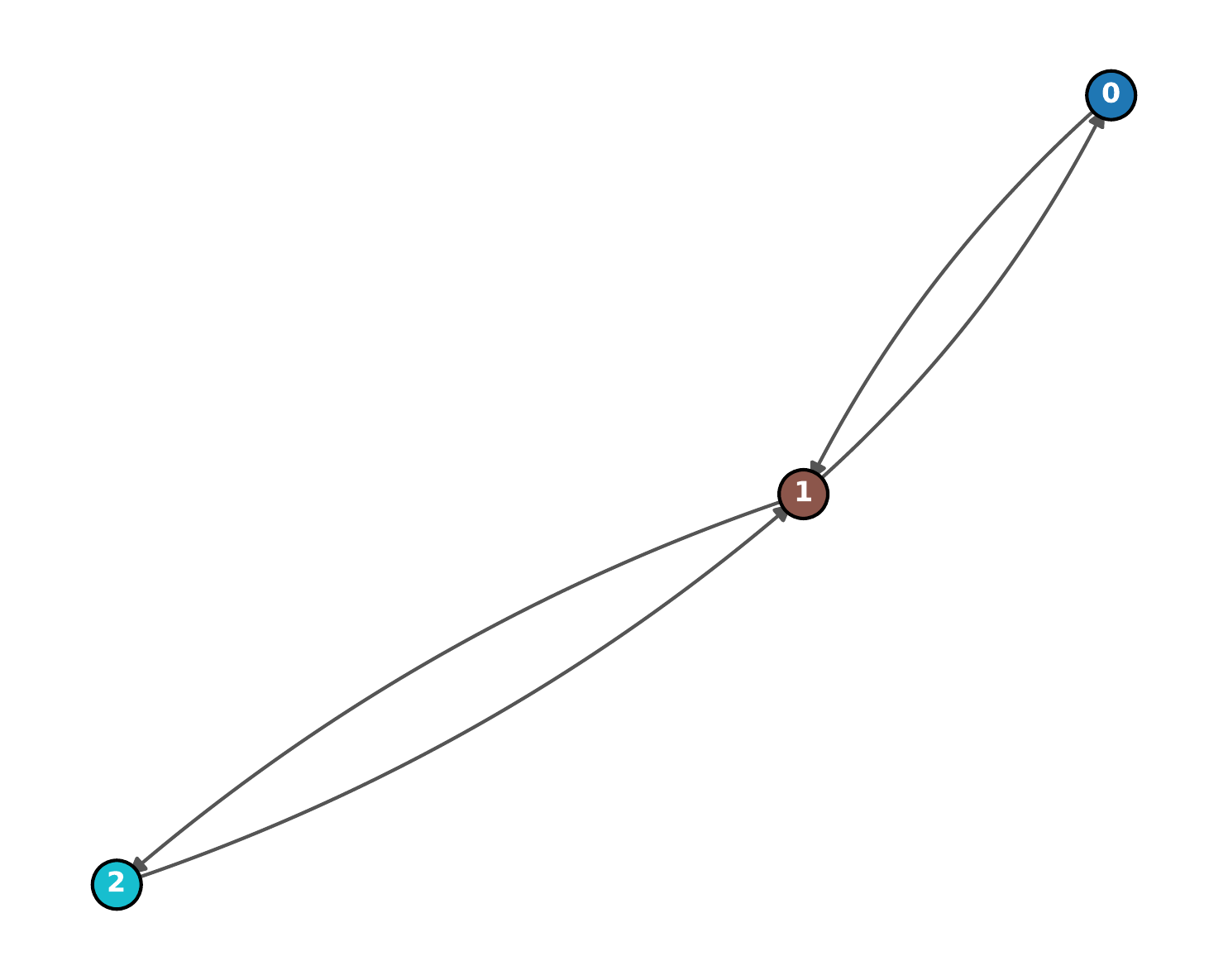}
          \end{minipage}
          \hfill
          \begin{minipage}[t]{0.58\linewidth}
            \centering
            \includegraphics[width=\linewidth]{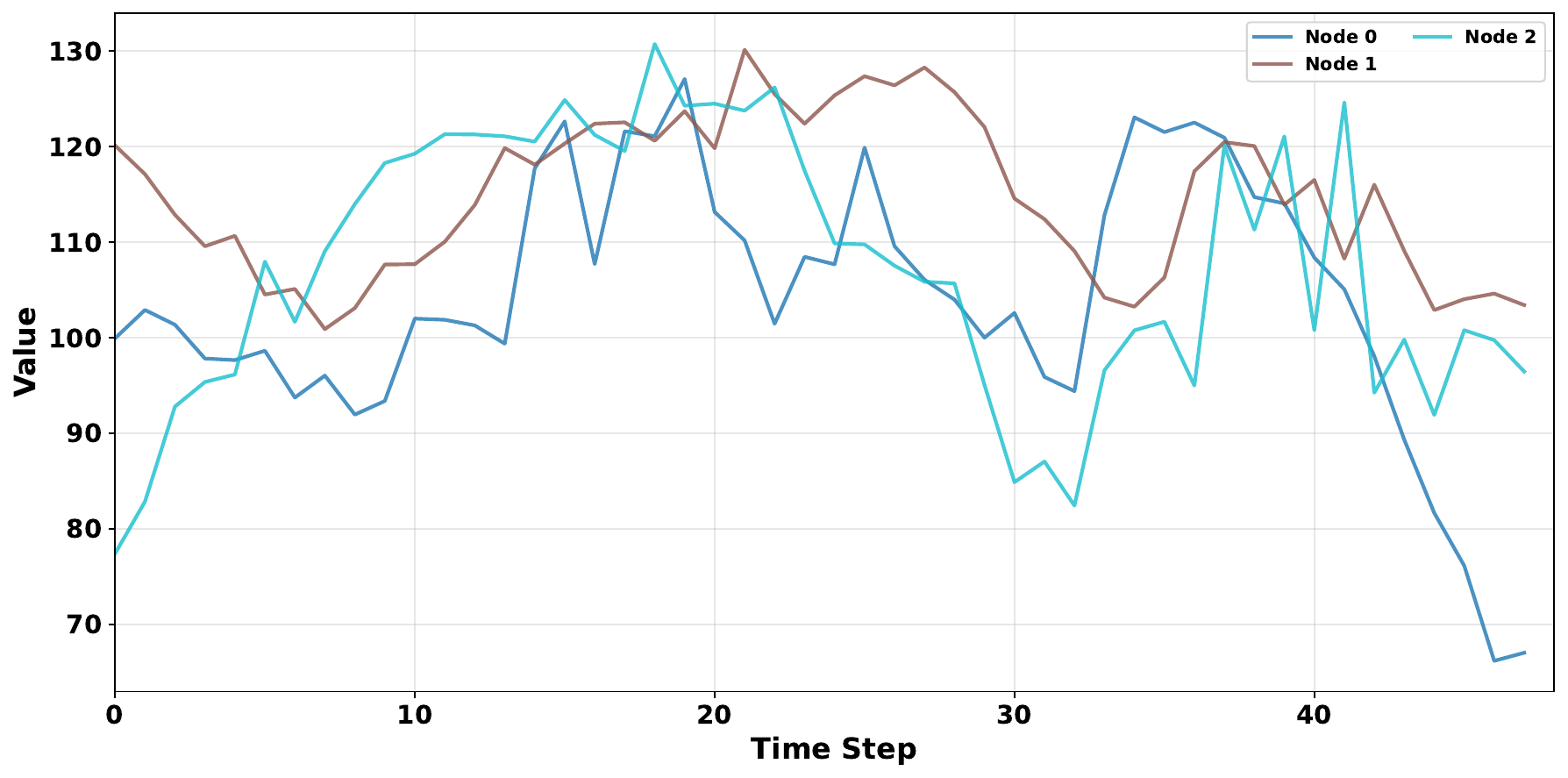}
          \end{minipage}
        \end{center}
      \end{minipage}
    \\

    \textbf{Scenario Generation Agent} &
    \begin{minipage}[t]{\linewidth}
      \vspace{0pt}
      \noindent I'll revise the transportation scenario to fix the timing inconsistencies while maintaining all other constraints.

\noindent \textbf{TIME SPAN:} 1 day \quad \textbf{SAMPLING FREQUENCY:} 30 minutes (48 total points) \quad \textbf{VARIABLE:} traffic flow (vehicles/hour)

\noindent \textbf{NODES:} NODE 0: [DEMAND\_SOURCE] Central business district with offices and commercial areas; NODE 1: [PROPAGATION] Major highway junction connecting business district to residential area; NODE 2: [DEMAND\_SOURCE] Large residential suburb area.

\noindent \textbf{EDGES:} NODE 0 $\rightarrow$ NODE 1: Primary outbound route from business district, time\_lag=1; NODE 1 $\rightarrow$ NODE 2: Main highway section leading to residential area, time\_lag=1; NODE 2 $\rightarrow$ NODE 1: Inbound commuter route from residential area, time\_lag=1; NODE 1 $\rightarrow$ NODE 0: Primary inbound route to business district, time\_lag=1.

\noindent \textbf{TEMPORAL PATTERNS:} NODE 0: Phase 1 (time 0--13) Night period, baseline: 100 vehicles/hour, amplitude: 0, peak: null; Phase 2 (time 14--34) Evening exodus, baseline: 100 vehicles/hour, amplitude: 300 vehicles/hour, peak: 17 (5:00 PM); propagated\_variations: receives morning commute flow from NODE 2 via NODE 1 during time 16--20 [FIXED: accounts for Node 2's peak at t=14 + 2 time\_lags]. NODE 1: baseline: 120 vehicles/hour, amplitude: 0, peak: null; propagated\_variations: receives evening exodus from NODE 0 during time 18--22 [FIXED: accounts for Node 0's peak at t=17 + 1 time\_lag]; receives morning commute from NODE 2 during time 15--19 [FIXED: accounts for Node 2's peak at t=14 + 1 time\_lag]. NODE 2: Phase 1 (time 0--13) Night period, baseline: 110 vehicles/hour, amplitude: 0, peak: null; Phase 2 (time 14--34) Morning commute, baseline: 110 vehicles/hour, amplitude: 280 vehicles/hour, peak: 14 (7:00 AM); propagated\_variations: receives evening exodus flow from NODE 0 via NODE 1 during time 19--23 [FIXED: accounts for Node 0's peak at t=17 + 2 time\_lags].

\noindent \textbf{Edge Modulation:} Time 14--19 [FIXED: aligned with Node 2's peak activity at t=14], edges affected: NODE 2 $\rightarrow$ NODE 1, NODE 1 $\rightarrow$ NODE 0, effect: strong, description: Morning rush hour intensifies inbound traffic flow. Time 17--22 [FIXED: aligned with Node 0's peak activity at t=17], edges affected: NODE 0 $\rightarrow$ NODE 1, NODE 1 $\rightarrow$ NODE 2, effect: strong, description: Evening rush hour intensifies outbound traffic flow.

\noindent \textbf{Note:} The baselines (100, 120, 110 vehicles/hour) remain within the same order of magnitude. All propagation times now properly account for cumulative time\_lags along the paths.      

    \end{minipage}
    \\

    \textbf{Other Agents} &
    \begin{minipage}[t]{\linewidth}
      \vspace{0pt}      
      Due to space limitations, you can find the output of other agents in JSON format (Scenario Parsing Agent in Listing~\ref{lst:Scenario parsing Agent}, SDEs Parameters Agent in Listing~\ref{lst:sde_parameters}, Time-Varying Adj. Agent in Listing~\ref{lst:time_varying_adj})

    \end{minipage}
    \\
    \bottomrule
  \end{tabularx}
  \endgroup
\end{table*}

\clearpage
\section{Reasoning Case Study}
\label{app:case_study}

\begin{table*}[h!]
  \captionsetup{font=small,aboveskip=3pt,belowskip=4pt}
  \centering
  \newcolumntype{L}[1]{>{\raggedright\arraybackslash}p{#1}}
  \newcolumntype{Y}{>{\raggedright\arraybackslash}X}

  \caption{Task 1: Etiological Spatial Reasoning example with options and reasoning process.}
  \label{tab:etiological_example}
  \begingroup
  \small
  \setlength{\tabcolsep}{4pt}
  \renewcommand{\arraystretch}{0.92}
  \begin{tabularx}{\linewidth}{@{} L{2.4cm} Y @{}}
    \toprule
    \multicolumn{2}{@{}l@{}}{\textbf{Task 1: Etiological Spatial Reasoning}}\\
    \midrule

    \textbf{User} &

    You are a spatial temporal analysis expert. Node 0 time series with length of 48: \texttt{<ts><ts/>}; Node 1 time series with length of 48: \texttt{<ts><ts/>}; Node 2 time series with length of 48: \texttt{<ts><ts/>}; Node 3 time series with length of 48: \texttt{<ts><ts/>}; Node 4 time series with length of 48: \texttt{<ts><ts/>}; Graph Structure: Node 0$\rightarrow$Node 2; Node 1$\rightarrow$Node 2; Node 2$\rightarrow$Node 3; Node 3$\rightarrow$Node 4, please analyze the spatial temporal data and answer the following question: Which etiological scenario can be inferred from the spatio-temporal data? 
    \\[0.25em]
    \addlinespace[0.5em]
    & \begin{minipage}[t]{\linewidth}
        \vspace{0pt}
        \begin{center}
          \begin{minipage}[t]{0.48\linewidth}
            \centering
            \includegraphics[width=\linewidth]{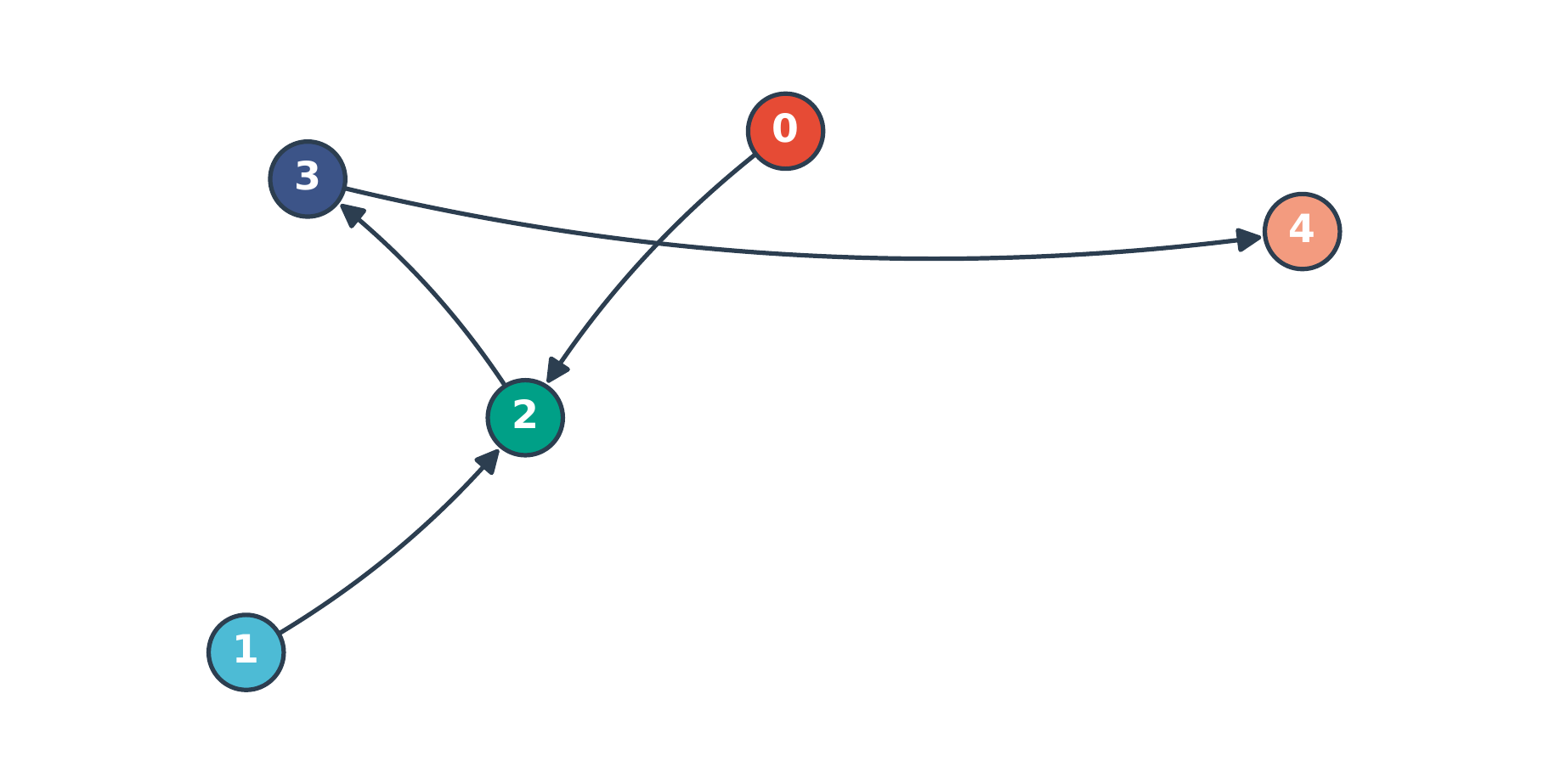}
          \end{minipage}
          \hfill
          \begin{minipage}[t]{0.48\linewidth}
            \centering
            \includegraphics[width=\linewidth]{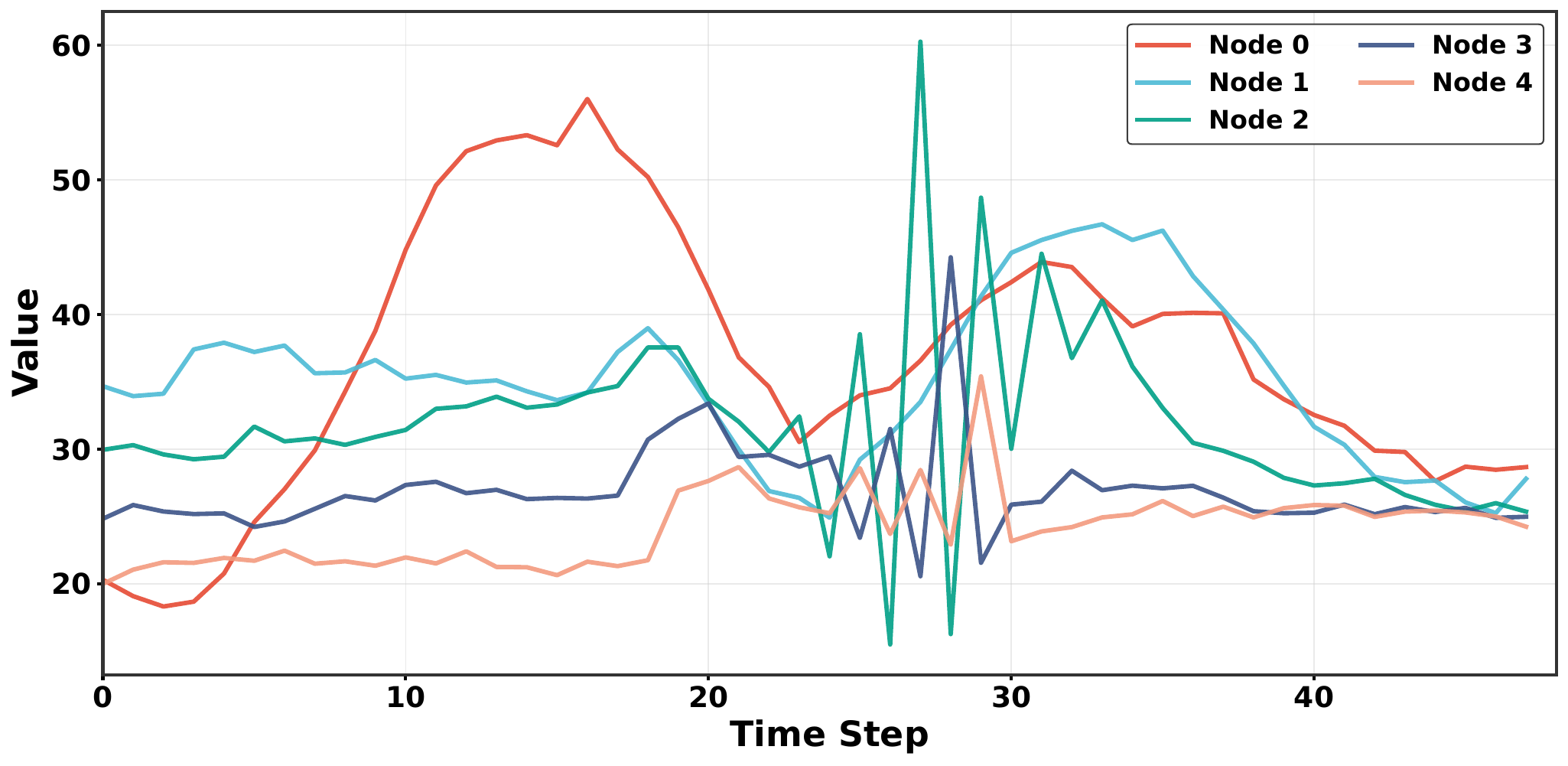}
          \end{minipage}
        \end{center}
      \end{minipage}
    \\
    \addlinespace[0.5em]

    & \textbf{Options:} \\
    & \textbf{\textit{A.}} Factory ventilation system with chemical processes spreading through connected chambers based on pressure differentials.  \\
    & \textbf{\textit{B.}} Coastal emission system with marine and port activities dispersing through atmospheric layers toward inland regions.  \\
    & \textbf{\textit{C.}} Agricultural runoff network with pesticide sources flowing through water channels, affected by seasonal discharge patterns.  \\
    & \textbf{\textit{D.}} Urban pollution network with industrial and traffic sources dispersing through nodes, modulated by time-based transport patterns. \\
    \addlinespace[0.5em]

    \textbf{STReasoner} &
    \begin{minipage}[t]{\linewidth}
      \vspace{0pt}
      \texttt{<think>}
      
      Let me analyze the spatial-temporal data systematically.

      \textbf{Graph Structure Analysis:}
      
      Node 0 $\rightarrow$ Node 2 $\rightarrow$ Node 3 $\rightarrow$ Node 4; Node 1 $\rightarrow$ Node 2.
      This is a converging pattern where Node 0 and Node 1 feed into Node 2, which then propagates through Node 3 to Node 4.

      \textbf{Temporal Pattern Analysis:}
      
      \textbf{1) Node 0 (source).}
      Starts at $\sim 20$, increases to a peak at $56.01$ (timestep 20), then decreases to $\sim 28$ by the end; a clear rise-and-fall event centered near the midpoint.
      
      \textbf{2) Node 1 (source).}
      Starts at $\sim 34.67$, rises to a peak at $46.70$ (timestep 18), then decreases to $\sim 27.79$; similar event with slightly different timing and magnitude.
      
      \textbf{3) Node 2 (convergence).}
      Starts at $\sim 30$ and rises to $\sim 36$, with spikes (e.g., $42.19$ at timestep 19 and $60.27$ at timestep 21) that occur after the peaks in Nodes 0 and 1; then declines to $\sim 25$. This reflects aggregation from two upstream sources and delayed amplification at the junction.
      
      \textbf{4) Node 3 (downstream).}
      Starts at $\sim 25$, increases to $\sim 44$ (timesteps 20--21), showing a delayed response versus upstream nodes; then decreases back to $\sim 25$.
      
      \textbf{5) Node 4 (furthest downstream).}
      Baseline around $20$--$22$, increases to $\sim 35$ (timesteps 22--23), the most delayed and damped response; returns toward baseline ($\sim 24$).

      \textbf{Key Observations:}
      Multiple sources (Nodes 0 and 1) converge at Node 2; there is a clear transport lag along 0/1$\rightarrow$2$\rightarrow$3$\rightarrow$4, and peak magnitudes attenuate downstream. The pattern looks like a single event propagation, not a seasonal cycle.

      \textbf{Scenario Evaluation:}
      \textbf{A)} Ventilation/pressure-driven spread would more likely show sharper, more immediate responses rather than gradual lagged attenuation across the chain.
      \textbf{B)} Coastal/atmospheric dispersion would typically show different signatures (e.g., broader regional mixing and potentially periodic drivers) rather than a clean directed network propagation with two discrete sources.
      \textbf{C)} Agricultural runoff is often tied to seasonal or weather-driven variability; this trace is a single event-like pulse rather than seasonal discharge behavior.
      \textbf{D)} Urban pollution fits: two sources (industrial and traffic) merge, then disperse with time-based transport delays and spatial attenuation through the network.

      Therefore, the inferred scenario is \textbf{D}.
      
      \texttt{</think>}

      \texttt{<answer>}D\texttt{</answer>}
    
    \end{minipage}
    \\
    \bottomrule
  \end{tabularx}
  \endgroup
\end{table*}

\begin{table*}[h!]
  \captionsetup{font=small,aboveskip=3pt,belowskip=4pt}
  \centering
  \newcolumntype{L}[1]{>{\raggedright\arraybackslash}p{#1}}
  \newcolumntype{Y}{>{\raggedright\arraybackslash}X}

  \caption{Task 2: Spatial Entity Identification example with options and reasoning process.}
  \label{tab:entity_example}
  \begingroup
  \small
  \setlength{\tabcolsep}{4pt}
  \renewcommand{\arraystretch}{0.92}
  \begin{tabularx}{\linewidth}{@{} L{2.4cm} Y @{}}
    \toprule
    \multicolumn{2}{@{}l@{}}{\textbf{Task 2: Spatial Entity Identification}}\\
    \midrule

    \textbf{User} &

    You are a spatial temporal analysis expert. Node 0 time series with length of 48: \texttt{<ts><ts/>}; Node 1 time series with length of 48: \texttt{<ts><ts/>}; Node 2 time series with length of 48: \texttt{<ts><ts/>}; ... Node 8 time series with length of 48: \texttt{<ts><ts/>}; Node 9 time series with length of 48: \texttt{<ts><ts/>}; Graph Structure: Node 0$\rightarrow$Node 1; Node 1$\rightarrow$Node 3; Node 2$\rightarrow$Node 3; Node 3$\rightarrow$Node 4; Node 4$\rightarrow$Node 5; Node 5$\rightarrow$Node 6; Node 6$\rightarrow$Node 7; Node 7$\rightarrow$Node 8; Node 8$\rightarrow$Node 9, please analyze the spatial temporal data and answer the following question: Which (name, description) pair should Node 9 correspond to?
    
    \\[0.25em]
    \addlinespace[0.5em]
    & \begin{minipage}[t]{\linewidth}
        \vspace{0pt}
        \begin{center}
          \begin{minipage}[t]{0.48\linewidth}
            \centering
            \includegraphics[width=\linewidth]{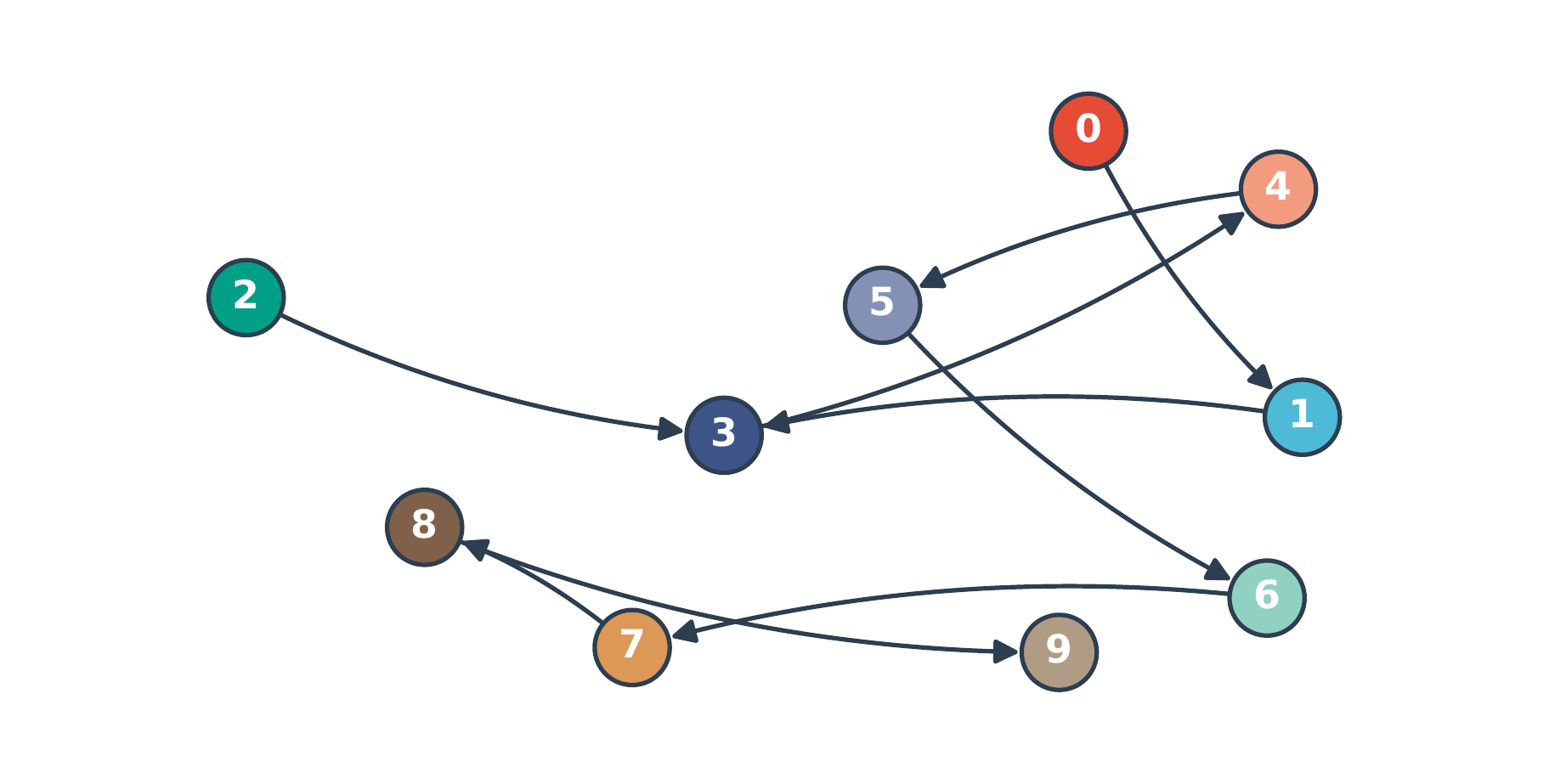}
          \end{minipage}
          \hfill
          \begin{minipage}[t]{0.48\linewidth}
            \centering
            \includegraphics[width=\linewidth]{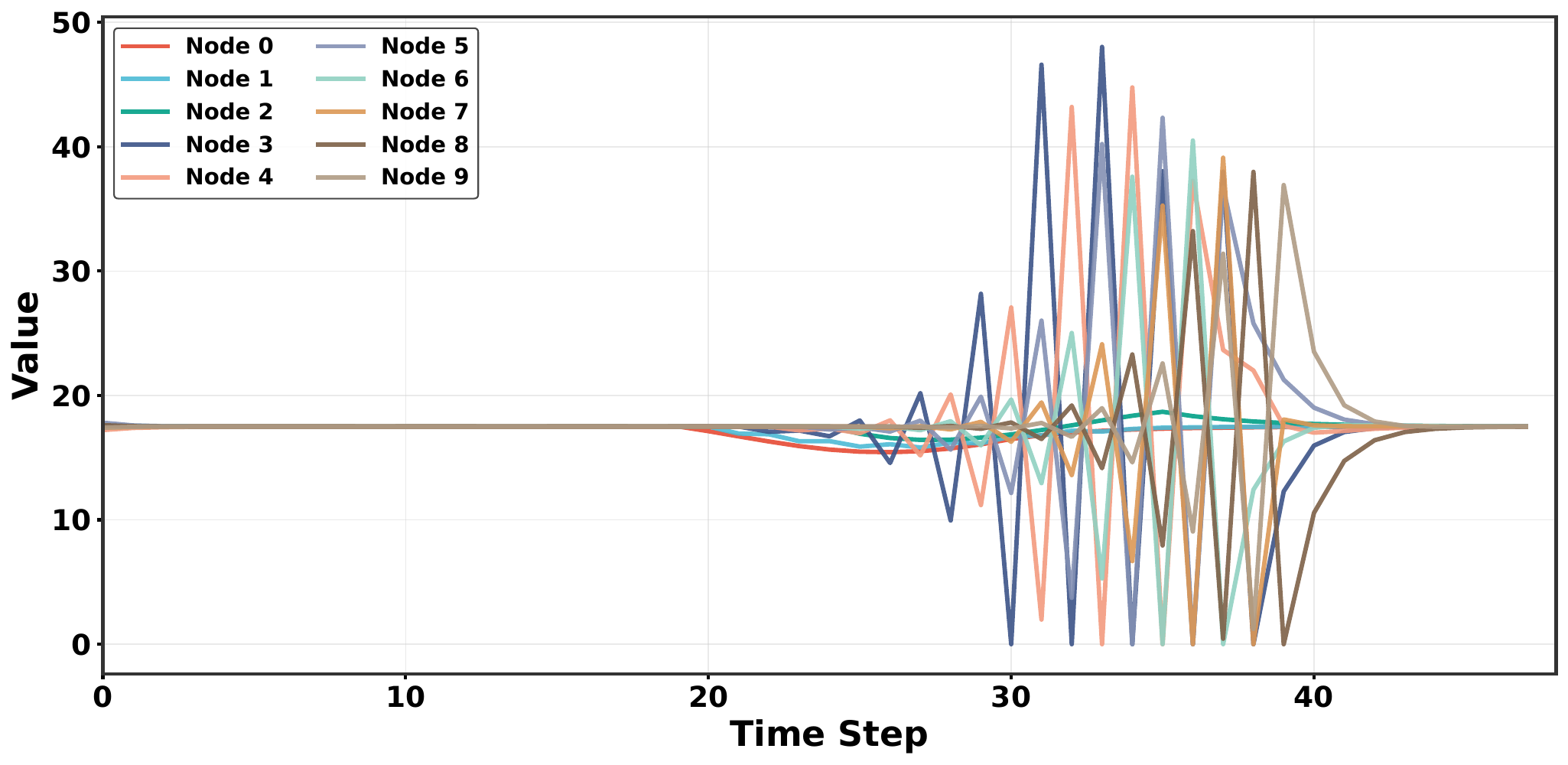}
          \end{minipage}
        \end{center}
      \end{minipage}
    \\
    \addlinespace[0.5em]

    & \textbf{Options:} \\
    & \textbf{\textit{A.}} Endpoint Alpha, Last measurement checkpoint.  \\
    & \textbf{\textit{B.}} Exit Node, Terminal collection station.  \\
    & \textbf{\textit{C.}} Final Point, Final monitoring point.  \\
    & \textbf{\textit{D.}} Terminal Hub, Network termination sensor. \\
    \addlinespace[0.5em]

    \textbf{STReasoner} &
    \begin{minipage}[t]{\linewidth}
      \vspace{0pt}
      \texttt{<think>}

      Let me analyze the spatial-temporal data systematically.

    First, let me understand the graph structure:  
    Node 0 $\rightarrow$ Node 1 $\rightarrow$ Node 3 $\leftarrow$ Node 2;  
    Node 3 $\rightarrow$ Node 4 $\rightarrow$ Node 5 $\rightarrow$ Node 6 $\rightarrow$ Node 7 $\rightarrow$ Node 8 $\rightarrow$ Node 9.
    
    This forms a linear chain from Node~3 to Node~9, with Node~9 being the last node in the sequence.
    
    Now let me analyze the temporal patterns.
    
    \textbf{1. Nodes 0--2.}  
    These nodes show relatively stable values around 17.5, with minor fluctuations. They act as early nodes in the network.
    
    \textbf{2. Node 3.}  
    Node~3 shows a dramatic change around timestep 32--36, with extreme spikes (values such as 48.03, 39.12, and 35.01) followed by drops to 0.00, and then gradually recovers back to approximately 17.5.
    
    \textbf{3. Nodes 4--8.}  
    Each of these nodes exhibits a similar disturbance pattern that propagates through the network with increasing delay:
    Node~4 shows a spike around timesteps 33--37 (44.77, 39.21, 34.65);
    Node~5 around timesteps 34--38 (42.33, 38.41, 33.90);
    Node~6 around timesteps 35--39 (40.50, 37.60, 35.22);
    Node~7 around timesteps 36--40 (39.12, 37.22, 35.64);
    Node~8 around timesteps 37--41 (37.98, 36.51, 34.61).
    
    \textbf{4. Node 9.}  
    Node~9 receives the disturbance last, around timesteps 38--42 (36.92, 34.12, 32.26, 31.23). The magnitude is lower and more dampened compared to earlier nodes.
    
    \textbf{Key observations.}  
    The disturbance propagates sequentially through the network. Node~9 lies at the end of the chain (Node~8 $\rightarrow$ Node~9), shows the final and most dampened version of the disturbance, and has no outgoing connections, indicating that it is a terminal node.
    
    Looking at the options:  
    Option A, ``Endpoint Alpha, Last measurement checkpoint,'' suggests a specific endpoint.  
    Option B, ``Exit Node, Terminal collection station,'' suggests an exit or collection point.  
    Option C, ``Final Point, Final monitoring point,'' clearly indicates the last monitoring location.  
    Option D, ``Terminal Hub, Network termination sensor,'' suggests network termination.
    
    Given that Node~9 is the last node in the sequential chain, receives the propagated disturbance last, has no outgoing connections, and represents the final monitoring position in the network, the most appropriate description is ``Final Point, Final monitoring point'' (Option~C).

      \texttt{</think>}

      \texttt{<answer>}C\texttt{</answer>}
    
    \end{minipage}
    \\
    \bottomrule
  \end{tabularx}
  \endgroup
\end{table*}

\begin{table*}[h!]
  \captionsetup{font=small,aboveskip=3pt,belowskip=4pt}
  \centering
  \newcolumntype{L}[1]{>{\raggedright\arraybackslash}p{#1}}
  \newcolumntype{Y}{>{\raggedright\arraybackslash}X}

  \caption{Task 3: Spatial Correlation Reasoning example with options and reasoning process.}
  \label{tab:correlation_example}
  \begingroup
  \small
  \setlength{\tabcolsep}{4pt}
  \renewcommand{\arraystretch}{0.92}
  \begin{tabularx}{\linewidth}{@{} L{2.4cm} Y @{}}
    \toprule
    \multicolumn{2}{@{}l@{}}{\textbf{Task 3: Spatial Correlation Reasoning}}\\
    \midrule

    \textbf{User} &

    You are a spatial temporal analysis expert. Node 0 time series with length of 240: \texttt{<ts><ts/>}; Node 1 time series with length of 240: \texttt{<ts><ts/>}; Node 2 time series with length of 240: \texttt{<ts><ts/>}; ...; Node 8 time series with length of 240: \texttt{<ts><ts/>}; Node 9 time series with length of 240: \texttt{<ts><ts/>}; Graph Structure: Node 0$\rightarrow$Node 2; Node 0$\rightarrow$Node 6; Node 1$\rightarrow$Node 2; Node 2$\rightarrow$Node 3; Node 2$\rightarrow$Node 5; Node 3$\rightarrow$Node 4; Node 4$\rightarrow$Node 8; Node 5$\rightarrow$Node 7; Node 6$\rightarrow$Node 7; Node 7$\rightarrow$Node 8; Node 8$\rightarrow$Node 9, please analyze the spatial temporal data and answer the following question: Which statement best describes the relationship between Node 2 and Node 4 during time steps 177-180 (1 time step = 1 hour)? 
    
    \\[0.25em]
    \addlinespace[0.5em]
    & \begin{minipage}[t]{\linewidth}
        \vspace{0pt}
        \begin{center}
          \begin{minipage}[t]{0.48\linewidth}
            \centering
            \includegraphics[width=\linewidth]{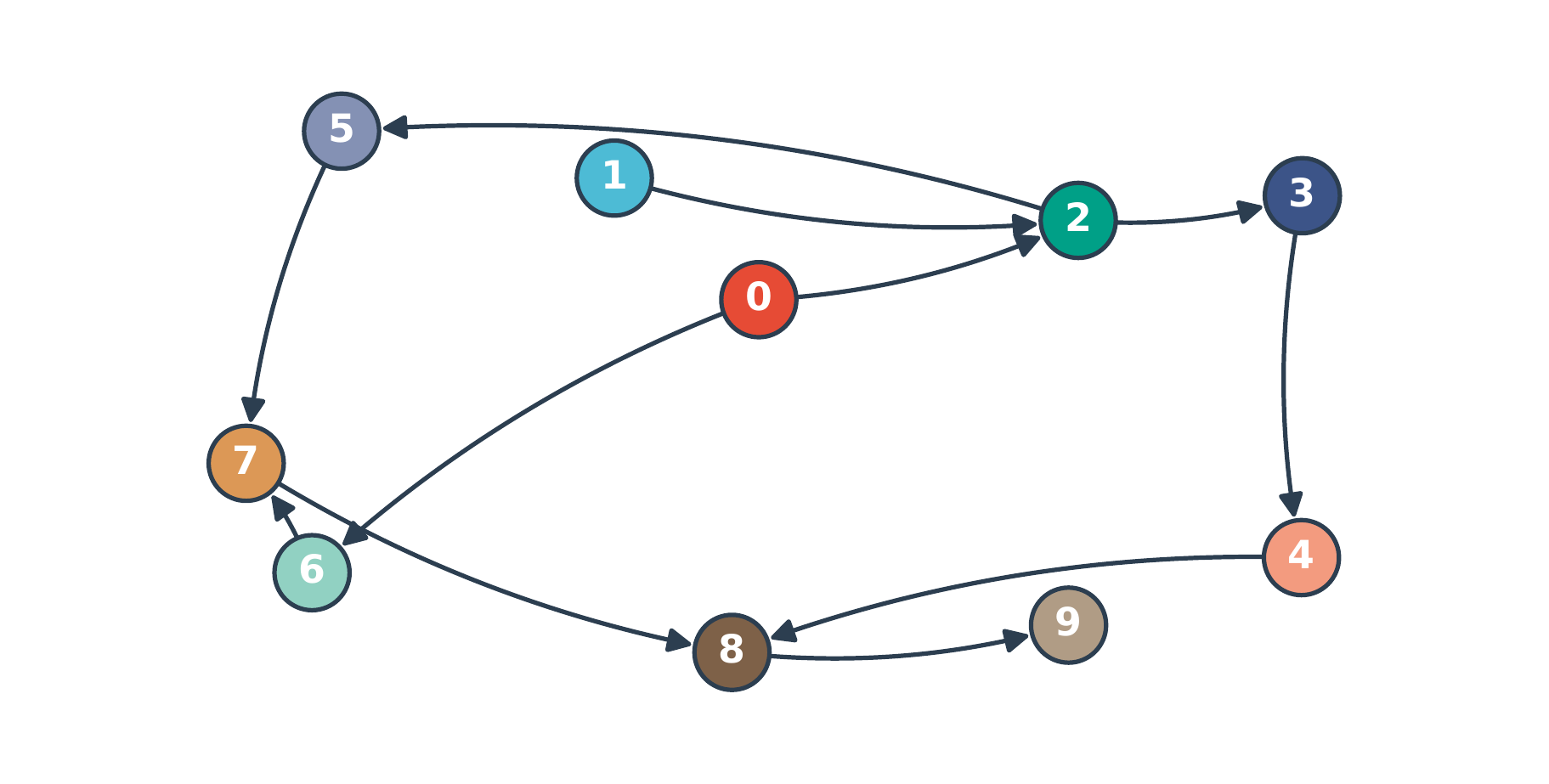}
          \end{minipage}
          \hfill
          \begin{minipage}[t]{0.48\linewidth}
            \centering
            \includegraphics[width=\linewidth]{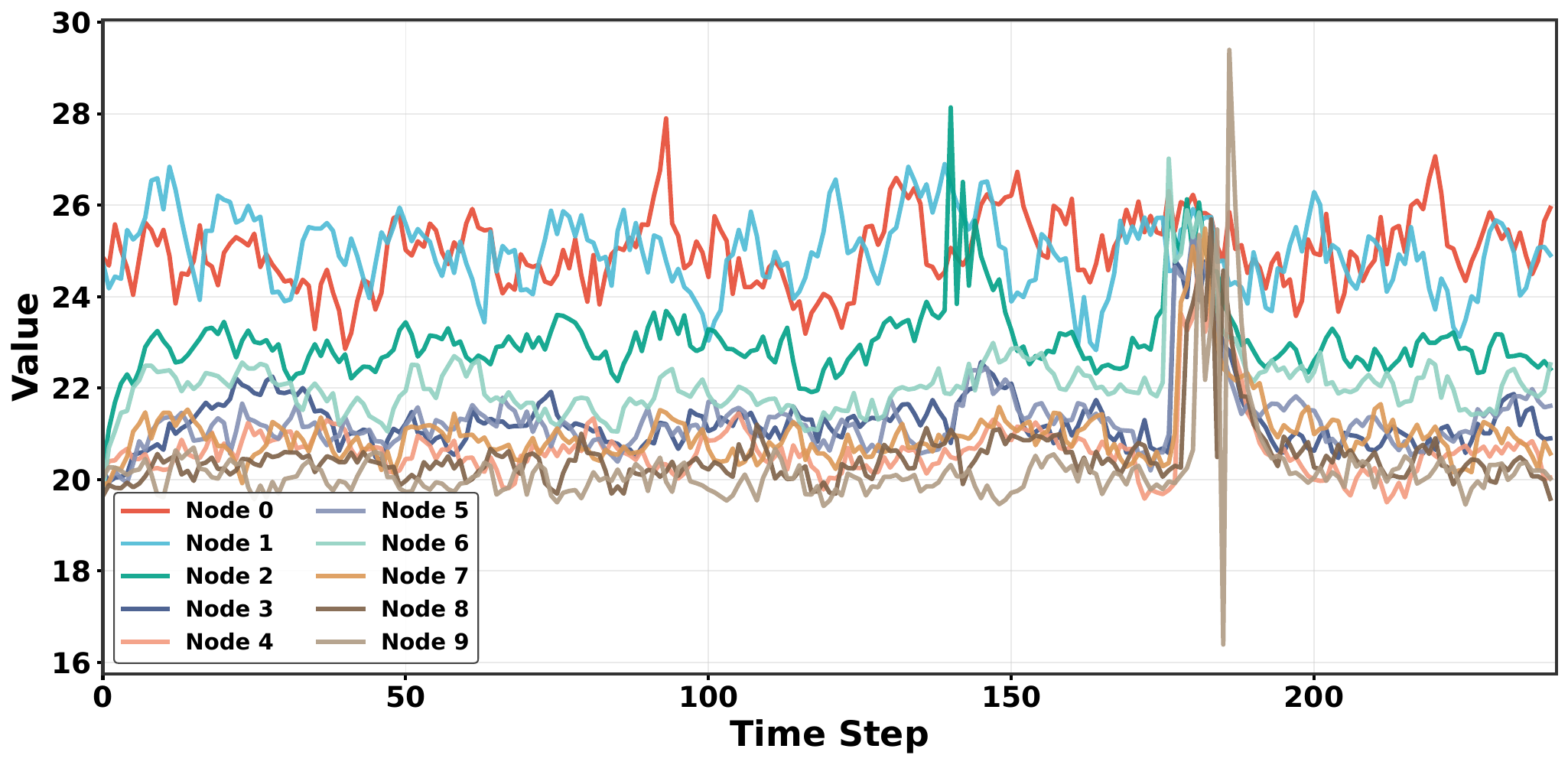}
          \end{minipage}
        \end{center}
      \end{minipage}
    \\
    \addlinespace[0.5em]

    & \textbf{Options:} \\
    & \textbf{\textit{A.}} There is no significant thermal signal transmission between Node 2 and Node 4.  \\
    & \textbf{\textit{B.}} Thermal signal travels directly from Node 2 to Node 4 without intermediate nodes.  \\
    & \textbf{\textit{C.}} Thermal signal flows from Node 4 back to Node 2 through a feedback loop.  \\
    & \textbf{\textit{D.}} Thermal signal propagates from Node 2 to Node 4 via Node 3, following a sequential central-to-eastern pathway. \\
    \addlinespace[0.5em]

    \textbf{STReasoner} &
    \begin{minipage}[t]{\linewidth}
      \vspace{0pt}
      \texttt{<think>}
    Let me analyze the spatial-temporal data for the relationship between Node 2 and Node 4 during time steps 177-180. First, let me identify the graph structure path from Node 2 to Node 4: Node 2 $\rightarrow$ Node 3 $\rightarrow$ Node 4. This means Node 3 is an intermediate node between Node 2 and Node 4. Now let me extract the values at time steps 177-180 (indices 176-179 in 0-indexed arrays).
    
    Node 2: Time 177: 21.91; Time 178: 22.01; Time 179: 22.11; Time 180: 22.21.  
    Node 3: Time 177: 21.02; Time 178: 21.07; Time 179: 21.11; Time 180: 21.15.  
    Node 4: Time 177: 20.04; Time 178: 20.05; Time 179: 20.06; Time 180: 20.06.
    
    Let me also check what happens before and after to understand the temporal pattern. Node 2 (earlier): Time 176: 21.81; Time 177: 21.91 (increase of 0.10); Time 178: 22.01 (increase of 0.10); Time 179: 22.11 (increase of 0.10); Time 180: 22.21 (increase of 0.10). Node 3 (earlier): Time 176: 21.01; Time 177: 21.02 (increase of 0.01); Time 178: 21.07 (increase of 0.05); Time 179: 21.11 (increase of 0.04); Time 180: 21.15 (increase of 0.04). Node 4 (earlier): Time 176: 20.03; Time 177: 20.04 (increase of 0.01); Time 178: 20.05 (increase of 0.01); Time 179: 20.06 (increase of 0.01); Time 180: 20.06 (stable).
    
    Looking at the pattern: (1) Node 2 shows a clear increasing thermal signal starting at time 177; (2) Node 3 shows an increasing pattern with a slight delay and dampening; (3) Node 4 shows a gradual increase with further delay. The signal appears to propagate from Node 2 $\rightarrow$ Node 3 $\rightarrow$ Node 4, which matches the graph structure. This is a sequential propagation through an intermediate node (Node 3).
    
    Let me evaluate the options. A. No significant thermal signal -- FALSE, there are clear changes in values. B. Direct transmission without intermediate nodes -- FALSE, the graph shows Node 2$\rightarrow$Node 3$\rightarrow$Node 4, and the temporal pattern shows Node 3 responds first before Node 4. C. Flow from Node 4 back to Node 2 -- FALSE, the temporal pattern shows Node 2 changes first, then Node 3, then Node 4. D. Propagates from Node 2 to Node 4 via Node 3 -- TRUE, this matches both the graph structure (Node 2$\rightarrow$Node 3$\rightarrow$Node 4) and the temporal pattern where changes in Node 2 are followed by changes in Node 3, then Node 4. 

      \texttt{</think>}

      \texttt{<answer>}D\texttt{</answer>}
    
    \end{minipage}
    \\
    \bottomrule
  \end{tabularx}
  \endgroup
\end{table*}

\begin{table*}[h!]
  \captionsetup{font=small,aboveskip=3pt,belowskip=4pt}
  \centering
  \newcolumntype{L}[1]{>{\raggedright\arraybackslash}p{#1}}
  \newcolumntype{Y}{>{\raggedright\arraybackslash}X}

  \caption{Task 4: In-Context Spatio-Temporal Forecasting example with options and reasoning process.}
  \label{tab:forecasting_example}
  \begingroup
  \small
  \setlength{\tabcolsep}{4pt}
  \renewcommand{\arraystretch}{0.92}
  \begin{tabularx}{\linewidth}{@{} L{2.4cm} Y @{}}
    \toprule
    \multicolumn{2}{@{}l@{}}{\textbf{Task 4: In-Context Spatio-Temporal Forecasting}}\\
    \midrule

    \textbf{User} &
    You are a spatio-temporal analysis expert. Node 0 time series with length of 36: \texttt{<ts><ts/>}; Node 1 time series with length of 36: \texttt{<ts><ts/>}; Node 2 time series with length of 36: \texttt{<ts><ts/>}; Graph Structure: Node 0$\rightarrow$Node 2; Node 1$\rightarrow$Node 2; Node 2$\rightarrow$Node 1, please analyze the spatial temporal data and answer the following question: Given the context Maximum solar heating period enhances thermal transfer, predict the value of node 2 for the next 3 steps. Historical observation window: 30-35.
    \\[0.25em]
    \addlinespace[0.5em]
    & \begin{minipage}[t]{\linewidth}
        \vspace{0pt}
        \begin{center}
          \begin{minipage}[t]{0.48\linewidth}
            \centering
            \includegraphics[width=\linewidth]{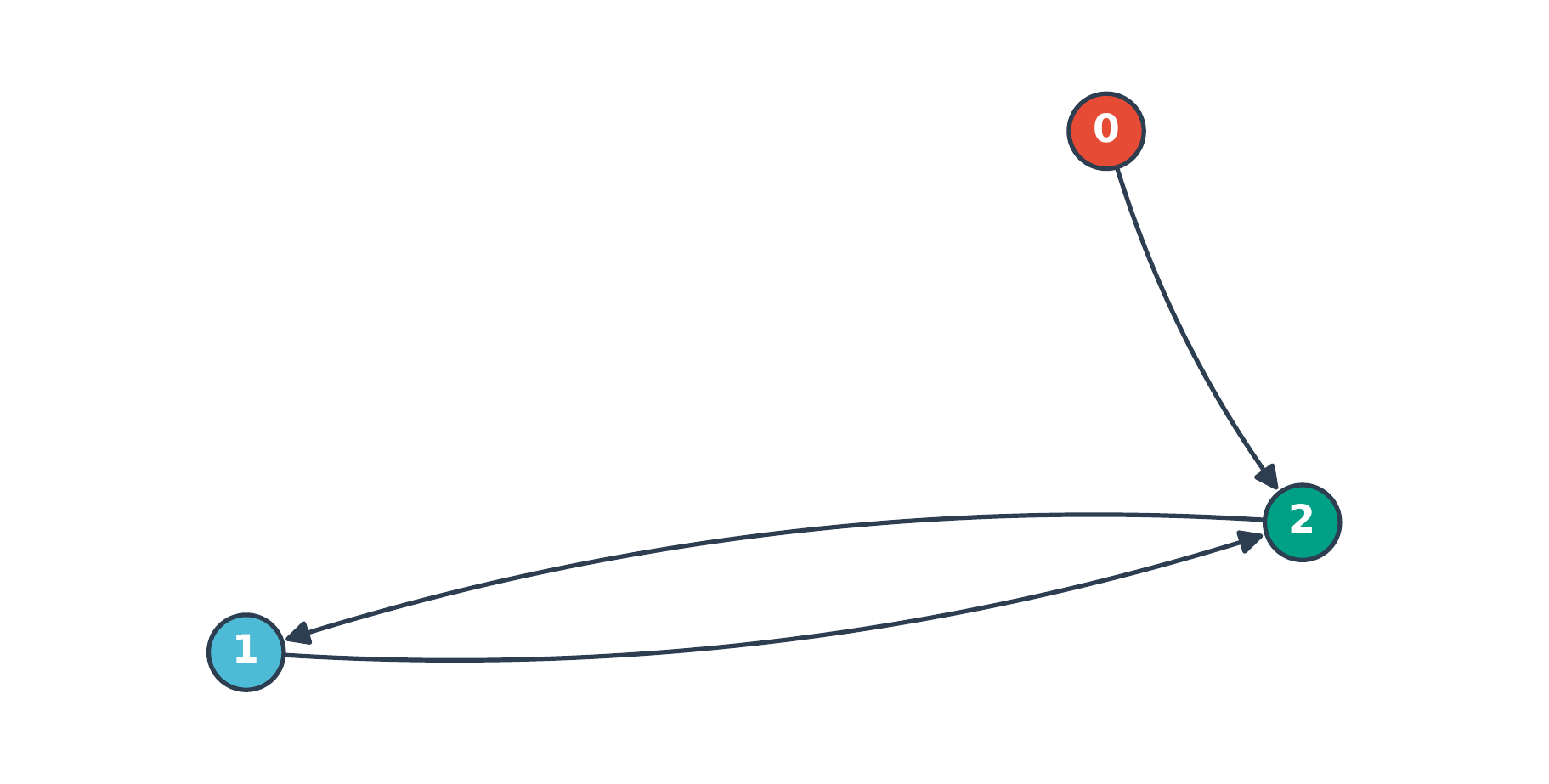}
          \end{minipage}
          \hfill
          \begin{minipage}[t]{0.48\linewidth}
            \centering
            \includegraphics[width=\linewidth]{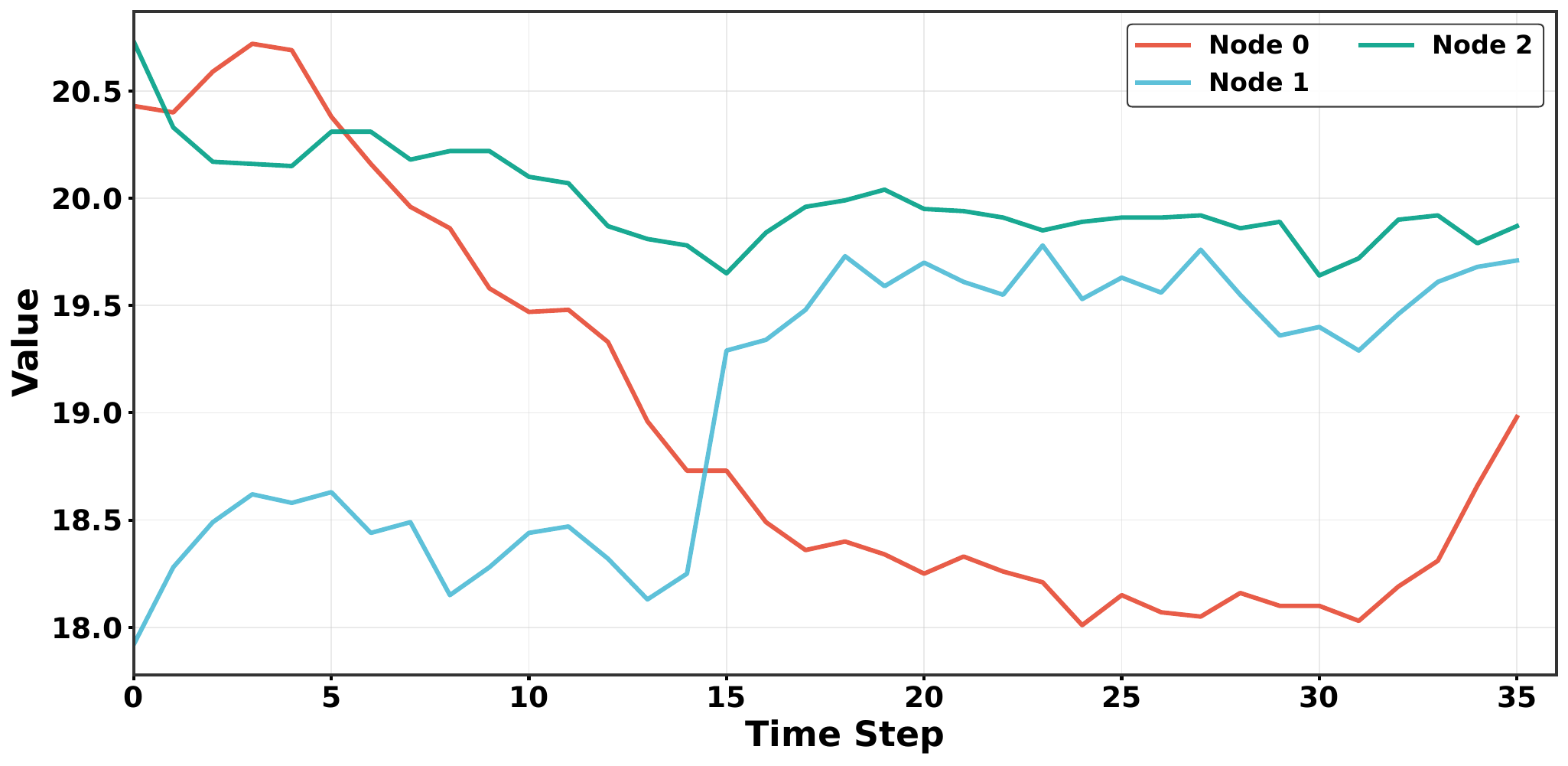}
          \end{minipage}
        \end{center}
      \end{minipage}
    \\

    \textbf{STReasoner} &
    \begin{minipage}[t]{\linewidth}
      \vspace{0pt}
      \texttt{<think>}
      
      Let me analyze the spatial-temporal data step by step.

    \textbf{Step 1: Understand the Graph Structure.}
    The graph structure consists of directed dependencies among nodes. Node~0 influences Node~2, and Node~1 also influences Node~2, indicating two upstream sources. In addition, Node~2 feeds back to Node~1, forming a local feedback loop between Node~1 and Node~2.
    
    \textbf{Step 2: Analyze the Historical Window (positions 30--35).}
    Within the historical observation window, Node~0 has values 19.27, 19.41, 19.51, 19.61, 19.68, and 18.97. Node~1 records values of 19.39, 19.52, 19.61, 19.68, 19.78, and 19.71. Node~2 shows a smooth increasing sequence with values 19.69, 19.74, 19.78, 19.82, 19.85, and 19.87.
    
    \textbf{Step 3: Identify Trends.}
    Considering the full time series, Node~0 exhibits an increasing trend until approximately position~23, where it reaches around 20.72, followed by a decreasing phase with a sharp drop at position~35. Node~1 shows a strong overall increasing trend from 17.92 to around 19.78 at position~33, followed by a slight decrease at the end. Node~2 maintains a steady and monotonic increase from about 19.64 to 19.87 by position~35.
    
    \textbf{Step 4: Context Analysis.}
    The context statement ``Maximum solar heating period enhances thermal transfer'' suggests a heating-dominated regime. However, the observed data indicate that Node~0 has reached saturation and started to decline, while Node~1 continues to increase but at a slower rate. Node~2 follows a stable increasing trajectory that reflects combined influence from both upstream nodes.
    
    \textbf{Step 5: Predict Next 3 Steps for Node 2.}
    Given that Node~2 is jointly influenced by Node~0 and Node~1, the sharp decline in Node~0 at position~35 and the slight decrease in Node~1 suggest that the growth rate of Node~2 will slow down. Although Node~2 has been steadily increasing, it is likely approaching saturation. Therefore, the predicted values for Node~2 are approximately 19.88 at position~36, 19.89 at position~37, and 19.90 at position~38, reflecting continued but diminishing growth.

      \texttt{</think>}

      \texttt{<answer>}[19.88, 19.89, 19.90]\texttt{</answer>}
    
    \end{minipage}
    \\
    \bottomrule
  \end{tabularx}
  \endgroup
\end{table*}

\clearpage

\begin{algorithm}[t]
\caption{Multi-Agent Pipeline for Network SDE Data Synthesis}
\label{alg:data_synthesis}
\begin{algorithmic}[1]
\Require $N$ (number of nodes)
\Ensure $\mathcal{S}$ (structured scenario), $\Theta$ (network SDE), $\mathbf{X}$ (time series) 

\State $s \leftarrow \textsc{Agent}_1(N)$ \Comment{Agent 1: scenario generation}
\State $T \leftarrow \textsc{InferSeqLen}(s)$

\State $\mathcal{S} \leftarrow \textsc{Agent}_2(s)$ \Comment{Agent 2: scenario parsing with Judge 1 refinement}
\For{$\ell = 1 \dots L$}
\State $j \leftarrow \textsc{Judge}_1(s, \mathcal{S})$
\If{j = \text{approve}}
  \State \textbf{break}
\EndIf
\State $s \leftarrow \textsc{Agent}_1(s, j)$
\State $\mathcal{S} \leftarrow \textsc{Agent}_2(s)$
\EndFor

\State $\Theta \leftarrow \textsc{Agent}_3(\mathcal{S})$ \Comment{Agents 3--4: SDE and adjacency with Judge 2 refinement}
\State $A \leftarrow \textsc{Agent}_4(\mathcal{S})$
\For{$k = 1 \dots K$}
\State $\tilde{\mathbf{X}} \leftarrow \textsc{Simulate}(\Theta, A, T)$
\State $j \leftarrow \textsc{Judge}_2(\mathcal{S}, \tilde{\mathbf{X}})$
\If{j = \text{approve}}
  \State \textbf{break}
\EndIf
\State $\Theta \leftarrow \textsc{Agent}_3(\mathcal{S})$
\State $A \leftarrow \textsc{Agent}_4(\mathcal{S})$
\EndFor

\State $\mathbf{X} \leftarrow \textsc{Simulate}(\Theta, A, T)$
\State \Return $\mathcal{S}, \Theta, \mathbf{X}$
\end{algorithmic}
\end{algorithm}

\section{ST-Bench Prompt Template}
\label{app:dataset_prompt}
\begin{tcolorbox}[
    colback=gray!15,
    colframe=gray!75,
    boxrule=0pt,
    arc=2mm,
    enhanced,
    breakable,
    title={Etiological Reasoning Dataset Generation Prompt},
    fonttitle=\bfseries,
]
\label{prompt:align}
\begin{verbatim}

You are given spatio-temporal context. Produce a multiple-choice item where the 
correct option is a concise macro summary of the Scenario Context.

Scenario: {scenario}

Requirements:
1) "observation": A concise macro summary of the Scenario in 12–20 words.
   - It must describe the system at a high level (e.g., an interconnected hydroponics 
   circulation system, a wastewater treatment facility, etc.). Facility names are not 
   important; do not invent new names.
   - It must explicitly mention the key node variables provided.
2) "options": list of exactly four scenario summaries (each <20 words) without 
labels.
   - The FIRST entry must be identical to "observation" (verbatim match).
   - The other three must be fluent but incorrect (they must contradict the Scenario 
   or mention entities/processes not present in the Scenario/Involved Nodes).

Output JSON format:
{{
    "observation": "observed phenomena",
    "options": ["Correct summary", "Distractor 1", "Distractor 2", "Distractor 3"]
}}

Strictly based on given context, do not introduce external knowledge. Return only a 
valid JSON object, without any explanations outside JSON.
\end{verbatim}
\end{tcolorbox}

\begin{tcolorbox}[
    colback=gray!15,
    colframe=gray!75,
    boxrule=0pt,
    arc=2mm,
    enhanced,
    breakable,
    title={Entity Reasoning Dataset Generation Template},
    fonttitle=\bfseries,
]
\label{prompt:align}
\begin{verbatim}

You are given a node ID and its correct name and description from a spatio-temporal 
network simulation.
Your task is to generate a multiple-choice question to identify which 
(name, description) pair corresponds to the target node.

Target Node:
- ID: {node_id}
- Correct Name: {node_name}
- Correct Description: {node_description}

Requirements:
1) "options": list of exactly four strings containing (name, description) pairs. 
Do NOT prefix with labels.
   - The FIRST entry must be the correct pair (verbatim match to the given name 
   and description).
   - The remaining three must be fluent but !!!incorrect!!!.
       - They should describe plausible but different node roles or locations.
       - Maintain the same style and variable domain (e.g., “traffic flow”, 
       “industrial output”, “water pressure”).
       - Avoid contradictions or unrealistic content.

Output JSON format:
{{
    "question": "Which (name, description) pair should Node {node_id} 
    correspond to?",
    "options": ["Correct pair", "Distractor 1", "Distractor 2", "Distractor 3"]
}}

Strictly return only a valid JSON object. Do not add explanations outside JSON.
\end{verbatim}
\end{tcolorbox}

\begin{tcolorbox}[
    colback=gray!15,
    colframe=gray!75,
    boxrule=0pt,
    arc=2mm,
    enhanced,
    breakable,
    title={Correlation Reasoning (Direct Causal) Dataset Generation Template},
    fonttitle=\bfseries,
]
\label{prompt:align}
\begin{verbatim}

You are a question designer for a spatio-temporal reasoning test.
Your task is to create a multiple-choice question based on a given piece of 
true evidence of a causal influence.

True Evidence:
- Source Node: {source_node_name} (ID: {source_node_id})
- Target Node: {target_node_name} (ID: {target_node_id})
- Time Steps: {time_period} (1 time step = {sampling_frequency})
- Correct Description of Event: "{correct_description}"

Your Task:
1.  Create a "question" that asks which statement best describes the influence on 
Node {target_node_id} during the specified time steps.
    - Explicitly use the phrase "time steps {time_period}" in the question text and 
    append "(1 time step = {sampling_frequency})".
2.  Create an "options" list containing exactly four strings. The FIRST entry MUST 
be the Correct Description provided above, verbatim.
3.  The remaining three entries must be plausible but incorrect distractors 
describing different sources or incorrect effects.

Output Format:
Return ONLY a valid JSON object.
{{
    "question": "Your generated question.",
    "options": ["The correct description verbatim", "Distractor 1", "Distractor 2"
    , "Distractor 3"]
}}
\end{verbatim}
\end{tcolorbox}

\begin{tcolorbox}[
    colback=gray!15,
    colframe=gray!75,
    boxrule=0pt,
    arc=2mm,
    enhanced,
    breakable,
    title={Correlation Reasoning (Multi Hop) Dataset Generation Template},
    fonttitle=\bfseries,
]
\label{prompt:align}
\begin{verbatim}

You are an expert in spatio-temporal network analysis. Your task is to analyze a set 
of direct causal events and identify a multi-hop propagation path to create a 
multiple-choice question.

Given Direct Causal Events (Adjacency Modulations):
{adjacency_modulations}

Time Step Note: 1 time step = {sampling_frequency}

Your Task:
1.  Analyze the events: Find a sequence of events that form a logical multi-hop path. 
The time periods of the events should be overlapping or consecutive.
2.  Synthesize a description: Create a concise, high-level description for the 
entire multi-hop event. This will be your correct answer.
3.  Identify Nodes and Time: State the start node, end node, and the overall time 
window for the multi-hop event in terms of time steps.
4.  Generate a Question: Create a "question" asking for the most appropriate 
description of the relationship between the start and end nodes during those 
time steps.
    - Explicitly reference the interval as "time steps X-Y" and append 
    "(1 time step = {sampling_frequency})" in the question text.
5.  Generate Options: Create an "options" list with exactly four strings. 
The FIRST entry must be your synthesized description. The other three should be 
plausible but incorrect distractors.

Output Format:
Return ONLY a valid JSON object.
{{
    "question": "Your generated question about the multi-hop relationship.",
    "options": ["Your synthesized correct description", "Distractor 1", 
        "Distractor 2", "Distractor 3"]
}}
"""
\end{verbatim}
\end{tcolorbox}

\begin{tcolorbox}[
    colback=gray!15,
    colframe=gray!75,
    boxrule=0pt,
    arc=2mm,
    enhanced,
    breakable,
    title={Forecasting Reasoning Dataset Generation Template},
    fonttitle=\bfseries,
]
\label{prompt:align}
\begin{verbatim}

Analyze the provided context and generate a detailed forecast description.

Scenario Details:
- Target Node: {target_node_name} (ID: {target_node_id})
- Variable: {target_node_variable}
- Observation Window: Steps {history_window}
- Prediction Window: Steps {prediction_window}
- Key Event: {events}
- Statistical Hints: {referenced_stats}

Task:
Based *only* on the information above, provide a JSON object describing the forecast.

Output Format:
```json
{
  "observation_window": "{history_window}",
  "prediction_window": "{prediction_window}",
  "prediction_length": {prediction_length},
  "target_node_id": {target_node_id},
  "context_description": "{events}",
  "summary": "text describing the expected behaviour during the prediction window",
  "confidence": "low/medium/high"
}
```
\end{verbatim}
\end{tcolorbox}

\section{ST-Align Prompt Template}
\label{app:dataset_stalign_prompt}
\begin{tcolorbox}[
    colback=gray!15,
    colframe=gray!75,
    boxrule=0pt,
    arc=2mm,
    enhanced,
    breakable,
    title={Alignment Dataset Generation Template},
    fonttitle=\bfseries,
]
\label{prompt:align}
\begin{verbatim}

{
  "temporal": {
    "drift_type": "Question: What is the evolution pattern of node {node_id} during 
        the time range {time_range}? Answer: {drift_type}",
    "baseline": "Question: What is the long-term baseline value of node {node_id} 
        during the time range {time_range}? Answer: {baseline}",
    "kappa": "Question: What is the mean reversion speed (kappa) of node {node_id} 
        during the time range {time_range}? Answer: {kappa}",
    "sigma": "Question: What is the random fluctuation intensity (sigma) of node 
        {node_id} during the time range {time_range}? Answer: {sigma}",
    "lambda": "Question: What is the coupling strength (lambda) of node {node_id} 
        during the time range {time_range}? Answer: {lambda}",
    "diffusion_shape": "Question: What is the diffusion shape (diffusion_shape) of
        node {node_id} during the time range {time_range}? Answer: {diffusion_shape}",
    "sinusoidal_A": "Question: What is the sinusoidal amplitude (A) of 
        node {node_id} during the time range {time_range}? Answer: {A}",
    "sinusoidal_omega": "Question: What is the sinusoidal frequency (omega) of 
        node {node_id} during the time range {time_range}? Answer: {omega}",
    "sinusoidal_phi": "Question: What is the sinusoidal phase (phi) of node 
        {node_id} during the time range {time_range}? Answer: {phi}"
  },
  "spatial": {
    "edge_relationship": "Question: Is there a direct   connection from node {src} to 
    node {tgt}? Answer: {answer}",
    "indirect_connection": "Question: Is there an indirect path (through one or more 
        intermediate nodes) from node {src} to node {tgt}? Answer: {answer}"
  },
  "spatial_temporal": {
    "node_type": "Question: What is the type of node {node_id}? demand_source or 
        propagation? Answer: {node_type}",
    "edge_lag": "Question: What is the time lag between node {src} and node {tgt}? 
        Answer: {lag}",
    "edge_modulation": "Question: What is the modulation multiplier (multiplier) of 
     edge {edge} during the time range {time_range}? Answer: {multiplier}",
    "effective_coupling_strength": "Question: What is the effective coupling 
        strength of edge {edge} during the time range {time_range}? Answer: 
        {multiplier * base_adjacency}"
  }
}
\end{verbatim}
\end{tcolorbox}

\section{Agent Prompt Template}
\label{app:agent_prompt}
\begin{tcolorbox}[
    colback=gray!15,
    colframe=gray!75,
    boxrule=0pt,
    arc=2mm,
    enhanced,
    breakable,
    title={Scenario Generation Prompt (Agent 1)},
    fonttitle=\bfseries,
]
\begin{verbatim}
You are Agent 1: Scenario Generation Agent.

Your task:
Generate a realistic scenario description for a spatial-temporal dataset 
with {num_nodes} interconnected nodes.

CORE PRINCIPLES:
- Create synthetic but realistic scenarios with SPECIFIC, CONCRETE details
- Provide information that enables accurate time series generation
- Describe the physical system clearly (what flows, where, when, why)

NODE TYPE DEFINITIONS:
1. DEMAND_SOURCE (1 or 2 nodes):
Definition: Nodes that independently generate or consume the monitored variable.
Characteristics:
- Must specify baseline and amplitude values
- Must have exactly ONE self_generated peak (exogenous cycle)
- Any additional variations must be explicitly marked as propagated 
from other nodes

2. PROPAGATION
Definition: Relay nodes that transmit flows without independent generation.
Characteristics:
- Must specify a baseline value (nonzero, low). This represents a small 
(much smaller than the demand_source nodes), ambient background level and 
ensures physical realism (e.g., a river junction is never completely dry).
- Amplitude must equal 0
- peak must be null
- All variations must be propagated from other nodes

BASELINE CONSISTENCY RULE:
- All nodes (both DEMAND_SOURCE and PROPAGATION) must have baseline values within 
the same order of magnitude
- Baseline values should be similar across all nodes
- This ensures network coupling effects are meaningful and nodes can effectively 
influence each other

BASELINE REALISM RULE:
- The `baseline` value must reflect a realistic, physically plausible state for 
the node, often representing its value during a "calm" or "initial" period.
- This ensures the simulation starts from a sensible state and that the 
mean-reverting behavior is anchored to a meaningful physical value.

REQUIREMENTS:
1. Number nodes as NODE 0, NODE 1, NODE 2, ... (0-indexed)
2. All nodes monitor the SAME variable.
3. Specify spatial relationships at different time
4. Specify TIME SPAN and SAMPLING FREQUENCY such that total points are 
smaller than {max_seq_len}
5. Temporal dynamics rules:
- DEMAND_SOURCE nodes follow the above constraints (single exogenous peak + 
possible propagated variations)
- PROPAGATION nodes follow the above constraints (only propagated variations, 
no self-generated peaks)
6. Edge Consistency Rule:
- Any propagated variation described in TEMPORAL PATTERNS must correspond to an 
explicitly declared directed edge in the EDGES section.
- No hidden or undeclared propagation is allowed.
- The graph must be connected, ensuring that the effects from demand_source nodes 
can propagate to all other nodes.
7. Direction Integrity Rule:
- If a demand_source node generates an outbound peak (e.g., evening exodus from 
downtown), the corresponding outbound edge (e.g., NODE 2 → NODE 1) must be explicitly
listed in EDGES.
- Temporal patterns cannot contradict or introduce flows that are missing from the 
graph structure.
8. Demand Source Connectivity Rule:
- Every DEMAND_SOURCE node must have at least one outgoing edge, i.e., it must appear 
as the source node in at least one directed edge in the EDGES section.
9. Propagated Event Timing Consistency Rule (CRITICAL):
- When describing Edge Modulation for a propagating event (e.g., morning rush hour
traveling through multiple edges), you MUST account for cumulative time lags.
- Key Principle: An event cannot activate an edge before it physically arrives 
at that edge's source node.
- Design Strategy: Create staggered, overlapping time windows that shift 
forward by the time_lag amount for each successive edge in the chain.
10. Time Lag Design Guideline:
- Use time_lag>=1 only when the physical travel/transmission time is significant 
relative to sampling frequency
- For long chains (>3 nodes), consider small sampling frequency to keep cumulative 
delays 

AVOID:
- Vague phrases ("depends on conditions", "may vary")
- Real geographic names (cities, countries)
- Specific calendar dates (use relative time: "weekdays", "weekend")
- Special events or holidays
- Assigning multiple independent peaks to a single demand_source node
- More than 2 demand_source nodes

OUTPUT FORMAT (STRICT):
- TIME SPAN: [exact duration, e.g., "1 year","1 day", "1 week"]
- SAMPLING FREQUENCY: [exact interval, e.g., "1 day", "1 week", "1 hour"]
- VARIABLE: [single variable name, e.g., "traffic flow (vehicles/hour)", 
"power demand (MW)", "water temperature (°C)", "network bandwidth (Gbps)", 
"migration intensity (individuals/day)"]
- NODES:
  - NODE 0: [type: DEMAND_SOURCE or PROPAGATION] [description]
  - NODE 1: [type: DEMAND_SOURCE or PROPAGATION] [description]
  - ...
- EDGES:
  - NODE 0 → NODE 1: [relationship description, including any time lag, 
  e.g., "with a 5-day delay"]
  - NODE 1 → NODE 2: [relationship description]
  - NODE 1 → NODE 0: [relationship description]
  - NODE 2 → NODE 1: [relationship description]
  - ...
- TEMPORAL PATTERNS:
For each node, describe its periodicity characteristics, which may vary over time.
- NODE 0:
    - Can have multiple phases, each with a time period and behavior.
    - Example Phase 1 (time 0-239):
      - Behavior: stable, mean-reverting around a baseline
      - baseline: [numerical value with unit, e.g., "100 individuals/day"]
      - amplitude: 0
      - peak: null
    - Example Phase 2 (time 240-260):
      - Behavior: sinusoidal increase/decrease
      - baseline: [numerical value, same as other phases]
      - amplitude: [numerical value, < 5*baseline]
      - peak: [time step number, e.g., "250"]
    - propagated_variations: [if any, describe which nodes and when, 
    e.g., "receives flow from NODE 0 with 3-step delay, peaking around step 63"]
- NODE 1:
    - baseline: [numerical value with unit, MUST be similar to NODE 0's baseline, 
    e.g., if NODE 0 is 100, use 90-110 range]
    - amplitude: [numerical value, must be 0 for PROPAGATION nodes]
    - peak: [time step number or null]
    - propagated_variations: [describe propagation sources and timing]
- ... (repeat for all nodes, maintaining similar baseline values)
- Edge Modulation: Describe how edge influences vary over time 
    (must have at least one modulation)
    Format for each time-dependent modulation:
    - Time [time period, e.g., "50-70" or "7-9"]:
      - Edges affected: [e.g., "NODE 0 → NODE 1"]
      - Effect: [strong/moderate]
      - Description: [brief explanation]
\end{verbatim}
\end{tcolorbox}

\begin{tcolorbox}[
    colback=gray!15,
    colframe=gray!75,
    boxrule=0pt,
    arc=2mm,
    enhanced,
    breakable,
    title={Scenario Parsing Prompt (Agent 2)},
    fonttitle=\bfseries,
]
\begin{verbatim}
You are Agent 2: Scenario Parsing Agent.

Your task: Convert a natural language scenario description into a STRICT, 
STRUCTURED JSON object.

INPUT: Natural language scenario description (from Agent 1)

OUTPUT: A single valid JSON object with NO markdown, NO explanations, NO comments, 
NO trailing commas.

---

JSON SCHEMA (STRICT):
{
  "time_span": "string (e.g., '7 days')",
  "sampling_frequency": "string (e.g., '1 hour')",
  "seq_len": "integer (number of time steps)",
  "variable": "string (exactly ONE variable monitored by all nodes)",
  "nodes": [
    {"id": 0, "type": "demand_source or propagation", "name": "string", 
    "description": "string"},
    {"id": 1, "type": "demand_source or propagation", "name": "string", 
    "description": "string"},
    ...
  ],
  "edges": [
    {
      "source": 0, 
      "target": 1, 
      "relationship": "string describing directional influence",
      "time_lag": "integer (optional, number of time steps for delay)"
    },
    {
      "source": 1, 
      "target": 2, 
      "relationship": "string describing directional influence",
      "time_lag": "integer (optional)"
    },
    ...
  ],
  "drift_patterns": {
    "repeat": "boolean (optional, indicates if the pattern sequence repeats)",
    "repeat_period": "integer (optional, defines the cycle duration in steps 
    if repeat is true, e.g., 24 for a daily cycle)",
    "nodes": [
      {
        "id": "integer (node ID)",
        "patterns": [
          {
            "time_range": "[start_time, end_time] (integer array)",
            "behavior": "string (e.g., 'mean_reverting', 'sinusoidal')",
            "baseline": "number (long-term average level, must be > 0)",
            "amplitude": "number (peak deviation from baseline, >= 0)",
            "peak": "integer or null (time step of the peak for sinusoidal behavior)"
          }
        ],
        "propagated_variations": [
          {
            "time": "string (time location or range)",
            "origin": "propagated",
            "source": "integer (node_id)",
            "delay": "string (optional, e.g., '3 days', '2 hours')",
            "description": "string (short explanation)"
          }
        ]
      }
    ]
  },
  "adjacency_modulation": {
    "patterns": [
      {
        "time_period": "string (e.g., '50-70', '7-9')",
        "effect": "string (strong/moderate)",
        "applies_to": "string or array of strings (e.g., '0->1' or ['0->1', '1->2'])",
        "description": "string (explanation of why this modulation occurs)"
      }
    ]
  },
  "spatial_layout": {
    "0": {"x": number, "y": number},
    "1": {"x": number, "y": number},
    ...
  }
}

PARSING RULES:
0. Calculate seq_len:
- Extract the numeric values from time_span and sampling_frequency
- Convert both to the same unit (e.g., hours, days)
- Calculate: seq_len = time_span / sampling_frequency
1. Node Classification:
- If description mentions "generate", "originate", "consume", "demand", "source"
→ demand_source
- If description mentions "relay", "connector", "junction", "pass through", 
"transmit" → propagation
- Each node must be classified based on its physical role
2. Edge Construction:
- Extract all directional influences from scenario description
- For each edge, extract these attributes:
 * source: source node ID
 * target: target node ID
 * relationship: brief description of the connection
 * time_lag: (optional) integer representing delay in time steps 
 (e.g., if scenario says "5 day delay" and sampling is "1 day", time_lag should be 5)
3. Drift Patterns:
- This section describes the time-varying behavior of each node.
- For each node, parse its temporal description into a list of `patterns`.
- Each pattern in the list must describe a specific behavior over a `time_range`, 
and include:
 * `baseline`: The typical long-term average value. This must be > 0.
 * `amplitude`: The peak deviation from the baseline. This must be >= 0.
 * `peak`: The time step where the peak occurs (for `sinusoidal` behavior). 
 Must be null for other behaviors.
- Parse any `propagated_variations` described for the node.
- If the patterns repeat (e.g., a daily cycle), set `repeat: true` and define the
cycle's duration in `repeat_period` (e.g., `24` for a 24-hour cycle).
- Coverage Constraint: If `repeat` is true, the `time_range` of all patterns for 
a node must completely and contiguously cover the range from `0` to `repeat_period`.
- Constraints per Node Type (CRITICAL):
 - For demand_source nodes: Can have patterns with `amplitude` > 0.
 - For propagation nodes: All patterns MUST have `amplitude: 0` and `peak: null`. 
 Their variation comes only from `propagated_variations`.
4. Adjacency Modulation:
- Extract concrete time-dependent edge effects from scenario
- CRITICAL: For propagating events (e.g., traffic flowing through a chain of nodes), 
each edge in the path should have its own modulation entry with a properly staggered 
time_period that accounts for the cumulative time_lag
- Describe modulation patterns with:
 * time_period: when the modulation occurs (e.g., "50-70", "7-9" - just numbers 
 representing time steps)
   - For event chains, ensure each edge's time_period starts AFTER the event could 
   have arrived from the previous edge
   - If edge A->B has time_lag=1 and modulation starts at t=15, then edge B->C should 
   have modulation starting at t>=16
 * effect: strength of the modulation (strong/moderate)
   - strong: significant enhancement of edge influence
   - moderate: moderate enhancement of edge influence
 * applies_to: which edge(s) are affected
   - Can be a single edge string (e.g., "0->1")
   - Can be an array of edge strings (e.g., ["0->1", "1->2"]) ONLY when they truly share 
   the exact same time window (be cautious with this for event chains)
 * description: explanation of why this modulation happens

5. Spatial Layout:
   - Generate simple 2D coordinates for visualization
   - Arrange nodes logically (e.g., source on left, propagation in middle, 
   sink on right)

6. Output Format:
   - Valid JSON only (RFC 8259)
   - Double quotes for strings
   - No trailing commas
   - No markdown code blocks
   - No extra text

Example: [See source code]

INPUT SCENARIO:
{scenario}
\end{verbatim}
\end{tcolorbox}

\begin{tcolorbox}[
    colback=gray!15,
    colframe=gray!75,
    boxrule=0pt,
    arc=2mm,
    enhanced,
    breakable,
    title={SDE Parameters Generation Prompt (Agent 3)},
    fonttitle=\bfseries,
]
\begin{verbatim}
You are Agent 3: SDE Parameters Generation Agent.

Your task: Generate hierarchical SDE parameters from a structured scenario JSON.

INPUT: Structured JSON from Agent 2 (scenario parsing agent)

OUTPUT: Hierarchical SDE parameters as strict JSON (NO markdown, NO comments)

SDE MODEL (per node i):
dX_i(t) = [ drift_i(t, X_i) + lambda_i * capital_sigma_j A_ji(t) * (X_j - X_i) ] dt 
+ sigma_i * g_i(X_i) dW_i(t)

Components:
- drift_i: drift term (type-dependent)
- lambda_i: coupling strength
- A_ji(t): time-varying adjacency (from Agent 4)
- sigma_i: base volatility
- g_i(X_i): diffusion shape function

DRIFT TYPES:
1. mean_reverting (default):
   - Formula: drift = kappa * (mu_t - X_t)
   - Parameters: kappa (mean reversion speed), baseline (mu_t)
   - Constraint: 0.01 < kappa < 0.5
   - Usage: REQUIRED for propagation nodes, allowed for demand_source nodes
2. constant:
   - Formula: drift = alpha
   - Parameters: alpha (constant drift rate)
   - Constraint: alpha \in R
   - Usage: ONLY allowed for demand_source nodes
3. sinusoidal:
   - Formula: drift = kappa * (baseline + A*sin(omega*t + phi) - X_t)
   - Parameters: A (amplitude), omega (frequency), phi (phase shift)
   - Constraint: A > 0, omega > 0, phi \in R (ALL SCALARS, NOT ARRAYS)
   - CRITICAL: Single harmonic only - no multi-frequency superposition
   - Usage: ONLY allowed for demand_source nodes
4. logistic:
   - Formula: drift = r * X_t * (1 - X_t/baseline)
   - Parameters: r (growth rate), baseline (carrying capacity)
   - Constraint: 0 < r < 0.1, baseline > 0
   - Usage: Allowed for both demand_source and propagation nodes

CRITICAL CONSTRAINTS (STRICTLY ENFORCED):
1. Node Type Constraints:
   - propagation nodes: MAY use mean_reverting or logistic drift (small r)
   - demand_source nodes: MAY use mean_reverting, sinusoidal, constant, or logistic
2. Parameter Ranges (for stability):
   - 0.01 < kappa < 0.5 (mean reversion speed)
   - 0.8 < lambda < 1.5 (coupling strength - high values for 
   realistic network dynamics)
   - sigma < 0.01*baseline (volatility, must be less than 1% of the baseline)
   - For sinusoidal: A, omega, phi must be scalars (not arrays)
   - For logistic: 0 < r < 0.1, baseline > 0
3. Propagation Node Special Rules:
   - Use LOW kappa (0.05-0.2) for weak self-reversion
   - Use HIGH lambda (1.0-1.5) for strong neighbor coupling
   - This ensures propagation nodes relay upstream flows effectively
4. Diffusion Shapes:
   - "constant": g(X) = 1
   - "sqrt": g(X) = sqrt(|X| + 1e-6)
   - "linear": g(X) = 1 + alpha*|X|

HIERARCHICAL OUTPUT STRUCTURE:
{{
  "global_defaults": {{
    "drift_type": "mean_reverting",
    "node_type": "demand_source",
    "kappa": 0.25,
    "baseline": 50.0,
    "lambda": 1.0,
    "sigma": 2.0,
    "diffusion_shape": "constant"
  }},
  "group_params": {{
    "demand_sources": {{
      "node_type": "demand_source",
      "drift_type": "sinusoidal",
      "baseline": 100.0,
      "A": 30.0,
      "omega": 0.2618,
      "phi": 0.0,
      "kappa": 0.25,
      "lambda": 1.0,
      "sigma": 2
    }},
    "propagation_nodes": {{
      "node_type": "propagation",
      "drift_type": "mean_reverting",
      "baseline": 50.0,
      "kappa": 0.1,
      "lambda": 1.2,
      "sigma": 2
    }}
  }},
  "node_overrides": {{
    "0": {{
      "group": "demand_sources",
      "drift_patterns": [
        {{
          "time_range": [0, 239],
          "drift_type": "mean_reverting",
          "baseline": 100,
          "kappa": 0.2
        }},
        {{
          "time_range": [240, 260],
          "drift_type": "sinusoidal",
          "baseline": 100,
          "A": 90,
          "omega": 0.0172,
          "phi": -2.557,
          "kappa": 0.35
        }}
      ],
      "description": "Node with time-varying drift"
    }},
    "1": {{
      "group": "propagation_nodes",
      "description": "Connector highway - pure relay"
    }},
    "2": {{
      "group": "demand_sources",
      "baseline": 80.0,
      "phi": 3.14159,
      "A": 30.0,
      "omega": 0.2618,
      "description": "Business district - midday peak (phase shifted)"
    }}
  }},
}}

INPUT JSON:
{structured_scenario}

RETURN ONLY VALID JSON (no markdown, no comments).
\end{verbatim}
\end{tcolorbox}

\begin{tcolorbox}[
    colback=gray!15,
    colframe=gray!75,
    boxrule=0pt,
    arc=2mm,
    enhanced,
    breakable,
    title={Agent Time-Varying Adjacency Generation Prompt},
    fonttitle=\bfseries,
]
\begin{verbatim}
You are Agent 4: Time-Varying Adjacency Generation Agent.

Your task: Generate time-varying modulation rules from structured scenario JSON.

INPUT: Structured JSON from Agent 2 (scenario parsing agent)

OUTPUT: Time modulation configuration as strict JSON (NO markdown, NO comments)

---

TASK SPECIFICATION:

Generate time_modulation: Rules for how edge weights vary over time
- Derived STRICTLY from "adjacency_modulation" field in input JSON
- Each pattern specifies time ranges and edge-specific multipliers

NOTE: Base adjacency matrix is handled separately by the system. 
You only need to generate time modulation rules.

TIME MODULATION CONSTRUCTION:
Extract from input JSON "adjacency_modulation.patterns" field:
For each pattern in the input JSON:
1. Extract "time_period" (e.g., "7-9", "25-55", "150-240") and convert to 
[start, end]
2. Extract "effect" (strong/moderate) and map to multiplier
3. Extract "applies_to" (e.g., "0->1" or "0->1, 1->2") and parse edges
4. Extract "description"

Effect to Multiplier mapping:
- strong → multiplier: 10-20
- moderate → multiplier: 5-10

Output format (simplified, no daily/seasonal distinction):
{{
  "time_modulation": {{
    "patterns": [
      {{
        "time_range": [start, end],
        "description": "...",
        "edge_modulations": {{
          "source->target": {{"multiplier": value, "description": "..."}},
          ...
        }}
      }}
    ]
  }}
}}

MULTIPLIER INTERPRETATION:
Final edge weight at time t:
- weight(t) = base_adjacency[i][j] * multiplier(t)
- Base adjacency is handled by the system (you don't generate it)
Multiplier values from effect mapping:
- strong effect: 10-20
- moderate effect: 5-10
- No modulation: 1.0 (edge weight unchanged, default when time is outside all 
pattern ranges)

CRITICAL RULES:
1. Time Modulation:
   - Extract patterns from input JSON "adjacency_modulation.patterns" array
   - Do NOT invent new patterns or time ranges
   - Map "effect" field to multiplier: strong=10-20, moderate=5-10
   - Output as unified "patterns" array (no daily/seasonal/weekly distinction)

2. Edge Specification:
   - Format: "source->target" (e.g., "0->1", "1->2")
   - Use "all_edges" if input JSON applies_to = "all_edges"
   - Otherwise, parse input JSON applies_to field (e.g., "0->1, 1->2" 
   → separate entries)

3. Time Ranges:
   - Parse "time_period" from input JSON (e.g., "7-9", "25-55", "150-240")
   - Convert to [start, end] integer array
   - No distinction between hourly/daily/seasonal - just numerical ranges

4. Output Format:
   - Valid JSON only (RFC 8259)
   - No markdown code blocks
   - No comments
   - No trailing commas

---

OUTPUT JSON SCHEMA:

{{
  "time_modulation": {{
    "patterns": [
      {{
        "time_range": [7, 9],
        "description": "Morning rush hour strengthens residential to highway flow",
        "edge_modulations": {{
          "0->1": {{"multiplier": 15, "description": "Strong effect on
          commuter flow"}}
        }}
      }},
      {{
        "time_range": [17, 19],
        "description": "Evening rush hour moderately strengthens highway to 
        business flow",
        "edge_modulations": {{
          "1->2": {{"multiplier": 10, "description": "Moderate effect on
          highway flow"}}
        }}
      }}
    ]
  }}
}}

EXAMPLE: [See source code]

INPUT JSON:
{structured_scenario}

RETURN ONLY VALID JSON (no markdown, no comments).
\end{verbatim}
\end{tcolorbox}

\begin{tcolorbox}[
    colback=gray!15,
    colframe=gray!75,
    boxrule=0pt,
    arc=2mm,
    enhanced,
    breakable,
    title={Agent Judge Scenario Parsing Prompt},
    fonttitle=\bfseries,
]
\begin{verbatim}

You are Judge Agent 1, a meticulous diagnostic expert responsible for 
two-level validation.

You will receive:
1. Original Scenario Text: Natural language description from Agent 1
2. Parsed Structured JSON: Structured data from Agent 2

Your mission is to determine if they are consistent, logical, and ready 
for simulation. Most importantly, if there is an error, you must diagnose the source: 
is it Agent 1's scenario logic or Agent 2's parsing accuracy?

DIAGNOSTIC PROCESS (FOLLOW THIS ORDER):
STEP 1: PARSING FIDELITY ASSESSMENT (Evaluating Agent 2)
Assume the Original Scenario Text is correct. Compare the Parsed JSON against 
it meticulously.
Check for:
1. Node Count Accuracy: Does the JSON contain exactly {expected_num_nodes} nodes 
as required?
2. Entity Completeness: Are all nodes and edges from the text present in JSON?
3. Type Accuracy: Are node types (demand_source/propagation) correctly assigned?
4. Attribute Accuracy: 
   - Are all edge relationships correctly represented?
   - Are time_lag values correctly extracted as integers?
5. Value Extraction:
   - Time span and sampling frequency correctly extracted?
   - Baseline, amplitude, and peak values match the text?
   - Propagated variations correctly parsed with source nodes and timings?
6. Structure Completeness:
   - Are adjacency_modulation patterns (time periods, effects, edges) 
   fully captured?
   - Are drift_patterns accurately representing the temporal evolution described?
If you find ANY discrepancy between the text and JSON, this is Agent 2's error.
- Set `error_source: "agent2"`
- List specific parsing mismatches in `issues`
- Stop here and do NOT proceed to Step 2

STEP 2: SCENARIO LOGIC ASSESSMENT (Evaluating Agent 1)
Only proceed if you are confident the JSON is a FAITHFUL representation of the text.
Now analyze the scenario's internal logic using ONLY the structured JSON:
1. Propagated Event Timing Consistency (CRITICAL):
- Identify event propagation chains in adjacency_modulation 
(e.g., edges forming a path like 0->1->3->2)
- For each edge in a chain, verify its modulation time_range respects preceding 
edges' time_lag
- Calculation Example:
 * Event path: 0 -> 1 -> 3 -> 2
 * Edge (0->1) has time_lag=1, modulation starts at t_start_1=15
 * Event arrives at Node 1 at t_arrival_1 = t_start_1 + time_lag = 15 + 1 = 16
 * Therefore, edge (1->3) modulation MUST start at t >= 16
 * If edge (1->3) modulation starts at t=15, this is IMPOSSIBLE
- This error indicates Agent 1 failed to account for propagation delays
2. Graph-Temporal Consistency:
- Does every propagated_variation have a corresponding incoming edge?
- Does every demand_source node have at least one outgoing edge?
- Are propagated_variation timings consistent with edge time_lags?
- Is the graph connected to ensure that the effects from demand_source nodes can 
propagate to all other nodes?
3. Physical Realism:
- Are all baseline values within similar order of magnitude?
- Do demand_source nodes have amplitude > 0 and exactly one self-generated peak?
- Do propagation nodes have amplitude = 0 and peak = null?
4. Cumulative Delay vs Event Duration:
- Calculate total time lag along critical paths
- Compare to the duration of edge_modulation events describing that path
- If cumulative lag >> event duration, the scenario is unrealistic
If you find logical inconsistencies in the scenario design, this is Agent 1's error.
- Set `error_source: "agent1"`
- Provide specific, actionable suggestions for scenario redesign

OUTPUT FORMAT (STRICT JSON):
You MUST respond with a single JSON object. No markdown blocks, no extra text.
{{
  "approved": boolean,
  "error_source": "agent1" | "agent2" | null,
  "feedback": "Brief one-sentence summary for control loop routing",
  "issues": [
    {{
      "type": "Parsing Fidelity" | "Scenario Logic",
      "field": "specific field or path (e.g., 'edges[2].time_lag', 
      'adjacency_modulation.path_0->1->3')",
      "problem": "Detailed description of the issue",
      "suggestion": "Clear, actionable fix for the responsible agent"
    }}
  ],
  "overall_comment": "Comprehensive assessment explaining the decision"
}}
Rules:
- If approved=true, set error_source=null and issues=[]
- If approved=false, error_source MUST be either "agent1" or "agent2"
- Step 1 errors always result in error_source="agent2"
- Step 2 errors always result in error_source="agent1"

Original Scenario Description:
{scenario}

Parsed Structured JSON:
{parsed_json}

Begin your two-step diagnostic analysis now.

\end{verbatim}
\end{tcolorbox}

\begin{tcolorbox}[
    colback=gray!15,
    colframe=gray!75,
    boxrule=0pt,
    arc=2mm,
    enhanced,
    breakable,
    title={Agent Judge Parameter Validation Prompt},
    fonttitle=\bfseries,
]
\begin{verbatim}
You are a validation agent responsible for checking if the SDE parameters and 
generated time series are reasonable and consistent with the scenario.

Scenario Description (from Agent 2):
{structured_scenario}

SDE Parameters Generated:
{sde_parameters}

Time-Varying Adjacency:
{time_varying_adjacency}
{previous_assessment_section}
Your Task:
Analyze the attached time series visualization and assess:
1. Verification of Fixes: If a previous assessment is provided, first verify if the 
suggested changes have been implemented and if the previous issues are resolved.
2. Time Series Patterns: Do the visualized curves match the scenario's 
`drift_patterns`? Check for correct transitions between behaviors 
(e.g., stable `mean_reverting` or `logistic` trends vs. dynamic `sinusoidal` peaks).
3. Parameter Plausibility: Within each `drift_pattern`, are the SDE parameter 
values (kappa, lambda, sigma) reasonable?
4. Baseline Consistency: Is the baseline value consistent across different patterns 
for the same node, as described in the scenario?
5. Drift Type Correctness: Does the sequence of assigned drift types in the 
parameters match the intended behaviors in the scenario?
6. Coupling Effects: Are the time-varying coupling strengths (edge weights) from 
the adjacency matrix correctly reflected in the simulation? For example, during a 
'strong' modulation period, is there a visible and significant influence between 
the connected nodes?
7. Dependency Flow: Do propagation nodes show clear dependency on the demand source 
nodes they are connected to?
8. Simulation Stability: Are there any unrealistic behaviors like explosive growth,
flatlining, or excessive, non-physical oscillations?
9. Noise Level (Sigma): Is the level of noise (random fluctuations) appropriate for 
the scenario? If the noise is so large that it completely hides the underlying trend 
(e.g., the sinusoidal peak), suggest reducing `sigma`. If the curve looks too smooth 
and artificial, suggest increasing `sigma`.

CRITICAL CONSTRAINTS FOR ADJACENCY ISSUES:
- You can give suggestions to increase or decrease the multiplier within the range of
10-20 for 'strong' and 5-10 for 'moderate'.
- If a propagated peak or pattern is not obvious, suggest INCREASING the `multiplier`
for the relevant edge. If a curve is excessively unstable or over-reacts to another 
node, suggest DECREASING the `multiplier`.

Response Format:
Provide your assessment in the following JSON format. Focus ONLY on suggesting 
specific parameter changes, do not give implementation advice.
```json
{{
  "approved": true/false,
  "parameter_issues": [
    {{
      "node_id": "node identifier",
      "parameter": "parameter name (e.g., 'kappa', 'baseline', 'drift_type')",
      "current_value": "current value",
      "problem": "description of the problem",
      "suggested_value": "suggested new value or range"
    }}
  ],
  "adjacency_issues": [
    {{
      "edge": "edge identifier (e.g., '0->1')",
      "problem": "description of the problem",
      "suggestion": "how to fix it"
    }}
  ],
  "visual_assessment": "analysis of the time series patterns and their consistency 
  with the scenario",
  "overall_comment": "overall assessment and guidance for revision"
}}
```

If everything looks good, set "approved": true and keep issue lists empty.
If there are problems, set "approved": false and provide detailed suggestions.
\end{verbatim}
\end{tcolorbox}

\section{Spatial Effect Prompt Template}
\label{app:spatial_effect_prompt}
\begin{tcolorbox}[
    colback=gray!15,
    colframe=gray!75,
    boxrule=0pt,
    arc=2mm,
    enhanced,
    breakable,
    title={Spatial Effect Prompt},
    fonttitle=\bfseries,
]
\label{prompt:align}
\begin{verbatim}

Does this response correctly reason with spatial/graph structure?

YES requires BOTH:
1. References graph structure (edges, connections, node relationships)
2. Uses it to reason about the answer (e.g., "A influences B because of edge A→B", 
"propagation follows the graph", "connected nodes show similar patterns")

NO if:
- Only describes the graph without reasoning with it
- Only analyzes nodes independently
- Mentions structure but reasoning is incorrect or irrelevant

Question: {question}
Response: {response}

Label: yes or no

\end{verbatim}
\end{tcolorbox}

\begin{lstlisting}[language=json,caption={Scenario parsing Agent},label={lst:Scenario parsing Agent}]
{
  "time_span": "1 day",
  "sampling_frequency": "30 minutes",
  "seq_len": 48,
  "variable": "traffic flow (vehicles/hour)",
  "nodes": [
    {
      "id": 0,
      "type": "demand_source",
      "name": "Central Business District",
      "description": "Central business district with offices and commercial areas"
    },
    {
      "id": 1,
      "type": "propagation",
      "name": "Highway Junction",
      "description": "Major highway junction connecting business district to residential area"
    },
    {
      "id": 2,
      "type": "demand_source",
      "name": "Residential Suburb",
      "description": "Large residential suburb area"
    }
  ],
  "edges": [
    {
      "source": 0,
      "target": 1,
      "relationship": "Primary outbound route from business district",
      "time_lag": 1
    },
    {
      "source": 1,
      "target": 2,
      "relationship": "Main highway section leading to residential area",
      "time_lag": 1
    },
    {
      "source": 2,
      "target": 1,
      "relationship": "Inbound commuter route from residential area",
      "time_lag": 1
    },
    {
      "source": 1,
      "target": 0,
      "relationship": "Primary inbound route to business district",
      "time_lag": 1
    }
  ],
  "drift_patterns": {
    "repeat": true,
    "repeat_period": 48,
    "nodes": [
      {
        "id": 0,
        "patterns": [
          { "time_range": [0,13], "behavior": "mean_reverting", "baseline": 100, "amplitude": 0, "peak": null },
          { "time_range": [14,34], "behavior": "sinusoidal", "baseline": 100, "amplitude": 300, "peak": 17 },
          { "time_range": [35,48], "behavior": "mean_reverting", "baseline": 100, "amplitude": 0, "peak": null }
        ],
        "propagated_variations": [
          { "time": "16-20", "origin": "propagated", "source": 2,
            "description": "Receives morning commute flow from residential area via highway junction" }
        ]
      },
      {
        "id": 1,
        "patterns": [
          { "time_range": [0,48], "behavior": "mean_reverting", "baseline": 120, "amplitude": 0, "peak": null }
        ],
        "propagated_variations": [
          { "time": "18-22", "origin": "propagated", "source": 0,
            "description": "Receives evening exodus from business district" },
          { "time": "15-19", "origin": "propagated", "source": 2,
            "description": "Receives morning commute from residential area" }
        ]
      },
      {
        "id": 2,
        "patterns": [
          { "time_range": [0,13], "behavior": "mean_reverting", "baseline": 110, "amplitude": 0, "peak": null },
          { "time_range": [14,34], "behavior": "sinusoidal", "baseline": 110, "amplitude": 280, "peak": 14 },
          { "time_range": [35,48], "behavior": "mean_reverting", "baseline": 110, "amplitude": 0, "peak": null }
        ],
        "propagated_variations": [
          { "time": "19-23", "origin": "propagated", "source": 0,
            "description": "Receives evening exodus flow from business district via highway junction" }
        ]
      }
    ]
  },
  "adjacency_modulation": {
    "patterns": [
      { "time_period": "14-19", "effect": "strong", "applies_to": ["2->1","1->0"],
        "description": "Morning rush hour intensifies inbound traffic flow" },
      { "time_period": "17-22", "effect": "strong", "applies_to": ["0->1","1->2"],
        "description": "Evening rush hour intensifies outbound traffic flow" }
    ]
  },
  "spatial_layout": {
    "0": { "x": 0, "y": 0 },
    "1": { "x": 1, "y": 0 },
    "2": { "x": 2, "y": 0 }
  },
  "domain": "Transportation",
  "task_id": "task_0030"
}
\end{lstlisting}

\begin{lstlisting}[
  language=json,
  caption={SDEs Parameters Agent},
  label={lst:sde_parameters}
]
{
  "global_defaults": {
    "drift_type": "mean_reverting",
    "node_type": "demand_source",
    "kappa": 0.25,
    "baseline": 50.0,
    "lambda": 1.0,
    "sigma": 0.5,
    "diffusion_shape": "sqrt"
  },
  "group_params": {
    "demand_sources": {
      "node_type": "demand_source",
      "drift_type": "sinusoidal",
      "baseline": 100.0,
      "A": 30.0,
      "omega": 0.1309,
      "phi": 0.0,
      "kappa": 0.25,
      "lambda": 0.9,
      "sigma": 0.5
    },
    "propagation_nodes": {
      "node_type": "propagation",
      "drift_type": "mean_reverting",
      "baseline": 120.0,
      "kappa": 0.25,
      "lambda": 0.8,
      "sigma": 0.3
    }
  },
  "node_overrides": {
    "0": {
      "group": "demand_sources",
      "drift_patterns": [
        { "time_range": [0,13], "drift_type": "mean_reverting", "baseline": 100, "kappa": 0.25, "sigma": 0.5 },
        { "time_range": [14,34], "drift_type": "sinusoidal", "baseline": 100, "A": 300, "omega": 0.1309, "phi": -1.7262, "kappa": 0.25, "sigma": 0.8 },
        { "time_range": [35,48], "drift_type": "mean_reverting", "baseline": 100, "kappa": 0.25, "sigma": 0.5 }
      ],
      "description": "Central business district with offices and commercial areas"
    },
    "1": {
      "group": "propagation_nodes",
      "drift_patterns": [
        { "time_range": [0,48], "drift_type": "mean_reverting", "baseline": 120, "kappa": 0.25, "lambda": 0.8, "sigma": 0.3 }
      ],
      "description": "Major highway junction connecting business district to residential area"
    },
    "2": {
      "group": "demand_sources",
      "drift_patterns": [
        { "time_range": [0,13], "drift_type": "mean_reverting", "baseline": 110, "kappa": 0.25, "sigma": 0.5 },
        { "time_range": [14,34], "drift_type": "sinusoidal", "baseline": 110, "A": 280, "omega": 0.1309, "phi": -1.2217, "kappa": 0.25, "sigma": 0.8 },
        { "time_range": [35,48], "drift_type": "mean_reverting", "baseline": 110, "kappa": 0.25, "sigma": 0.5 }
      ],
      "description": "Large residential suburb area"
    }
  }
}
\end{lstlisting}

\begin{lstlisting}[
  language=json,
  caption={Time-Varying Adj. Agent},
  label={lst:time_varying_adj}
]
{
  "time_modulation": {
    "patterns": [
      {
        "time_range": [14,19],
        "description": "Morning rush hour intensifies inbound traffic flow",
        "edge_modulations": {
          "2->1": { "multiplier": 15, "description": "Strong inbound commuter flow" },
          "1->0": { "multiplier": 15, "description": "Strong inbound business district flow" }
        }
      },
      {
        "time_range": [17,22],
        "description": "Evening rush hour intensifies outbound traffic flow",
        "edge_modulations": {
          "0->1": { "multiplier": 15, "description": "Strong outbound business district flow" },
          "1->2": { "multiplier": 15, "description": "Strong outbound residential flow" }
        }
      }
    ]
  },
  "base_adjacency": [
    [0.0, 0.1, 0.0],
    [0.1, 0.0, 0.1],
    [0.0, 0.1, 0.0]
  ]
}
\end{lstlisting}

\end{document}